\documentclass[a4paper,11pt]{article}\usepackage{jmlr2e,url,colortbl,graphicx,subcaption,appendix,
xparse,amsmath,mathtools,etoolbox,patchcmd,stmaryrd,xspace,booktabs,
colortbl,enumitem,microtype,mathptmx}
\definecolor{citegreen}{rgb}{0.1,0.5,0.1}
\definecolor{tabgrey}{rgb}{0.85,0.85,0.85}
\usepackage[dvipsnames]{xcolor}
\usepackage{rotating}
\usepackage{pdflscape}
\mathtoolsset{showonlyrefs}
\usepackage[a4paper,lmargin=1.25in,rmargin=1.25in,bottom=1.25in,top=1in]{geometry}

\usepackage[colorlinks=true, linkcolor=NavyBlue, citecolor=NavyBlue, %
	urlcolor=RoyalBlue, linktocpage, plainpages=false]{hyperref} 

\definecolor{Links}{RGB}{0,0,128}
\makeatletter

\NewDocumentCommand\Def{m g}{%
    \colorlet{temp}{.}\color{DarkRed}\relax\ifmmode{#1}\else\emph{#1}\fi\color{temp}%
    \IfNoValueF{#2}{\coloneqq #2}
    \xspace
    }

\makeatother

\definecolor{Base00} {HTML}{ffffff} 
\definecolor{Base01} {HTML}{e0e0e0} 
\definecolor{Base02} {HTML}{d6d6d6} 
\definecolor{Base03} {HTML}{8e908c} 
\definecolor{Base04} {HTML}{969896} 
\definecolor{Base05} {HTML}{4d4d4c} 
\definecolor{Base06} {HTML}{282a2e} 
\definecolor{Base07} {HTML}{1d1f21} 
\definecolor{Red}    {HTML}{c82829} 
\definecolor{Base09} {HTML}{f5871f} 
\definecolor{Yellow} {HTML}{eab700} 
\definecolor{Green}  {HTML}{718c00} 
\definecolor{Cyan}   {HTML}{3e999f} 
\definecolor{Blue}   {HTML}{4271ae} 
\definecolor{Magenta}{HTML}{8959a8} 
\definecolor{Base0F} {HTML}{a3685a} 

\colorlet{LightBase00}{Base00!20}
\colorlet{LightBase01}{Base01!20}
\colorlet{LightBase02}{Base02!20}
\colorlet{LightBase03}{Base03!20}
\colorlet{LightBase04}{Base04!20}
\colorlet{LightBase05}{Base05!20}
\colorlet{LightBase06}{Base06!20}
\colorlet{LightBase07}{Base07!20}
\colorlet{LightRed}{Red!20}
\colorlet{LightBase09}{Base09!20}
\colorlet{LightYellow}{Yellow!20}
\colorlet{LightGreen}{Green!20}
\colorlet{LightCyan}{Cyan!20}
\colorlet{LightBlue}{Blue!20}
\colorlet{LightMagenta}{Magenta!20}
\colorlet{LightBase0F}{Base0F!20}

\colorlet{DarkBase00}{Base00!90!black}
\colorlet{DarkBase01}{Base01!90!black}
\colorlet{DarkBase02}{Base02!90!black}
\colorlet{DarkBase03}{Base03!90!black}
\colorlet{DarkBase04}{Base04!90!black}
\colorlet{DarkBase05}{Base05!90!black}
\colorlet{DarkBase06}{Base06!90!black}
\colorlet{DarkBase07}{Base07!90!black}
\colorlet{DarkRed}{Red!90!black}
\colorlet{DarkBase09}{Base09!90!black}
\colorlet{DarkYellow}{Yellow!90!black}
\colorlet{DarkGreen}{Green!90!black}
\colorlet{DarkCyan}{Cyan!90!black}
\colorlet{DarkBlue}{Blue!90!black}
\colorlet{DarkMagenta}{Magenta!90!black}
\colorlet{DarkBase0F}{Base0F!90!black}

\AtBeginDocument{
    \newcolumntype{L}[1]{>{\raggedright\let\newline\\\arraybackslash\hspace{0pt}}m{#1}}
    \newcolumntype{C}[1]{>{\centering\let\newline\\\arraybackslash\hspace{0pt}}m{#1}}
    \newcolumntype{R}[1]{>{\raggedleft\let\newline\\\arraybackslash\hspace{0pt}}m{#1}}
}
\usepackage{accents}

\usepackage{tikz, pgfplots}
\usetikzlibrary{intersections, calc, positioning, decorations.pathreplacing}
\usetikzlibrary{external}
\tikzsetexternalprefix{figures/}
\usepgfplotslibrary{fillbetween}
\pgfplotsset{compat=1.14}
\tikzset{
  set/.style   = thick,
  faint/.style = {color=black!30},
  dot/.style = {draw, circle, fill=black, inner sep=1.5pt},
  dashrounded/.style = {draw, rounded corners=3pt, dashed, faint},
  rounded/.style   = {draw, rounded corners=3pt, minimum height=1.5em},
  prominent/.style   = {fill=white, inner sep=1pt, rounded corners=3pt},
  every picture/.style={font issue=\footnotesize},
  font issue/.style={execute at begin picture={#1\selectfont}},
  right angle quadrant/.code={
      \pgfmathsetmacro\quadranta{{1,1,-1,-1}[#1-1]}     
      \pgfmathsetmacro\quadrantb{{1,-1,-1,1}[#1-1]}},
  right angle quadrant=1, 
  right angle length/.code={\def\rightanglelength{#1}},   
  right angle length=1.5ex, 
  right angle symbol/.style n args={3}{
      insert path={
          let \p0 = ($(#1)!(#3)!(#2)$) in     
              let \p1 = ($(\p0)!\quadranta*\rightanglelength!(#3)$), 
              \p2 = ($(\p0)!\quadrantb*\rightanglelength!(#2)$) in 
              let \p3 = ($(\p1)+(\p2)-(\p0)$) in  
          (\p1) -- (\p3) -- (\p2)
      }
  },
}
\pgfplotsset{
    tick label style={font=\footnotesize},
    axis lines=middle,
    axis x line=center,
    axis y line=center,
    inner axis line style={-latex},
    xlabel={$l_1$},
    xlabel style={anchor=north, align=center},
    ylabel={$l_2$},
    ylabel style={anchor=south, align=center},
    xtick={1},
    ytick={1},
    xtick style={draw=none},
    ytick style={draw=none},
    no markers,
    cycle list={{black,solid}},
    samples=200,
    domain=0.05:0.95,
    axis on top=true,
    xmin=0, xmax=2,
    ymin=0, ymax=2,
    width=0.5\textwidth, height=0.5\textwidth,
    xticklabel style={/pgf/number format/.cd,frac,frac TeX=\varfrac},
    yticklabel style={/pgf/number format/.cd,frac,frac TeX=\varfrac},
}

\newcommand{\cl}{\operatorname{cl}}

\NewDocumentCommand\bnch{G{X}}{{\mathrsfso{#1}}}         
\def\fcone{{X}_{\raisebox{2pt}{$\mkern-1mu\scriptscriptstyle{+}\mkern-1mu$}}^*}      
\def\pcone{{X}_{\raisebox{2pt}{$\mkern-1mu\scriptscriptstyle{+}\mkern-1mu$}}}      
\def\probm{\Delta}                                       
\DeclareMathOperator{\super}{spr}                        
\NewDocumentCommand\prop{}{\mathcaleu{P}}               
\NewDocumentCommand\shdy{}{\mathcaleu{S}}               
\NewDocumentCommand\rdnt{}{\mathcaleu{R}}               
\NewDocumentCommand\cvx{}{\mathcaleu{K}}               
\NewDocumentCommand\rv{m}{\mathsf{#1}}                   

\def\agauge{\mathrm{\beta}\mkern-1mu}    
\def\gauge{\mathrm{\gamma}\mkern-1mu}    
\def\cvsprt{\mathrm{\rho}}               
\def\cxsprt{\mathrm{\sigma}\mkern-2mu}   
\def\apolar{\diamond}                           
\def\polar{\circ}                               
\def\ppolar{\apolar}                            

\NewDocumentCommand{\msum}{s E_{M}}{\IfBooleanT{#1}{\mathbin\bgroup}
    \operatorname{\oplus}_{#2}\IfBooleanT{#1}{\egroup}
}
\NewDocumentCommand{\dualmsum}{s E_{M}}{\IfBooleanT{#1}{\mathbin\bgroup}
    \operatorname{\oplus}^*_{#2}\IfBooleanT{#1}{\egroup}
}

\makeatletter
\NewDocumentCommand{\@mtwrap}{m m m m m m m}{
    \def\tmp{#7}
    \patchcmd{\tmp}{|}{\mathpunct{}\delimsize\vert\mathpunct{}}{}{}
    \def\paireddelim{#6}
    \IfBooleanTF{#1}{
        \paireddelim{\bgroup\mathchoice{\textstyle}{}{}{}\tmp\egroup}
    }{
        \IfBooleanTF{#2}{\paireddelim[\big]{\tmp}}{
        \IfBooleanTF{#3}{\paireddelim[\Big]{\tmp}}{
        \IfBooleanTF{#4}{\paireddelim[\bigg]{\tmp}}{
        \IfBooleanTF{#5}{\paireddelim[\Bigg]{\tmp}}
                    {\paireddelim*{\tmp}}
                }
            }
        }
    }
}

\DeclarePairedDelimiterX{\@rbr}[1]{(}{)}{#1}
\DeclarePairedDelimiterX{\@cbr}[1]{\lbrace}{\rbrace}{#1}
\DeclarePairedDelimiterX{\@sbr}[1]{\lbrack}{\rbrack}{#1}
\DeclarePairedDelimiterX{\restrict}[1]{}{|}{\vphantom{\int}#1}

\NewDocumentCommand\group{s t\big t\Big t\bigg t\Bigg g o d()}{
    \IfNoValueTF{#6}{
        \IfNoValueTF{#7}{
            \IfNoValueTF{#8}{}{
                \@mtwrap{#1}{#2}{#3}{#4}{#5}{\@rbr}{#8}
            }
        }{
            \@mtwrap{#1}{#2}{#3}{#4}{#5}{\@sbr}{#7}
            \IfNoValueF{#8}{(#8)}
        }
    }{
        \@mtwrap{#1}{#2}{#3}{#4}{#5}{\@cbr}{#6}
        \IfNoValueF{#7}{[#7]}\IfNoValueF{#8}{(#8)}
    }
}

\NewDocumentCommand{\cbr}{l m}{\group#1{#2}}
\NewDocumentCommand{\rbr}{l m}{\group#1(#2)}
\NewDocumentCommand{\sbr}{l m}{\group#1[#2]}

\NewDocumentCommand\setqtfy{ >{\SplitList{;}} m}{
    \NewDocumentCommand\qtfier{m}{\itemdelim##1}
    \def\itemdelim{\def\itemdelim{, }} 
    \ProcessList{#1}{\qtfier}:
}
\let\oldsetminus\setminus
\RenewDocumentCommand\setminus{g}{
    \oldsetminus\IfNoValueF{#1}{\set{#1}}
}
\NewDocumentCommand\@set{m m m}{
    \group#1{#2\IfNoValueF{#3}{\mathrel{}\delimsize\vert\mathrel{}#3}}
}
\DeclareDocumentCommand\set{ l >{\SplitArgument{1}{;}} m}{
    \bgroup
        \DeclareDocumentCommand\and{}{,\,\mathpunct{}}
        \DeclareDocumentCommand\or{}{\hat}
        \def\quantify{\setqtfy}
        \def\qtfy{\setqtfy}
        \@set{#1}#2
    \egroup
}

\NewDocumentCommand\scriptparse{m m}{\IfNoValueF{#1}{^{#1}}\IfNoValueF{#2}{_{#2}}}

\NewDocumentCommand\scriptgroup{s e{^_}}{
    \scriptparse#2
    \IfBooleanTF{#1}{\def\@grouped{\group*}}{\def\@grouped{\group}}
    \@grouped
}

\NewDocumentCommand\setscriptgroup{s e{^_} g}{
    \scriptparse#2
    \IfBooleanTF{#1}{\def\@setd{\set*}}{\def\@setd{\set}}
    \IfNoValueF{#3}{\@setd{#3}}
}

\let\g\group

\let\single\cbr

\def\eul{\mathrm{e}}    
\def\diffd{\mathrm{d}}  

\NewDocumentCommand\Lp{E{^}{p}}{\mathrm{\mathcalbd{L}\mkern-1mu}{^{#1}}}  
\NewDocumentCommand\Lq{E{^}{p}}{\mathrm{\mathcalbd{L}\mkern-1mu}{^{#1}}}  
\NewDocumentCommand\lp{E{^}{p}}{\mathrm{\elll\mkern1mu}{^{#1}}}  
\DeclareDocumentCommand\lq{E{^}{p}}{\mathrm{\elll\mkern1mu}{^{#1}}}  

\DeclareDocumentCommand\L{t{1} t{2}}{
    \mathrm[-0.27]{\mathcalbd{L}\mkern-1mu}\IfBooleanT{#1}{^1}\IfBooleanT{#2}{^2}
}  

\DeclareMathOperator\Cont{C}
\newcommand{\ball}{\mathrm{B}}
\makeatother

\DeclareUnicodeCharacter{00B0}{\circ}       
\DeclareUnicodeCharacter{2022}{\bullet}     
\DeclareUnicodeCharacter{00B7}{\cdot}       
\DeclareUnicodeCharacter{2202}{\subdiff}    
\DeclareUnicodeCharacter{2264}{\leq}        
\DeclareUnicodeCharacter{2265}{\geq}        
\DeclareUnicodeCharacter{2260}{\neq}        

\DeclareFontFamily{U}{BOONDOX-calo}{\skewchar\font=45 }
\DeclareFontShape{U}{BOONDOX-calo}{m}{n}{
  <-> s*[1.05] BOONDOX-r-calo}{}
\DeclareFontShape{U}{BOONDOX-calo}{b}{n}{
  <-> s*[1.05] BOONDOX-b-calo}{}
\DeclareMathAlphabet{\mathcalbd}{U}{BOONDOX-calo}{m}{n}
\SetMathAlphabet{\mathcalbd}{bold}{U}{BOONDOX-calo}{b}{n}
\DeclareMathAlphabet\mathcaleu{U}{eus}{m}{n} 
\pdfmapfile{+rsfso.map}
\DeclareMathAlphabet{\mathrsfso}{U}{rsfso}{m}{n}
\DeclareMathAlphabet{\mathrsfs}{U}{rsfs}{m}{n}
\DeclareMathAlphabet{\mathupsf}{\encodingdefault}{\sfdefault}{m}{n}
\DeclareMathAlphabet{\mathsf}{\encodingdefault}{\sfdefault}{m}{sl}
\let\mathscr\mathrsfso

\SetSymbolFont{stmry}{bold}{U}{stmry}{m}{n} 

\newsavebox{\foobox}
\NewDocumentCommand\slantbox{O{0} m}{
    \mbox{%
        \sbox{\foobox}{#2}%
        \hskip\wd\foobox
        \pdfsave
        \pdfsetmatrix{1 0 #1 1}%
        \llap{\usebox{\foobox}}%
        \pdfrestore
    }
}
\NewDocumentCommand\mathup{O{-0.25} m}{\mathchoice
    {\mkern1mu\slantbox[#1]{$\displaystyle #2$}\mkern-1mu}
    {\mkern1mu\slantbox[#1]{$\textstyle #2$}\mkern-1mu}
    {\mkern1mu\slantbox[#1]{$\scriptstyle #2$}\mkern-1mu}
    {\mkern1mu\slantbox[#1]{$\scriptscriptstyle #2$}\mkern-1mu}
}
    
\let\elll\ell
\def\jay{\mathcalbd{j}}
\def\kay{\mathcalbd{k}}
\def\ell{\mathcalbd{l}}
\def\emm{\mathcalbd{m}}

\let\cal\mathcal    \let\scr\mathscr

\NewDocumentCommand\mathrslap{m m}{\ooalign{\hidewidth$#1$\cr$\phantom{#2}$}}
\NewDocumentCommand\mathcslap{m m}{\ooalign{\hidewidth$#1$\hidewidth\cr$\phantom{#2}$}}

\makeatletter
\NewDocumentCommand\missingarg{O{\,} O{\,} m}{\ifblank{#3}{#1\cdot#2}{#3}}
\edef\marg{\missingarg{}}
\NewDocumentCommand{\MTWrapPairedDelimiter}{m m}{
    \NewDocumentCommand#1{s t\big t\Big t\bigg t\Bigg m}{
        \@mtwrap{##1}{##2}{##3}{##4}{##5}{#2}{\missingarg{##6}}
    }
}
\NewDocumentCommand\@inlinecontents{m}{\bgroup\mathchoice{\textstyle}{}{}{} #1\egroup}
\DeclarePairedDelimiterX{\@inner}[1]{\langle}{\rangle}{#1}
\DeclarePairedDelimiterX{\@abs}[1]{\lvert}{\rvert}{\@inlinecontents{#1}}
\DeclarePairedDelimiterX{\@norm}[1]{\lVert}{\rVert}{\@inlinecontents{#1}}
\DeclarePairedDelimiterXPP{\@lnorm}[1]{}{\lVert}{\rVert}{_2}{\@inlinecontents{#1}}
\DeclarePairedDelimiterXPP{\@maxnorm}[1]{}{\lVert}{\rVert}{_\infty}{\@inlinecontents{#1}}
\DeclarePairedDelimiterXPP{\@tvnorm}[1]{}{\lVert}{\rVert}{_\mathrm{TV}}{\@inlinecontents{#1}}
\DeclarePairedDelimiterX{\@floor}[1]{\lfloor}{\rfloor}{\@inlinecontents{#1}}
\DeclarePairedDelimiterX{\@ceil}[1]{\lceil}{\rceil}{\@inlinecontents{#1}}
\DeclarePairedDelimiterX{\@iver}[1]{\llbracket}{\rrbracket}{\@inlinecontents{#1}}

\MTWrapPairedDelimiter{\abs}{\@abs}
\MTWrapPairedDelimiter{\lnorm}{\@lnorm}
\MTWrapPairedDelimiter{\maxnorm}{\@maxnorm}
\MTWrapPairedDelimiter{\tvnorm}{\@tvnorm}
\MTWrapPairedDelimiter{\floor}{\@floor}
\MTWrapPairedDelimiter{\ceil}{\@ceil}
\MTWrapPairedDelimiter{\iver}{\@iver}

\NewDocumentCommand\norm{s t{1} t{2} t{\infty} t\big t\Big t\bigg t\Bigg m}{%
    \@mtwrap{#1}{#5}{#6}{#7}{#8}{\@norm}{\missingarg{#9}}
    \IfBooleanT{#2}{_1}
    \IfBooleanT{#3}{_2}
    \IfBooleanT{#4}{_\infty}
}

\NewDocumentCommand\@innerparse{m m m m m m m m}{
    \def\bra{\missingarg[\mkern2mu][\mkern2mu]{#6}}
    \def\ket{\IfNoValueTF{#7}{
            \missingarg[\mkern2mu][\mkern2mu]{}
        }{
            \missingarg[\mkern2mu][\mkern2mu]{\IfNoValueTF{#8}{#7}{#8}}
        }
    }
    \@mtwrap{#1}{#2}{#3}{#4}{#5}{\@inner}{
        \mathchoice{\textstyle}{}{}{}\bra
        ,
        \mathchoice{\textstyle}{}{}{}\ket
    }
}
\NewDocumentCommand\inner{s t\big t\Big t\bigg t\Bigg >{\SplitArgument{2}{;}} m}{
    \@innerparse{#1}{#2}{#3}{#4}{#5}#6
}
\makeatother

\NewDocumentCommand\varfrac{s m m}{
    \def\sfrac {\raisebox{.5ex}{\(\scriptstyle#2\)}
                \mkern-3mu/\mkern-3mu
                \raisebox{-.5ex}{\(\scriptstyle#3\)}}
    \def\ssfrac{\raisebox{.3ex}{\(\scriptscriptstyle#2\)}
                \scriptstyle{\mkern-3mu/\mkern-3mu}
                \raisebox{-.3ex}{\(\scriptscriptstyle#3\)}}
    \mathinner{
        \IfBooleanTF#1{
            \mathchoice{\sfrac}{\frac{#2}{#3}}{\frac{#2}{#3}}{\frac{#2}{#3}}
        }{
            \mathchoice{\frac{#2}{#3}}{\sfrac}{\ssfrac}{\ssfrac}
        }
    }
}

\NewDocumentCommand\DeclareFraction{m m}{
    \edef#1{\varfrac{1}{#2}}
}

\DeclareFraction\half  {2} \DeclareFraction\third{3} \DeclareFraction\quarter{4}
\DeclareFraction\fifth {5} \DeclareFraction\sixth{6} \DeclareFraction\seventh{7}
\DeclareFraction\eighth{8} \DeclareFraction\ninth{9} \DeclareFraction\tenth  {10}
\NewDocumentCommand\oneon{s s O{1} m}{
    \IfBooleanTF#2{#4^{-1}}{
        \IfBooleanTF#1{\varfrac*{#3}{#4}}{\varfrac{#3}{#4}}
    }
}

\DeclareMathOperator*{\argmin}{arg\,min}
\DeclareMathOperator*{\argmax}{arg\,max}
\DeclareMathOperator*{\arginf}{arg\,inf}
\DeclareMathOperator*{\argsup}{arg\,sup}

\let\integral\int
\def\int{\integral\scriptgroup}

\NewDocumentCommand\minimise{s e_ m g}{
    \IfNoValueTF{#2}{\mathrm{minimise}}{\underset{#2}{\mathrm{minimise}}}\quad #3 
    \IfNoValueF{#4}{\quad\mathrm{subject\ to}\quad#4}
}
\NewDocumentCommand\maximise{m g}{
    \begin{aligned}
        \mathrm{maximise} &&& #1
        \IfNoValueF{#2}{
            \\
            \mathrm{subject\ to} &&&
            \begin{aligned}
                #2
            \end{aligned}
        }
    \end{aligned}
}

\makeatletter
\NewDocumentCommand{\grad}{e{_}}{\nabla\IfNoValueF#1{_{\mkern-3mu #1}}}
\NewDocumentCommand{\hes}{e_}{\nabla^2\IfNoValueF{#1}{_{\mkern-3mu #1}}}

\DeclareDocumentCommand\evaluated{s m}{
    \def\@vsp{\vphantom{\mathchoice{\int}{}{}{}}}
    \left.#2\@vsp\right\vert
}
\DeclareDocumentCommand\eval{}{\evaluated} 

\let\@FirstD\BooleanTrue
\RenewDocumentCommand\d{e{^_} m e{^_} d() t{.} t{,} t{;}}{ 
    \IfBooleanT\@FirstD{\mathinner\bgroup\let\@FirstD\BooleanFalse}
    \diffd{\scriptparse#1}{#2}{\scriptparse#3}
    \IfNoValueF{#4}{(#4)}%
    \@ifnextchar\d{
        \mkern1mu
    }{%
        \let\@FirstD\BooleanTrue%
        \egroup
    }%
    \IfBooleanT#5{\mathclose{}.}\IfBooleanT#6{\mathclose{},}\IfBooleanT#7{\mathclose{};}
}

\def\d{\diffd}

\DeclareDocumentCommand\dv{s e{^_} m g}{
    \edef\num{\IfNoValueF{#4}{#3}}
    \edef\den{\IfNoValueTF{#4}{#3}{#4}}
    \IfBooleanTF#1{\varfrac*}{\varfrac}{\d{\scriptparse#2}\num}{{\d\den}{\scriptparse#2}}%
    \scriptgroup
}

\newcommand{\partialderivative}{\partial}

\DeclareDocumentCommand\pdv{}{\partialderivative} 

\newcommand{\ddef}[1]{\expandafter\def\csname d#1\endcsname{\d #1}}
\makeatother

\NewDocumentCommand\smplx{}{\Delta}

\newcommand{\subseq}{\subseteq}
\newcommand{\supseq}{\supseteq}

\DeclareMathOperator{\bd}{bd}
\DeclareMathOperator{\interior}{int}     

\DeclareMathOperator{\conv}{co}          
\let\co\conv
\DeclareMathOperator{\relbd}{rbd}        
\DeclareMathOperator{\cone}{cone}        
\DeclareMathOperator{\rec}{rec}          
\NewDocumentCommand{\harmsum}{}          
    {\mathbin{\#}}

\DeclareMathOperator{\lev}{lev}          
\DeclareMathOperator{\dom}{dom}          
\DeclareMathOperator{\relint}{ri}        
\DeclareMathOperator{\epi}{epi}          
\DeclareMathOperator{\hyp}{hyp}          

\NewDocumentCommand{\moreau}{m m}        
    {\prescript{#2}{}{#1}}

\def\infconv{\mathbin{\square}}          
\DeclareMathOperator{\breg}{B}           

\DeclareMathOperator{\subdiff}{\partial}        

\newcommand{\union}{\bigcup}
\newcommand{\inter}{\bigcap}

\def\logloss{\ell_{\log}}                       
\def\brierloss{\ell_{\scriptscriptstyle\mathrm{Br}}}                
\def\ooneloss{\ell_{\scriptscriptstyle\mathrm{0/1}}}                
\NewDocumentCommand\cdloss{s E_{a}}{\IfBooleanTF#1{\bar\ell}{\ell}_{\scriptscriptstyle\mathrm{CD}_{#2}\!}}                
\NewDocumentCommand\normloss{E_{\alpha}}{\ell_{\scriptscriptstyle\norm{\cdot}_{#1}\!}}                 

\NewDocumentCommand\ppt{s}{\IfBooleanTF{#1}{\mathrel{\psi}}{\psi}} 
\NewDocumentCommand\elicitloss{E_{s} E_{\ppt}}{\ell_{{#1},{#2}}} 

\newcommand{\E}{\mathbb{E}}
\newcommand{\R}{\mathbb{R}}
\newcommand{\reals}{\mathbb{R}}
\newcommand{\Rx}{\overline{\mathbb{R}}}
\newcommand{\Rp}{\mathbb{R}_{\ge 0}}
\newcommand{\Rpp}{\mathbb{R}_{> 0}}
\newcommand{\Rm}{\mathbb{R}_{\le 0}}
\newcommand{\Rmm}{\mathbb{R}_{< 0}}
\newcommand{\N}{\mathbb{N}}

\newcommand{\dir}{\operatorname{dir}}
\def\Ds{\mathsf{D}}

\def\Fcal{{\cal{F}}}

\newcommand{\hzn}{\operatorname{hzn}}
\def\Hs{\mathsf{H}}

\def\minL{\underline{L\mkern-2mu}\mkern2mu}     
\def\minM{\underline{M\mkern-4mu}\mkern4mu}     
\def\minN{\underline{N\mkern-4mu}\mkern4mu}     

\newcommand{\oneb}{{\boldsymbol{1}}}

\newcommand{\Pcal}{{\mathcal{P}}}

\newcommand{\st}{\,|\,}

\def\infconv{\mathrel{\square}}

\newcommand{\llangle}{\smash{\langle}\hspace*{-0.8mm}|}
\newcommand{\rrangle}{|\hspace*{-0.8mm}\smash{\rangle}}

\newcommand{\Tfrac}[2]{%
  \ooalign{%
    $\genfrac{}{}{1.2pt}1{#1}{#2}$\cr%
    $\color{white}\genfrac{}{}{.4pt}1{\phantom{#1}}{\phantom{#2}}$}%
}
\newcommand{\Dfrac}[2]{%
  \ooalign{%
    $\genfrac{}{}{1.2pt}0{#1}{#2}$\cr%
    $\color{white}\genfrac{}{}{.4pt}0{\phantom{#1}}{\phantom{#2}}$}%
}
\allowdisplaybreaks  

\title{The Geometry and Calculus of Losses}
\jmlrheading{?}{2023}{$n$--$(n+l-1)$}{8/23}{?}{Williamson, Cranko} 
\ShortHeadings{Geometry and Calculus of Losses}{Williamson, Cranko}
\author{%
    Robert C. Williamson \email{Bob.Williamson@uni-tuebingen.de}\\
    \addr{University of T\"{u}bingen and T\"{u}bingen AI Center,\\
	  Germany} 
    \AND
    Zac Cranko \email{Zac.Cranko@gmail.com}\\
    \addr{Sydney, Australia }
}
\editor{?}
\begin{document}

\maketitle
\begin{abstract}
	Statistical decision problems lie at the heart of statistical
	machine learning. The simplest problems are binary and multiclass
	classification and class probability estimation. Central to their
	definition is the choice of loss function, which is the means by
	which the quality of a solution is evaluated.  In this paper we
	systematically develop the theory of loss functions for such
	problems from a novel perspective whose basic ingredients are
	convex sets with a particular structure.  The loss function is
	defined as the subgradient of the support function of the convex
	set. It is consequently automatically proper (calibrated for
	probability estimation).  This perspective provides three novel
	opportunities. It enables the  development of a fundamental
	relationship between losses and (anti)-norms that appears to have
	not been noticed before.  Second, it enables the development of a
	calculus of losses induced by the calculus of convex sets which
	allows the interpolation between different losses, and thus is a
	potential useful design tool for tailoring losses to particular
	problems. In doing this we build upon, and considerably extend,
	existing results on $M$-sums of convex sets. Third, the perspective
	leads to a natural theory of ``polar''  loss
	functions, which are derived from the polar dual of the convex set
	defining the loss, and which form a natural universal substitution
	function for Vovk's aggregating algorithm.
\end{abstract}

\begin{keywords}
	convex sets, support functions, gauges, polars, concave duality,
	proper loss functions, $M$-sums, distorted probabilities, polar
	losses, Shephard duality, anti-norms, Bregman divergences, semi
	inner products, Finsler geometry, 
	aggregating algorithm, substitution functions, direct and inverse
	addition.
\end{keywords}

\section{Introduction} 
\label{sec:introduction}

Most machine learning research focusses on methods (algorithms). But these
methods are designed to solve particular problems.  \citet{Platt1962}
argued for the greater importance of problem-oriented research.  Our
premise is that we need to better understand the elements of machine
learning problems, and their permissible transformations.  We focus on some
of the simplest possible machine learning problems, namely multiclass
classification and probability estimation. 

Stateless machine learning problems have three key ingredients: 
\begin{enumerate}[itemsep=0pt]
	\item the \emph{loss function} $\ell$: specifies how predictive 
		performance is evaluated; 
	\item the \emph{data generating process}: in the statistical 
		setting corresponds to an underlying probability distribution 
		$P$ from which samples are drawn;
	\item the \emph{model class} $\Fcal$:  the analyst's choice,
		 informed by their prior 
		 knowledge\footnote{The 
			claim that all the analyst brings is the model
			class $\Fcal$ is a simplification that captures
			much of ML; in general the analyst provides a
			learning ``algorithm'' (a function) 
			$A\colon \mathsf{S}\mapsto f\in
			\Fcal$ which given a sample $\mathsf{S}$ produces
			an $f$ \citep{Herbrich:2002aa}. For the purposes of
			the present paper the simplification stated in the main
			text suffices. }.
\end{enumerate}
Implicit in this is the protocol by which the learner or analyst interacts
with the data; we presume the usual statistical batch setting for now, but
most of the technical results of the paper are not so restricted. Thus an
(idealised) machine learning problem can be parametrised by the triple
$(\ell,\Fcal, P)$.

Much research in machine learning focusses upon the classes
of models $\Fcal$ and methods for searching for the best element within
$\Fcal$ for data generated by $P$, or on theoretical results concerning the
complexity of $\Fcal$ and its effect on convergence of empirical estimates
\citep{Vapnik1998}, or the intrinsic geometry induced by $\Fcal$
\citep{Amari:2016aa}.  Little attention has been paid to research on the
loss function $\ell$, and its interaction with the other ingredients
$\Fcal$ and $P$.  A recent exception is \citep{van-Erven:2015aa} which
showed how the \emph{joint} interaction of $\ell$, $\Fcal$ and $P$ control
the speed of convergence of learning algorithms.
The lack of attention is surprising because the choice of loss function
matters, especially when (as is typical) one has limited data,
and so one cares about the speed of convergence of empirical estimates, and
the best model in the class $\Fcal$ has non-zero expected loss (again
typically the case).  Understanding the implications and options for the
choice of loss function also matters when one considers the integration of
machine learning technologies into larger socio-technical systems, since
the loss function serves as an abstraction of what matters at the larger
system level, and can be used, for example, to abstract a range of notions
of fairness in ML problems \citep{Menon2018}.

Loss functions are central to statistical decision problems, and have a
long history \citep{Wald1950}; see \citep{Vernet:2016aa,Williamson2013} for
some pointers to the literature.  The present paper focusses upon
understanding at a deeper level the loss functions for multiclass
probability estimation, and their possible transformations.  Our results
are, to use the apposite term of \citet{Rota:1997aa}, ``cryptomorphic'' ---
an isomorphism that was previously hidden from view, which once decoded is
illuminating.  Our approach is parametrisation independent in the sense of
the distinction made in \citep{van-Erven:2015aa,Vernet:2016aa} (in essence,
what matters is the geometry of the set induced by the loss function which
does not change under reparametrization). 

Losses in machine learning play a role analogous to metrics in other
applied problems.  \citet{Menger1928} introduced distance geometry (in
order to view the world in terms of distances) and there is an incredible
variety of distances to choose from \citep{Deza:2009aa}.  But as we shall
see, it is the simpler notion of norms, and normed spaces, that are the
most relevant in the study of losses.  The development of functional
analysis critically depended upon the development of finite dimensional
normed spaces by \citet{Minkowski:1896aa}, who, in his \emph{Geometrie der
Zahlen}, developed the notion of a symmetric convex body and its
equivalence to a norm ball $\set{x;p(x)\le 1}$, as well as introducing the
notion of a supporting hyperplane and the corresponding
support function and showed\footnote{See \citep[Section 2]{Martini:2001aa} and
	\citep[Section 1.5]{Thompson1996} for a more detailed  history.
	The extension of these concepts to infinite dimensional spaces
	underpinned the development of functional analysis; \citet[page
	130]{Dieudonne1981} credits \citet{Helly:1921aa} with the idea of
	abstracting away from \emph{particular} spaces such as $\lp$, $\Lp$
	or $\Cont([a,b])$ to the notion of \emph{general} normed sequence
	spaces by methods which do not depend upon special features of the
	space. While apparently rather elementary, these  finite
	dimensional normed spaces (``Minkowski Spaces''
	\citep{Thompson1996}) underpin the general theory of Banach spaces.
	\citet[Page xxii]{Pietsch:2007aa} quotes Dvoretzky (1960) inspired
	by Grothendieck: ``many problems in the theory of Banach spaces may
	be reduced to the finite-dimensional case, i.e. to problems concerning
	Minkowski spaces.''  }  
it was the dual to the norm $p(x)$.  
We shall see that these concepts that were central to the development
of normed spaces are, with minor modification, the foundation for an
understanding of loss functions.  Recapitulating history, we concentrate in
this paper on the finite dimensional case, corresponding to multiclass
classification and conditional probability estimation problems.

The rest of the paper is organised as follows. Section
\ref{sec:preliminaries} introduces the mathematical machinery we utilise.
Section \ref{sec:loss_functions} introduces (proper) loss functions,
including the antipolar loss which is a natural ``inverse'' of a loss
function.  Section \ref{sec:polar_applications} presents examples,
illustrating the new perspective, and the antipolar loss in particular.
Section \ref{sec:designing_losses_via_their_superprediction_sets} presents
some design strategies for loss functions in terms of their superprediction
sets.  Section \ref{sec:new_losses_from_old} shows how to construct new
proper losses from old ones by suitable combination of their
superprediction sets. These results are based on the new results in Section
\ref{sub:the_m-sum_polar} which substantially extend the theory of $M$-sums
of convex sets, including a general duality result for $M$-sums of norms
and anti-norms.  Section \ref{sec:conclusion} concludes.  

\subsection{Motivation, Expectations, Context and Significance}
The goals and results of this paper are different in nature to those of the
majority of papers in machine learning\footnote{But not different in nature
	to many papers in economics.
	Indeed, economists have, over a long
	period, conducted investigations on the foundations of their
	discipline (utilities). 
	As we shall see below in footnote
	\ref{footnote:economic-duality}, the similarity turns out not to be
	just in style, but there is a remarkable parallel in content as
	well.}. To that end, we give
some context and set expectations. The paper contains no new algorithms and
no experimental results. What it does contain is a new way of
looking at loss functions which 1) illustrates the close connection between
losses and norms and anti-norms; 2) presents the new idea of an antipolar
loss; 3) develops a
calculus for loss functions that allows multiple proper loss
functions to be combined in a manner that the resulting loss is guaranteed
proper; 4) shows how the geometrical perspective can be used to design loss
functions.  

Why embark on this complex endeavour? Currently loss functions are widely
used, but there is little insight to be had regarding the
consequences of particular choices. This is especially true when these
functions are identified with their algebraic formulas.  There were
insights derived in \citep{Reid2011}  for the design of loss functions
(following \cite{HandVinciotti2003}), and in \citep[Appendix
B]{Menon:2018aa}, but these approaches, whilst tractable enough in the
binary case, become  intractable for the multiclass situation.  As we
will show, there is an intrinsic geometry to loss functions which controls
the nature of the learning problem at hand. There is already evidence for
this in \citep{van-Erven:2012vz,Cabrera:2023aa} which showed how the mixability constant
of a loss (which directly appears in bounds for the regret in online
learning) is directly controlled by the intrinsic geometry of the loss
function.

Some of the value of the viewpoint developed in the paper is
only realised in the companion paper \citep{Williamson:2022aa} which uses
the geometric approach developed here to derive, in a much
simpler manner, the bridge between loss functions and measures of
information that was previously presented in \citep{Reid2011} (binary case)
and \citep{Garcia-Garcia2012} (multiclass case).  In
\citep{Williamson:2022aa} we show that the geometric way of viewing
information measures allows one to derive results seemingly unobtainable by
others means. In particular, we derive a  general data processing
\emph{equality} from which one can derive the classical strong data
processing inequality. It turns out that the geometric viewpoint is central
to these novel results.

The new perspective has  been used by \citet{Kamalaruban:2015aa} to
show the connection between exp-concavity and mixability, which is relevant
to online learning algorithms, as well as to the understanding of fast
rates in statistical learning \citep{van-Erven:2015aa}. It was  used by
\citet{Mhammedi:2018aa} to solve an open question regarding  generalised
mixability as well as to draw a connection to mirror descent.  It also
underpins the results of  \citet{Cranko:2021aa}, which develops the theory
of proper composite losses in an infinite dimensional setting.  The present
paper is a substantially extended and improved version of
\citep{williamson2014geometry}.  Beyond the correction of some errors, the
present paper fully develops the theory of $M$-sums of superprediction sets
in a general and rigorous manner.

\section{Preliminaries} 
\label{sec:preliminaries}


We introduce some standard machinery from the  theory of convex sets and
functions \citep[see][]{hiriarturruty2001fca,Rockafellar:1970,RockafellarWets2004,
        Schneider2014,Penot1997Duality}\footnote{We recognise that there is
	a significant quantity of background material needed before we get
	to the machine learning problem and the results about loss
	functions.  But this really illustrates the point of the paper: all
	of the deeper structure of loss functions arises from more
	fundamental geometrical concepts. And while some of the material in
	this section is widely known, the results for concave gauges and
	their polar duals, which are
	central to the analysis of loss functions, are both less well known and
	\emph{not} a trivial variation of the convex case.}.
The concave cases of some of these results can be found in the works of  
\citet{Pukelsheim1983} and \citet{Barbara1994}.  In choosing our notational
conventions, we have adopted notation more common in the mathematical
literature, even though some of this may be unfamiliar to a machine
learning audience (since we refer to the mathematical literature for a
number of the results upon which we build).

\subsection{Basic Notation} Let $\Def{\bnch}$ be a finite dimensional
Euclidean space over the reals. The space of linear functionals on $\bnch$
is $\Def{\bnch^*}$, and the natural coupling is
$\Def{\inner{}}\colon\bnch^*\times\bnch\to\R$; the usual inner product.
Define the special sets $\Def{\Rpp}{(0,\infty)}$; $\Def{\Rp }{
[0,\infty)}$; $\Def{\Rmm }{ (-\infty,0)}$; $\Def{\Rm }{ (-\infty,0]}$;
$\Def{\Rx}{[-\infty,+\infty]}$; $\Def{\Rx_{\ge 0}}{[0,+\infty]}$.
Denote the cardinality of a set
$S\subset\bnch$ by $\Def{\abs{S}}$.  If $S=\{T_\alpha\st \alpha\in A\}$ is
a set of sets, then $\Def{\bigcap S}{\bigcap_{\alpha\in A} T_\alpha}$ and 
$\Def{\bigcup S}{\bigcup_{\alpha\in A} T_\alpha}$.
We refer to the components of $x\in\bnch$ by $x_i$ and $x = (x_1,\dots,x_n)$. 
Let $\Def{\smplx(S)}$ denote the set of probability measures on a set $S$,
and  $\Def{[n]}{\set{1,2,\dots,n}}$; then 
$\smplx([n])\simeq\set{x\in\Rp^n; \sum_{i=1}^n x_i = 1}$. 
Let $\Def{(\eul^i)_{i\in[n]}}$ be the canonical basis vectors in $\R^n$. 
The family of $p$-norms (with $p\in[1,\infty]$) on the space $\bnch$ are
defined by $\Def{\norm{x}_p}{\g\big(\sum_{i\in[n]} \abs{x_i}^p )^{1/p}}$
for finite $p$, and $\Def{\norm\infty{x}}{\max_{i\in[n]}
\abs{x_i}}=:\bigvee_{i\in [n]} x_i$. The
\Def{$p$-unit ball} is $\Def{\ball_p}{\set*{x\in\bnch; \norm{x}_p≤1}}$, 
and if there is no subscript we take $\Def{\ball}{\ball_2}$.
The Iverson bracket $\Def{\iver{}}$ takes a proposition and returns 1 if it
is true, and 0 otherwise. We use the common conventions
$\Def{\inf(\varnothing)\coloneqq +\infty}$, 
$\Def{\sup(\varnothing)\coloneqq -\infty}$, 
$\Def{\infty \cdot 0}{0}$ and
$\Def{1/0}{\infty}$.
If $v\in\R^n$,  then $v'$ denotes its transpose. The all ones vector is
defined as
$\Def{\oneb_n}{(1,\ldots,1)'}\in\R^n$.

\subsection{Convex Sets}
Let $S,T\subseteq\bnch$, $x\in\bnch$, $\alpha\in\R$ and $U\subset\R$.  Let
$\Def{\alpha S}{\set{\alpha s; s\in S}}$, $\Def{U\!\cdot\! S}{\set{\alpha
s;\alpha\in U,\ s\in S}}$, $\Def{S+x}{\set{s+x; s\in S}}$. 
The \Def{Minkowski sum} is $\Def{S+T}{\set{s+t; s\in S\mbox{\ and\ } t\in T}}$. 
For $\varnothing\subset S\subseteq\bnch$, $\Def{\cl(S)}$ and
$\Def{\overline{S}}$ both denote its \Def{closure}. The collection of
\Def{closed, nonempty, convex subsets of $S$} is $\Def{\cvx(S)}$. The
\Def{interior} and \Def{boundary} of $S$ are 
\[
    \Def{\interior (S)}{\{x \in S \st \exists{\epsilon>0},\ 
    (\epsilon\ball + x)\subseq S\}} \mbox{\ \ and\ \  }
    \Def{\bd (S)}{S \setminus \interior(S)}.
\]
If $S$ is convex its \Def{relative interior} and \Def{relative boundary} are
\begin{gather}
    \Def{\relint(S)}{\{x\in S \st \forall{y\in S},\ \exists{\lambda>1},\  \lambda
    x + (1-\lambda) y \in S\}} \mbox{\ \ and\ \ }
    \Def{\relbd(S)}{S\setminus\relint(S)}.\label{eq:relint_defn}
\end{gather}
Its \Def{convex hull} is 
$\Def{\co(S)}{\bigcap \{ T\subseteq\bnch \st S\subseteq T 
	\text{\ and $T$ convex}\}}$; 
	its \Def{conic hull} is
$\Def{\cone(S)}{\bigcup_{t>0} t S}$. For the closure of these
operations we sometimes write $\Def{\cl\co(S)}{\cl (\co S)}$, and
$\Def{\cl\cone(S)}{\cl (\cone S)}$.

%

\subsection{Starry, Radiant and Shady Sets}
\label{sec:starry}
A nonempty, proper subset $S\subset\bnch$ is: 
\begin{itemize} 
	\setlength{\itemsep}{0mm}
	\item \Def{star-shaped} if $(0,1]\cdot S\subseteq S$ and $0\in\interior S$;
	\item \Def{co-star-shaped} if $[1,\infty)\cdot S\subseteq S$ and $0\notin S$;
	\item \Def{radiant} if it is star-shaped and convex; 
	\item \Def{shady} if it is co-star-shaped and convex. 
\end{itemize}
By convention the empty set is star-shaped (and radiant),
and the entire space is co-star-shaped. Thus the star-shaped sets
are the complements of the co-star-shaped sets and vice versa. If
$-S=S$ we say $S$ is \Def{symmetric}; if $S$ is symmetric and
radiant we say it is a \Def{norm ball}. Let $\Def{\rdnt(X)}$ and
$\Def{\shdy(X)}$ denote, respectively, the collections of \Def{closed
radiant} and \Def{closed shady subsets} of $X\subseteq\bnch$. These
definitions are illustrated in
Figure \ref{fig:relationships_between_sets_gauges}.

\begin{figure}
    \centering
    \tikzsetnextfilename{relationships_between_sets_gauges}
    \begin{tikzpicture}
        \node [rounded] (cvx) {convex};
        \node [rounded, above left  = 3em of cvx] (rad) {radiant};
        \node [rounded, above right = 3em of cvx] (shd) {shady};
        \node [rounded, below left  = 3em of rad] (str) {star-shaped};
        \node [rounded, below right = 3em of shd] (css) {co-star-shaped};
        \node [rounded, above = 8em of str.west, anchor = west] (bal) {norm ball};
        \node [rounded, below left  = 3em of bal] (sym) {symmetric};
        \path (cvx) edge[bend left, -latex]  (rad);
        \path (str) edge[bend right, -latex]  (rad);
        \path (rad) edge[bend left, -latex]  (bal);
        \path (sym) edge[bend right, -latex]  (bal);
        \path (cvx) edge[bend right, -latex]  (shd);
        \path (css) edge[bend left, -latex]  (shd);

        \node[draw=none] (poshomnw) at (str.south west  |- bal.north west) {};
        \node[draw=none] (poshomse) at (str.south west -| rad.south east) {};
        \node[draw=none] (sublinearse) at (bal.south east |- rad.south east) {};

        \draw[dashrounded] ($(poshomnw)+(-0.3,0.3)$) rectangle ($(poshomse)+(0.3,-0.6)$)
            node[faint,anchor=south east,draw=none] {\textit{$1$-homogeneous}};;
        \draw[dashrounded] ($(poshomnw)+(-0.2,0.2)$) rectangle ($(rad.south east)+(0.2,-0.2)$) node[draw = none, pos=0.5] {} ;
        \draw[dashrounded] ($(poshomnw)+(-0.1,0.1)$) rectangle ($(sublinearse)+(0.1,-0.1)$);
        \node[draw=none,faint,anchor=north east] at 
            ($(rad.north east |- poshomnw) + (+0.2,+0.2)$) {\textit{subadditive}} ;
        \node[draw=none,faint,anchor=south] at 
            ($(rad.south east -| bal)-(0,0.1)$) {\textit{norm}} ;
        \node[draw=none,anchor=south,faint] at 
            ($(rad.north east |- poshomnw)!0.5!(poshomnw) + (0,0.4)$) {\large$\textcolor{gray}{\gauge_{S}}$};
                
        \node[draw=none] (poshomne) at (css.south east |- poshomnw) {};
        \node[draw=none] (poshomsw) at  (css.south east -| shd.south west) {};
        
        \draw[dashrounded] ($(poshomne)+(0.3,0.3)$) rectangle ($(poshomsw)+(-0.3,-0.6)$) 
            node[faint, anchor=south west,draw=none] {\textit{$1$-homogeneous}};
        \draw[dashrounded] ($(poshomne)+(0.2,0.2)$) rectangle ($(shd.south west)+(-0.2,-0.2)$);
        \node[draw=none,faint,anchor=north west] at 
            ($(shd.north west |- poshomnw) + (-0.2,+0.2)$) 
            {\textit{superadditive}};
        \node[draw=none,anchor=south,faint] at 
            ($(shd.north west |- poshomne)!0.5!(poshomne) + (0,0.4)$) 
            {\large$\textcolor{gray}{\agauge_{S}}$};
    \end{tikzpicture}
    \caption{The relationship between various classes of sets
	    $S\subseq\bnch$ defined in \S\ref{sec:starry}, and the 
	    corresponding properties of the
	    gauge $\gauge_S$ and antigauge $\agauge_S$ defined in 
	    \S\ref{sec:gauge_functions_and_polar_duality}.}
    \label{fig:relationships_between_sets_gauges}
\end{figure}

\subsection{Cones and Recession Cones}
\label{sec:cones}
A set $C\subseteq \bnch$ is said to be a \Def{cone} if
$\Rpp·C\subseq C$. A cone $C$ is \Def{pointed} if $0\in C$; \Def{salient}
if $x,-x\in C$ implies $x=0$; and \Def{blunt} if  $0\notin C$. 
Every closed cone is pointed. 
Every blunt, convex cone is salient, but this is not the case for pointed
convex cones. If a convex cone $C$ is salient, then $C\setminus{0}$ is also
a convex cone.
For a cone $C\subseq\bnch$ there is a natural counterpart
$C^*\subseq\bnch^*$ called the \Def{dual cone}, where
\begin{gather}
    \Def{C^*}{\set{ x^*\in\bnch^*; \forall{x\in C},\ \inner{x^*;x} ≥ 0}}.
    \label{eq:dual_cone_def}
\end{gather}
A pointed convex cone $C\subseq\bnch$ induces a partial ordering on $\bnch$
which we denote $\Def{\succeq_C}$; for $x,y\in\bnch$ we say $x\succeq_C y$
if $x-y\in C$. For a set $S\subseq \bnch$ we say $d\in \bnch$ is a
\Def{recession direction of $S$} if $S+d\subseq S$. The collection of
recession directions of $S$ is called its \Def{recession cone} which we
denote by
\begin{gather}
    \Def{\rec(S)}{\set{d\in\bnch; S + d \subseq S}}.\label{eq:rec_cone_defn}
\end{gather}
The recession cone is illustrated in Figure \ref{fig:recession_cone}.
If $S\in\cvx(\bnch)$ then it is easily verified that $\rec(S)$ is indeed a
closed convex cone. We say a set $S\subseq \bnch$ is \Def{$C$-oriented} if
$\rec(S)=C$. Unless otherwise stated we use $\Def{\pcone}$ to denote a
salient, closed, convex cone that is a proper subset of $\bnch$: $
\pcone\subset \bnch$. 

\begin{figure}
    \centering
    \tikzsetnextfilename{recession_cone}
    \begin{tikzpicture}
        \begin{axis}[domain=0.05:0.95, xmin=0, xmax=2, ymin=0, ymax=2, width=0.4\textwidth, height=0.4\textwidth, xtick={0}, ytick={0}, xlabel={$x_1$}, ylabel={$x_2$}]
            \coordinate (o) at (axis cs: 0,0);
            \path[name path=upper_axis] (axis cs:0.35,5) -- (axis cs:5,5) -- (axis cs:5,0.35);
            
            \draw[set, Green, name path=rec] (5,0) -- (o) -- (0,5);
            \addplot[fill=LightGreen] fill between[of=rec and upper_axis];

            \draw[set, Blue, name path=left_s] 
                (axis cs:0.35,1.35) -- (axis cs:0.35,3);
            \draw[set, Blue, name path=right_s] 
                (axis cs:1.35,0.35) -- (axis cs:3,0.35);
            \addplot[set, Blue, name path=round_s, domain=pi:1.5*pi,samples=300]
                ({cos(deg(x)) + 1.35},{sin(deg(x)) + 1.35});
            \addplot[fill=LightBlue] fill between[of=round_s and upper_axis];
            \addplot[fill=LightBlue] fill between[of=right_s and upper_axis];

            \coordinate (point) at (axis cs:0.6,0.9);
            \coordinate (dir) at ($(point) + (axis cs:0.2,0.5)$);

            \draw[set, color=Red, name path=rec_cone] 
                (point |- {{(3,3)}}) -- (point) -- (point -| {{(3,3)}}) ;
                
            \addplot[fill=LightRed] fill between[of=rec_cone and upper_axis];
            
            \draw (axis cs:2,0.9) node[anchor=south east, color=Red] {$\rec(S) + x$};
            \draw (axis cs:2,0.35) node[anchor=south east, color=Blue] {$S$};
            \draw (axis cs:2,0) node[anchor=south east, color=Green] {$\rec(S)$};

            \draw (point) node[dot] {} node[anchor=south east] {$x$} ;
            \draw (dir) node[dot] {} node[anchor=south west] {$x + d\in S$} ;
            \draw ($(dir) - (point)$) node[dot] {} node[anchor=south west] {$d$} ;
            \draw[shorten >=0cm, shorten >=1.5pt, -latex] (point) -- (dir);
            \draw[shorten >=0cm, shorten >=1.5pt, -latex] (o) -- ($(dir) - (point)$);
        \end{axis}
        \draw (o) node[anchor = north east] {$0$};
        \draw (o) node[anchor = north east] {$0$};
    \end{tikzpicture}
    \caption{
	An illustration of the set $S\subseq\bnch$ (which extends
	infinitely far to the ``north-east'', and thus only a finite
	portion is illustrated, a convention which we adopt throughout the
	paper) and its recession cone
	$\rec(S)=\Rp^2$. For all $x\in S$ and all $d\in\rec(S)$ the point
	$x+d$ is contained within $S$. 
    }
    \label{fig:recession_cone}
\end{figure}

\begin{proposition}[Calculus of Recession Cones]\label{prop:rccalc}
    Let $S\in\cvx(\bnch)$. The following hold:
    \begin{enumerate}
        \item $\rec(S) = \single{0}$ if and only if $S$ is bounded;
		 \label{prop:rccalc_bounded}
        \item $\rec(S)$ is a closed convex cone, thus $S + \rec(S) = S$; 
		\label{prop:rccalc_cvx}
	\item $d\in\rec(S)$ if and only if there exist sequences
		$(x_n)_{n\in\N}$, $x_n\in S$, and $(t_n)_{n\in\N}\searrow
		0$, $t_n\in\Rp$, such that $x_n t_n \to d$;
		\label{prop:rccalc_seq}
	\item if $C\subseq\bnch$ is a cone then $\rec(C) = \cl(C)$;
		\label{prop:rccalc_cone}
	\item if $A \subseq B\subseq \bnch$ then $\rec(A) \subseq \rec(B)$.
		\label{prop:rccalc_subset}
    \end{enumerate}
    If $(S_i)_{i\in I}$ with $S_i\in\cvx(\bnch)$ is a family with an
    arbitrary index set $I$ with $\inter_{i\in I}S_i\neq \varnothing$, then
    \begin{enumerate}[resume]
	\item $\rec(\inter_{i\in I} S_i) = \inter_{i\in I} \rec S_i$;
		\label{prop:rccalc_inter}
	\item $\rec(\union_{i\in I} S_i)\supseq \union_{i\in I} \rec S_i$.
		\label{prop:rccalc_union}
    \end{enumerate}
    If $[m]\subseteq I$ is a finite subcollection, then
    \begin{enumerate}[resume]
	\item $\rec\big( \sum_{i\in[m]} S_i) \supseq  \sum_{i\in[m]}
		\rec(S_i)$. \label{prop:rccalc_sum}
    \end{enumerate}
\end{proposition}
\begin{proof}
    These are all well-known and can be found in a variety of common
    references \citep{Auslender2003,Rockafellar:1970,RockafellarWets2004}.
\end{proof}

\subsection{Convex, Concave and Homogeneous Functions}
For the remainder of this section let $f\colon\bnch\to\Rx$; we define its
\Def{domain}, \Def{epigraph} and \Def{hypograph} respectively as 
\begin{align}
    \Def{\dom (f)} &\coloneqq {\set{ x \in\bnch ; f(x) \in\R}},\\
    \Def{\epi (f)} &\coloneqq {\set{(x,t) \in \dom f\times\R ; f(x) ≤ t}},\\
    \Def{\hyp (f)} &\coloneqq {\set{(x,t) \in \dom f\times\R ; f(x) ≥ t}}.
\end{align}
Let $\alpha\in\R$. The \Def{below}, \Def{level} and \Def{above sets} are,
respectively, 
\begin{gather}
    \Def{\lev_{\le\alpha}(f)}{\set{x\in\dom(f); f(x)\le \alpha}},\\
    \Def{\lev_{=\alpha}  (f)}{\set{x\in\dom(f); f(x)= \alpha}},\\
    \Def{\lev_{\ge\alpha}(f)}{\set{x\in\dom(f); f(x)\ge \alpha}}.
\end{gather}

We say $f$ is \Def{convex} if it satisfies
\begin{gather}
    \forall{x,y\in\dom f}\ \forall t\in(0,1),\  f(t x + (1-t)y) ≤ tf(x) + (1-t)f(y).
\end{gather}
If $-f$ is convex, then $f$ is \Def{concave}. Equivalently $f$ is convex
(resp.\ concave) if $\epi (f)$ is convex (resp.\ $\hyp (f)$ is convex).
We say $f$ is \Def{quasi-convex} (resp.~\Def{quasi-concave}) 
if $\lev_{\le\alpha} f$  (resp.~$\lev_{\ge \alpha} f$)
is convex for all $\alpha\in\R$.  We
say a convex (resp.\ concave) function $f$ is \Def{closed} if $\epi (f)$
(resp.\ $\hyp (f)$) is closed. Thus if $f$ is closed and convex, $-f$ is
closed and concave. The \Def{closure} of a convex (resp.\ concave) function
$f$ is the convex (resp.\ concave) function $g$ such that $\epi(g) =
\cl(\epi f)$ (resp.\ $\hyp(g) = \cl(\hyp f)$). The function $g$ is denoted
by $\Def{\cl(f)}$. Finally for a set $S\subseq\bnch$, define
$\Def{\argsup_{x\in S} f(x)}{\{x\in\bnch\st f(x)=\sup_{x\in S} f(x)\}}$ and
$\Def{\arginf_{x\in S} f(x)}{\{x\in\bnch\st f(x)=\inf_{x\in S} f(x)\}}$;
either of these sets can be empty.

If for some $k\in\R$, $f$ satisfies $f(\alpha x) = \alpha^k f(x)$ for 
$\alpha>0$ and for all $x$, we say $f$ is
\Def{homogeneous of degree $k$} or \Def{$k$-homogeneous} (there is an
obvious extension to set-valued functions). If $f$ is
1-homogeneous, we also say $f$ is \Def{positively homogeneous}. If for all
$x,y\in\dom (f)$ we have $f(x+y)≤f(x) + f(y)$ then $f$ is called
\Def{subadditive}. If $-f$ is subadditive then $f$ is called
\Def{superadditive}. If $f$ is positively homogeneous and subadditive
(resp.\ superadditive) then we say $f$ is \Def{sublinear} (resp.
\Def{superlinear}). All sublinear functions are convex and all superlinear
functions are concave.  Suppose $f_1,\ldots,f_m\colon \bnch\rightarrow\Rx$.
Their \Def{infimal convolution} is the function $\bnch\rightarrow\Rx$
defined by
\begin{gather}
	\label{eq:infimal-convolution-def}
	\Def{(f_1 \infconv \cdots \infconv f_m)(x)}\coloneqq
	\inf\{f_1(x_1)+\cdots +f_m(x_m)\st x_1+\cdots +x_m =x\}.
\end{gather}


\subsection{Support Functions, Subdifferentials and Superdifferentials}
For a  set $S\subseq\bnch$ we define its \Def{convex support function}  
\begin{gather}
	\label{eq:convex-support-function-def}
    \bnch^*\ni x^*\mapsto \Def{\cxsprt_S(x^*)}{\sup\set{\inner{x^*;x} ;
      x\in S}} \in\Rx.
\end{gather}
However, in our setting, it will often be more natural to consider the  
\Def{concave support function} 
\begin{gather}
	\label{eq:concave-support-function-def}
    \bnch^*\ni x^*\mapsto \Def{\cvsprt_S(x^*)}{ \inf\set{\inner{x^*;x};
       x\in S}}\in\Rx.
\end{gather}
The convex and concave support functions are related as follows:
\begin{align}
    \forall{x^*\in\bnch^*},\  \cxsprt_S(x^*) 
    = \sup_{x\in S}\inner{x^*;x}
    = \sup_{x\in -S}\inner{x^*;-x}
    = -\inf_{x\in -S}\inner{x^*;x}
    = -\cvsprt_{-S}(x^*).
    \label{eq:cvx_to_ccv_sprt}
\end{align}
It is easy to see that $\cxsprt$ and $\cvsprt$ are both 1-homogeneous,
$\cxsprt$ is subadditive and thus sublinear; and $\cvsprt$ is superadditive
and thus superlinear. 



We introduce the mappings $\subdiff_\prime f,\subdiff^\prime f \colon \bnch
\to 2^{\bnch^*}$ with
\begin{gather}
    \bnch\ni x \mapsto \Def{\subdiff_\prime f(x)}{\set{x^*\in\bnch^*; 
	    \forall{y\in\bnch},\   
	    f(y)-  f(x) ≥ \inner{x^*; y-x}}},\label{eq:subdiff_defn}\\
    \bnch\ni x \mapsto \Def{\subdiff^\prime f(x)}{\set{x^*\in\bnch^*; 
	    \forall{y\in\bnch},\   
	    f(y)-  f(x) ≤ \inner{x^*; y-x}}}.\label{eq:superdiff_defn}
\end{gather}
The first mapping is the classical \Def{subdifferential}, and the second is
the less-common \Def{concave subdifferential}, or \Def{superdifferential}.
These sets are related by $\subdiff_\prime(f) = -\subdiff^\prime(-f) $. The
mapping
\begin{gather}
    \bnch\ni x \mapsto \subdiff f(x) \coloneqq
    \subdiff_\prime f(x) \cup \subdiff^\prime f(x)
\end{gather}
is known as the \Def{symmetric subdifferential}
\citep{mordukhovich1995nonconvex}.  When $f$ is convex $\subdiff f =
\subdiff_\prime f$, and when $f$ is concave $\subdiff f = \subdiff^\prime
(f)$. In the event that $f$ is both convex and concave $\subdiff f =
\subdiff_\prime f =\subdiff^\prime f$. More importantly, the symmetric
subdifferential satisfies $\subdiff(-f) = -\subdiff f$, which makes it a
convenient choice for us since we deal with (sub/super)-differentials of both
convex and concave functions. We refer to elements of $\subdiff f$ as
\Def{subgradients} (recognising the slight terminological abuse in the
choice of name) and write $\Def{\dom(\subdiff f)}{\set{x\in\bnch;
\subdiff f ≠ \varnothing}}$.
If there is a function $g\colon\bnch\to\bnch^*$ that
is always a subgradient of $f$ in the sense that for all $x\in\dom(\subdiff
f)$ we have $g(x) \in \subdiff f(x)$, then $g$ is called a \Def{selection
of $\subdiff f$} and we write $g\in\subdiff f$. The following proposition
is a standard result \citep[see][]{Penot2012Calculus,Bauschke2011Convex}:
\begin{lemma}\label{prop:subdiff_nonempty}
    Let $f\colon\bnch\to\Rx$ be convex with nonempty domain. 
    Then $\relint(\dom f) \subseq \dom(\subdiff f)$.
\end{lemma}
It is easy to show that for some convex functions the inclusion in
Lemma \ref{prop:subdiff_nonempty} is not strict; For example take
$\subdiff\inner{;s}$, then $\dom(\subdiff\inner{;s}) = \dom(\inner{;s})$.

\begin{proposition}\label{prop:subdiff-zero-homog}
    Suppose $f\colon\bnch\to\Rx$ is 1-homogeneous. Then $\partial f$
    is 0-homogeneous.
\end{proposition}
\begin{proof} From the definition of the subdifferential,
        for all $x\in\bnch$ and all $\alpha>0$, 
    \begin{align}
        \subdiff_\prime f(\alpha x)
	&=\set{x^*\in\bnch^*; \forall{y\in\bnch},\   f(y)-  f(\alpha x) ≥
		\inner{x^*; y-\alpha x}}\\
        &=\set{x^*\in\bnch^*; \forall{\alpha y\in\bnch},\   
	    f(\alpha y)-  f(\alpha x) ≥ \inner{x^*; \alpha y-\alpha x}}\\
        &=\set{x^*\in\bnch^*; \forall{y\in\bnch},\   
	    \alpha f(y)-  \alpha f( x) ≥ \alpha\inner{x^*;  y- x}}\\
        &=\subdiff_\prime f(x),
    \end{align}
    and $\subdiff_\prime f$ is thus 0-homogenous. The proof for the
    superdifferential is similar.
\end{proof}
\vspace*{-4mm}
\begin{lemma}[\protect{\citealp[Corollary~3]{zualinescu2013differentiability}}]\label{thm:cxsprt_differentiable}
    Assume $C\in\cvx(\bnch)$. Then $\cxsprt_C$ is differentiable on
    $\dom(\subdiff\cxsprt_C)\setminus{0}$ if and only if either $C$ is a
    singleton or $\interior(C)≠\varnothing$ and $C$ is strictly convex.
\end{lemma}
Lemma \ref{thm:cxsprt_differentiable} together with \eqref{eq:cvx_to_ccv_sprt} gives us the corollary for the concave case:
\begin{corollary}\label{cor:cvsprt_differentiable}
    Assume $C\in\cvx(\bnch)$. Then $\cvsprt_C$ is differentiable on
    $\dom(\subdiff\cvsprt_C)\setminus{0}$ if and only if either $C$ is a
    singleton or $\interior(C)≠\varnothing$ and $C$ is strictly convex.
\end{corollary}


The \Def{gradient operator} on $\bnch$ is $\Def{\nabla}{(\pdv{x_1},\dots,\pdv{x_n})}$. The following lemma has an obvious extension for the concave case:

\begin{lemma}[\protect{\citealp[Theorem~25.1]{Rockafellar:1970}}]\label{lem:subdiff_differentiable}
    Let $f\colon\bnch\to\Rx$ be convex. If $f$ is differentiable at $x\in\dom(f)$, then $\subdiff f(x)=\single{\nabla f(x)}$. Conversely, if $\subdiff f(x)$ is a singleton at $x\in\dom(f)$ then $f$ is differentiable at $x$.
\end{lemma}
Support functions are oblivious to closures and convex hulls:
\begin{lemma}[\protect{\citealp[Proposition~C.2.2.1]{hiriarturruty2001fca}}]\label{lem:closed_convex_hull_cvx}
    \ \ Suppose $S\subseq\bnch$ is nonempty. Then  
    $\cxsprt_S = \cxsprt_{\cl(S)} = \cxsprt_{\conv(S)}$; 
    whence $\cxsprt_S = \cxsprt_{\cl\conv(S)}$.
\end{lemma}
Using \eqref{eq:cvx_to_ccv_sprt} we find:
\begin{corollary}\label{lem:closed_convex_hull_ccv}
   Suppose  $S\subseq\bnch$ is nonempty. Then  $\cvsprt_S = \cvsprt_{\cl(S)} =
   \cvsprt_{\conv(S)}$; whence $\cvsprt_S = \cvsprt_{\cl\conv(S)}.$
\end{corollary}

\begin{lemma}\label{lem:sprt_domain} 
	Let $S\subseq\bnch$, $S\ne\varnothing$. Then  $\overline{\dom(\cxsprt_S)}
	= -(\rec(\cl\conv S))^*$ and $\overline{\dom(\cvsprt_{S})} =
	\rec(\cl\conv S)^*$.
\end{lemma}
\begin{proof}
    Firstly $\cxsprt_S = \cxsprt_{\cl\conv S}$
    (Lemma \ref{lem:closed_convex_hull_cvx}).  Then from
    \citep[Theorem~2.2.1, p.~32]{Auslender2003}
    $-(\dom\cxsprt_{S})^*=\rec(\cl\conv S)$. Thus since $\dom(\cxsprt_{S})$
    is convex, $\overline{\dom(\cxsprt_{S})} = (-\rec(\cl\conv S))^* =
    -(\rec(\cl\conv S))^*$. For the concave case using
    \eqref{eq:cvx_to_ccv_sprt} we have $\dom(-\cvsprt_{\cl\conv
    S})=\dom(\cxsprt_{- \cl\conv S})=(-\rec(- \cl\conv S))^*=\rec(\cl\conv
    S)^*$.
\end{proof}

\subsection{Gauge Functions} 
\label{sec:gauge_functions_and_polar_duality}

The theory of gauges (Minkowski functionals) and polars has been
traditionally developed for radiant sets
\citep[see][]{hiriarturruty2001fca,Rockafellar:1970,Schneider2014,Thompson1996},
whereas the theory of gauges for shady sets is less well known
\citep{Rockafellar1967,Pukelsheim1983,Barbara1994,Penot2000}. However gauge
functions for shady sets have been used in statistics in a manner similar to
that which we will use them \citep{Pukelsheim1983} and in economics; see
\citep[e.g.][]{Hasenkamp1978} and footnote \ref{footnote:economic-duality}
below.


For convex sets the support function is a natural object to consider.
Likewise, when working with star-shaped and co-star-shaped sets the
\Def{gauge} and \Def{anti-gauge} are a natural parallel. For a set
$S\subseq\bnch$ we define its gauge and anti-gauge:
\begin{align}
    \bnch\ni x\mapsto \Def{\gauge_S(x)}{\inf\set{\lambda>0; x\in\lambda S}} \in\Rx,\label{eq:gauge_defn}\\
    \bnch\ni x\mapsto \Def{\agauge_S(x)}{\sup\set{\lambda>0; x\in\lambda S}}\in\Rx.\label{eq:agauge_defn}
\end{align}
If $S$ is closed and radiant, then $\gauge_S$ is closed and sublinear
\citep[Proposition~2.3]{Penot2000}. Alternately, if $S$ is closed and shady,
then $\agauge_S$ is closed and superlinear \citep[Proposition~2.4]{Penot2000}.
Thus $\gamma_S$ is a \Def{convex gauge} and $\beta_S$ a \Def{concave
gauge}, as they are sometimes described in the literature.
We list some properties of gauge functions and their associated sets in
Table \ref{tab:summary_of_convex_concave_gauges} and graphically in
Figure \ref{fig:relationships_between_sets_gauges}. 
For closed $S$, the base
star-shaped and co-star-shaped sets can be recovered with the inverse
mappings $\gauge_S\mapsto \lev_{≤1}(\gauge_S) =S$ and $\agauge_S\mapsto
\lev_{≥1}(\agauge_S)=S$ respectively. Mirroring Lemma \ref{lem:sprt_domain},
\citet{Penot2000} observed that when $S$ is closed
\begin{gather}
	\dom(\gauge_S) = \cone(S) \cup \single{0}\subseq\rec(S) \mbox{\ \
	and\ \ } \dom(\agauge_S) = \cone(S) \cup \single{0}\subseq\rec(S).
\end{gather}
As one might expect 
\begin{gather}
	\dom(\gauge_S) = \cl\cone (S)\mbox{\ \ and\ \ }
	\dom(\agauge_S) = \cl\cone(S).
\end{gather}
If $0\in\interior S$, then $\dom(\gauge_S) =\bnch$ for all star-shaped sets
$S$. Conversely, for any co-star-shaped set $S\subset\bnch$ we have
$\cone(S) \subset \bnch$. In general this is the key difference between
gauge and anti-gauge functions. That is, while gauge functions can be
finite on the whole space, anti-gauges are usually finite only on a conic subset
of $\bnch$. This is significant with regard to the equivalence of gauges or
norms.  Recall two gauge functions
$\gamma_{T_1},\gamma_{T_2}\colon\bnch\rightarrow\R$ are equivalent if there
exists constants $c,C\in\Rpp$ such that for all $x\in\bnch$,
$c\gamma_{T_2}(x)\le \gamma_{T_1}(x)\le C\gamma_{T_2}(x)$.
Whilst all convex gauges and norms in finite dimensional
space are equivalent, 
that is \emph{not} true in general for concave gauges or anti-norms
$\beta_{S_1},\beta_{S_2}\colon\bnch\rightarrow\Rx$ if at least
one of the concave gauges can take on the value $+\infty$ for some
$x\in\bnch$.  Unbounded concave gauges correspond to unbounded loss
functions; a point which will be elaborated below.

\begin{table}
	    \arrayrulecolor{lightgray}
    \centering
    \small
    \begin{tabular}{C{5em} C{5.9em} C{7em} C{5.5em} C{2.7em}} \toprule
        &\multicolumn{4}{c}{Gauge function $\gauge_S$} \\ \cmidrule{2-5}
        $S$ & Non-negative & 1-Homogeneous &  Subadditive  & Norm \\ \midrule
        \emph{star-shaped} & $\bullet$ & $\bullet$ &  & \\
        \emph{radiant} & $\bullet$ & $\bullet$ & $\bullet$ &\\
        \emph{norm ball} & $\bullet$ & $\bullet$ & $\bullet$ & $\bullet$\\ 
	\bottomrule
    \end{tabular}
    
    \bigskip
    
    \begin{tabular}{C{6.3em} C{5.9em} C{7em} C{6.2em} } \toprule
         &\multicolumn{3}{c}{Anti-gauge function $\agauge_S$} \\ \cmidrule{2-4}
        $S$ & Non-negative &  1-Homogeneous  & Superadditive \\ \midrule
        \emph{co-star-shaped} & $\bullet$ & $\bullet$ & \\
        \emph{shady} & $\bullet$ & $\bullet$ & $\bullet$ \\ \bottomrule
    \end{tabular}
    \bigskip
    \caption{Properties of gauge and anti-gauge functions when restricted
	    to their domains, as determined by their defining sets. See
	    also Figure \ref{fig:relationships_between_sets_gauges}.}
    \label{tab:summary_of_convex_concave_gauges}
\end{table}

The attentive reader will notice the similarity between the properties of
the gauge of a norm ball and the properties of a norm on $\bnch$. Indeed
every norm on $\bnch$, $\norm{}$, can be written as a gauge of the norm
ball $\lev_{≤1}\norm{}$, and conversely the gauge of every norm ball, as
defined in \S\ref{sec:preliminaries}, is a norm. If one restricts
analysis to the set $\cone(S)$ for a shady set $S\subset\bnch$---in effect
ruling out multiplication by negative scalars---the function $\agauge_S:
\cone(S) \to \R$ is a natural counterpart to a norm on this space, which we
call an \Def{anti-norm}. As we shall see in \S\ref{sec:loss_functions},
the conditional Bayes' risks associated with proper losses are in fact
anti-norms\footnote{Anti-gauges are  sometimes called ``anti-norms''
	\citep{Berestovskii:2004aa,Moszynska:2012aa,Merikoski:1991aa},
	although confusingly this name is sometimes used to refer to the
	dual (polar) of a traditional norm 
	\citep{Horvath:2017aa,Martini:2006aa}. We will thus stick
with the terminology ``anti-gauge''.}.

\subsection{Legendre-Fenchel Conjugates}
The \Def{Legendre--Fenchel conjugate} or \Def{convex conjugate} of $f$ is the function
\begin{gather}
    \bnch^* \ni x^*\mapsto
    \Def{f^*(x^*)}{\sup_{x\in\bnch}{\left(\inner{x^*;x} - f(x)\right)}} \in \Rx. \label{eq:convex_conjugate}
\end{gather}
In addition to the more common convex Legendre--Fenchel conjugate, we will
make use of the concave conjugate, which, like the case of the
concave support function, will be more appropriate for our purposes. The
\Def{concave conjugate} of $f$ is 
\begin{gather}
    \bnch^* \ni x^*\mapsto
    \Def{f_*(x^*)}{\inf_{x\in\bnch}{\left(\inner{x^*;x} - f(x)\right)}} \in \Rx, \label{eq:concave_conjugate}
\end{gather}
and is related to the convex conjugate as follows:
\begin{gather}
f^*(x^*)= \sup_{x\in\bnch}{\left(\inner{x^*;x} - f(x)\right)}
=-\inf_{x\in\bnch}{\left(\inner{-x^*;x} - (-f)(x)\right)}
=
-(-f)_*(-x^*).
\end{gather}
The concave conjugate therefore satisfies a \Def{reverse Fenchel--Young
inequality}:
\begin{gather}
	\forall{x\in \bnch},\ \forall x^*\in \bnch^*,\  f(x) + f_*(x^*) ≤ \inner{x^*;x}.\label{eq:rev_fy_inequality}
\end{gather}

A function $f$ is \Def{lower semi-continuous} if for all $x\in\bnch$ and for all sequences $(x_n)_{n\in\N}$ with $x_n\in\bnch$ and $x_n\to x$ we have $f(x) \leq \liminf_{n\to\infty} f(x_n)$. If $-f$ is lower semi-continuous then $f$ is said to be \Def{upper semi-continuous}.

\subsection{Polar Duality}


While we do make some use of the above Legendre-Fenchel duality, more
important for our purposes is the polar duality of convex sets
\citep[Chapter 13]{Moszynska:2006ua}.  It arises from the classical
polarity between points and lines relative to the unit circle in inversive
geometry \citep{Askwith:1917tc}.  Following
\citet[\S VII.5.C]{Berger:2010wx},  define a bijection between the set of all
points other than the origin onto the set of hyperplanes not containing the
origin via $x\mapsto h_x\coloneqq \{y\in\R^n\st \inner{x;y}=1\}$, which is known
as the polar hyperplane of the point $x$. Associated with $h_x$ is the
halfspace $H_x\coloneqq\{y\in\R^n\st\inner{x;y}\le 1\}$.  Given a set
$C\subset\R^n$ containing the origin, the \emph{polar} of $C$ is simply
$\bigcap_{x\in C} H_x$.  

More formally, there is a bijection between the
sublinear functions $\cxsprt_A$ and closed convex sets $A$; and another
bijection between the superlinear functions $\cvsprt_B$ and closed convex sets
$B$ \cite{hiriarturruty2001fca}. 
Noting Table \ref{tab:summary_of_convex_concave_gauges} we see that the
gauge and anti-gauge functions also satisfy the same criteria as the
support functions when restricted to (respectively) the radiant and shady subsets. A
natural question to ask then is: can we find sets $A'$ and $B'$ such that
$\cxsprt_{A'}=\gauge_A$ and $\cvsprt_{B'}=\agauge_B$? The answer in the
convex case with respect to the radiant sets is well known, but there is an
equally rich, parallel structure in the concave case with respect to the
shady sets. \citet{Penot2012Calculus} and \citet{zalinescu2002convex} show the
following results for the convex case, and  \citet{Penot2000} prove the
concave case.

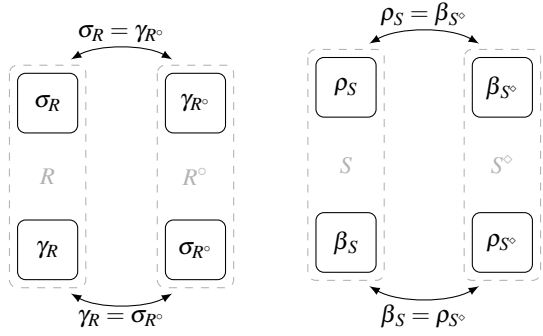
\begin{figure}[t]
    \centering
    \tikzsetnextfilename{relationships_between_polars1}
    \begin{tikzpicture}[every node/.style = {rounded corners=3pt, draw, minimum size=2.5em}]
      \node [draw=none] (o) {};
      \node [above left  = 0.75em of o] (a) {$\cxsprt_R$};
      \node [below left  = 0.75em of o] (b) {$\gauge_R$};
      \node [above right = 0.75em of o] (c) {$\gauge_{R^\polar}$};
      \node [below right = 0.75em of o] (d) {$\cxsprt_{R^\polar}$};
      
      \path (a.north) edge [bend left, latex-latex, shorten >=10pt, shorten <=10pt] 
        node[draw=none, midway, above, minimum size=0] {$\cxsprt_R=\gauge_{R^\polar}$} (c.north);
      \path (d.south) edge [bend left, latex-latex, shorten >=10pt, shorten <=10pt] 
        node[draw=none, midway, below, minimum size=0] {$\gauge_R=\cxsprt_{R^\polar}$} (b.south);
      \node[draw=none,faint] (S) at ($(a)!0.5!(b)$) {$R$};
      \node[draw=none,faint] (Sp) at ($(c)!0.5!(d)$) {$R^\polar$};
      \draw[dashrounded] 
        ($(a.north west)+(-0.1,0.1)$) rectangle ($(b.south east)+(0.1,-0.1)$);
      \draw[dashrounded] 
        ($(c.north west)+(-0.1,0.1)$) rectangle ($(d.south east)+(0.1,-0.1)$);
    \end{tikzpicture}
    \qquad
    \tikzsetnextfilename{relationships_between_polars2}
    \begin{tikzpicture}[every node/.style = {rounded corners=3pt, draw, minimum size=2.5em}]
      \node [draw=none] (o) {};
      \node [above left = 1em of o] (a) {$\cvsprt_S$};
      \node [below left = 1em of o] (b) {$\agauge_S$};
      \node [above right = 1em of o] (c) {$\agauge_{S^\apolar}$};
      \node [below right = 1em of o] (d) {$\cvsprt_{S^\apolar}$};
      \path (a.north) edge [bend left, latex-latex, shorten >=10pt, shorten <=10pt] 
        node[draw=none, midway, above, minimum size=0] {$\cvsprt_S=\agauge_{S^\apolar}$} (c.north);
      \path (d.south) edge [bend left, latex-latex, shorten >=10pt, shorten <=10pt] 
        node[draw=none, midway, below, minimum size=0] {$\agauge_S=\cvsprt_{S^\apolar}$} (b.south);
      \node[draw=none,faint] (S) at ($(a)!0.5!(b)$) {$S$};
      \node[draw=none,faint] (Sp) at ($(c)!0.5!(d)$) {$S^\apolar$};
      \draw[dashrounded] 
        ($(a.north west)+(-0.1,0.1)$) rectangle ($(b.south east)+(0.1,-0.1)$);
      \draw[dashrounded] 
        ($(c.north west)+(-0.1,0.1)$) rectangle ($(d.south east)+(0.1,-0.1)$);
    \end{tikzpicture}
    \caption{The polar duality of support functions and gauge functions for radiant
    ($R$) and shady ($S$) sets.}
    \label{fig:relationships_between_polars}
\end{figure}

For a set $S\subseq\bnch$ its \Def{polar} and \Def{antipolar} are the sets
\begin{gather}
	\Def{S^\polar} \coloneqq
    \set{x^*\in\bnch^*;\forall{x\in S},\ \inner{x^*;x}≤1}
    \mbox{\ \ and\ \ }  \Def{S^\apolar} \coloneqq
    \set{x^*\in\bnch^*;\forall{x\in S},\ \inner{x^*;x} ≥ 1}.
    \label{eq:polar_antipolar_dfn}
\end{gather}
Equivalently $S^\polar = \lev_{≤1}\cvsprt_S$ and $S^\apolar =
\lev_{≥1}\cvsprt_S$. It is easy to show that the polarity operation
$S\mapsto S^\polar$ takes closed radiant sets to closed radiant sets, and
the antipolarity operation $S\mapsto S^\apolar$ takes closed shady sets to
closed shady sets such that if $S\in\shdy(\pcone)$, then
$S^\apolar\in\shdy(\fcone)$. The polars and bipolars satisfy
\citep[Lemma~4.2]{Penot2000}:
\def\t{(0,1]}
\begin{gather}
    \begin{aligned}
        S^{\polar}&=(\cl\co((0,1]·S))^\polar\\
        S^{\polar\polar}&=\cl\co(\t·S)
    \end{aligned}
    \quad\mbox{\ and\ } \quad
    \begin{aligned}
        S^{\apolar}&=(\cl\co([1,\infty)·S))^\apolar\\
        S^{\apolar\apolar}&=\cl\co([1,\infty)·S),
    \end{aligned}
    \label{eq:bipolar_theorem}
\end{gather}
where the operations $S\mapsto\cl\co(\t·S)$ and
$S\mapsto\cl\co([1,\infty)·S)$ are known as the \Def{radiant hull} and
	\Def{shady hull}
	respectively. The polar and antipolar operations also induce a
	natural duality relationship between the gauge and support
	functions:
\begin{gather}
    \begin{gathered}
        \forall R\in\rdnt(X),\quad 
	   \cxsprt_{R}= \gauge_{R^\polar},\quad \cxsprt_{R^\polar}= \gauge_{R};
    \end{gathered}
    \quad\mbox{\ and\ } \quad
    \begin{gathered}
	    \forall S\in\shdy(X),\quad 
        \cvsprt_{S}= \agauge_{S^\apolar},\quad
        \cvsprt_{S^\apolar}= \agauge_{S},\label{eq:gauge_support_polar}
    \end{gathered}
    \intertext{such that we may define the \Def{function polar and
    antipolar}:}
    \begin{gathered}
	    \label{eq:function-polar-def}
	    \Def{\cxsprt_{R}^\polar\coloneqq \cxsprt_{R^\polar}},\quad
	    \Def{\gauge^\polar_{R}\coloneqq \gauge_{R^\polar}};
    \end{gathered}
    \quad\mbox{\ and\ } \quad
    \begin{gathered}
	    \Def{\cvsprt_{S}^\apolar\coloneqq \cvsprt_{S^\apolar}},\quad
	    \Def{\agauge^\apolar_{S}\coloneqq\agauge_{S^\apolar}}.
    \end{gathered}
\end{gather}
The above relationships are presented diagrammatically in
Figure \ref{fig:relationships_between_polars}.

The polar (antipolar) relationship between the convex (concave) 
support functions and gauge functions motivates a
definition of the polar (antipolar) for sub/super-linear 
non-negative functions that is
independent of its definition as a support function or gauge of a set. Let
$f,g\colon\bnch\to\R$ with $f$ sublinear, and $g$ superlinear. Convex and
concave polar duality correspondences for $f$ and $g$ are given by 
\begin{gather}
    \bnch^*\ni x^*\mapsto
    \Def{f^\polar(x^*)}{\sup_{x≠0}\frac{\inner{x^*;x}}{f(x)}} \in\Rx
    \mbox{\ \ \ and\ \ \ } \bnch^*\ni x^*\mapsto 
    \Def{g^\apolar(x^*)}{\inf_{x≠0}\frac{\inner{x^*;x}}{g(x)}} \in \Rx.
\end{gather}
Thus $f^\polar$ and $g^\apolar$ satisfy a \Def{generalised Hölder} and
\Def{reverse Hölder} inequality respectively:
\begin{gather}
    \forall{x\in \bnch, x^*\in \bnch^*},\  \inner{x^*;x}≤f^\polar(x^*)f(x)
    \mbox{\ \ \ and\ \ \ }  \inner{x^*;x}≥g^\apolar(x^*)g(x).
    \label{eq:holder}
\end{gather}
The case of Hölder conjugate norms --- $\norm{}_p$ and $\norm{}_q$ with
$\oneon{p}+\oneon{q}=1$ --- can easily be derived as a special case of the
polar duality relationships above, with $\ball_q = \ball_p^\polar$ and
$\ball_p = \ball_q^\polar$.

We will make use of the following result of \cite{Barbara1994} which can be
seen to be analogous to the classical result \citep[Proposition
E.1.4.3]{hiriarturruty2001fca} regarding subdifferentials of
Legendre--Fenchel conjugates. We express the result for the concave case
because that is what we need for losses.  An analogous result holds for
convex gauges.

\begin{lemma}[\protect{\citealt[Theorem~3.1]{Barbara1994}}]\label{lem:polars_and_inverses}
    Let  $S\in\shdy(\bnch)$. Then for $x\in\dom\agauge_S$ and $x^*\in\dom\agauge_{S^\apolar}$ 
    \begin{gather}
        \frac{x}{\agauge_{S}(x)}\in\subdiff\agauge_{S^\apolar}(x^*)
                    \iff
                    \frac{x^*}{\agauge_{S^\apolar}(x^*)}\in\subdiff\agauge_S(x)
                    \iff
                    \agauge_{S^\apolar}(x^*)\agauge_S(x)= \inner{x^*;x}.
        \label{eq:subdifferential-of-gauge}
    \end{gather}
\end{lemma}
\citet{Barbara1994} provided a sketch of a proof. However as it is central to
what follows we present a complete proof below.
\begin{proof}
    We first prove
    \begin{gather}
        \forall{s\in S},\ \ \underbrace{\varfrac{x^*}{\agauge_{S^\apolar}(x^*)}\in \subdiff\agauge_S(s)}_{\mathrm{(A)}} \iff
        \underbrace{\g(
                    \varfrac{x^*}{\agauge_{S^\apolar}(x^*)}\in
		    S^\apolar\mbox{\ and\ } 
                    \agauge_S(s)=\inner{\varfrac{x^*}{\agauge_{S^\apolar}(x^*)};s}
                )}_{\mathrm{(B)}}.
        \label{eq:partial-claim}
    \end{gather}
    
    For the sufficient condition suppose $s\in S$ and
    $\varfrac{x^*}{\agauge_{S^\apolar}(x^*)}\in\subdiff\agauge_S(s)$. Then from 
    \eqref{eq:superdiff_defn} we have
    \begin{gather}
        \forall{y\in\bnch},\  \agauge_S(y) ≤ \agauge_S(s) + \inner{\varfrac{x^*}{\agauge_{S^\apolar}(x^*)}; y-s}.\label{eq:r1} 
    \end{gather}
    Since by assumption $S$ is shady, $s≠0$. Setting $y=0$ and then $y=2s$, and
    exploiting the 1-homogeneity of $\agauge_S$ and \eqref{eq:r1} 
    \begin{gather}
        \agauge_S(0) = 0 
        ≤ \agauge_S(s) - \inner{\varfrac{x^*}{\agauge_{S^\apolar}(x^*)};s} 
        \implies \agauge_S(s) 
        ≥ \inner{\varfrac{x^*}{\agauge_{S^\apolar}(x^*)};s}\label{eq:r1a},\\
        \agauge_S(2s) = 2\agauge_S(s)
        ≤ \agauge_S(s) + \inner{\varfrac{x^*}{\agauge_{S^\apolar}(x^*)};s} 
        \implies
        \agauge_S(s) ≤\inner{\varfrac{x^*}{\agauge_{S^\apolar}(x^*)};s}\label{eq:r1b}.\\
        \intertext{Together, \eqref{eq:r1a} and \eqref{eq:r1b} give}
        \agauge_S(s)=\inner{\varfrac{x^*}{\agauge_{S^\apolar}(x^*)};s}. \label{eq:rr}
    \end{gather}
    Combining \eqref{eq:r1} with \eqref{eq:rr} gives
    \begin{align}
         & \mathrm{(A)} \implies \forall{y\in\bnch},\  \agauge_S(y) ≤ \inner{\varfrac{x^*}{\agauge_{S^\apolar}(x^*)};s} + \inner{\varfrac{x^*}{\agauge_{S^\apolar}(x^*)}; y-s} = \inner{\varfrac{x^*}{\agauge_{S^\apolar}(x^*)};y},\\
        \iff\ \ \ \  & \mathrm{(A)} \implies \forall{y\in\bnch},\  \agauge_S(y) ≤ \inner{\varfrac{x^*}{\agauge_{S^\apolar}(x^*)};y}. \label{eq:r2}
    \end{align}
    Since $S$ is closed and shady, $\lev_{\ge 1}\agauge_S=S$ and
    \begin{gather}
        \forall{y\in\bnch},\  1 ≤ \agauge_S\g(s) \overset{\eqref{eq:rr}}{=} \inner{\varfrac{x^*}{\agauge_{S^\apolar}(x^*)};s} \overset{\eqref{eq:r2}}{≤} \inner{\varfrac{x^*}{\agauge_{S^\apolar}(x^*)};y}
        \implies \forall{s'\in S},\  1 ≤ \inner{\varfrac{x^*}{\agauge_{S^\apolar}(x^*)};s'}.
    \end{gather}
    Thus $\varfrac{x^*}{\agauge_{S^\apolar}(x^*)}\in S^\apolar$ by \eqref{eq:polar_antipolar_dfn}.

    For the necessary condition suppose now that 
    $\varfrac{x^*}{\agauge_{S^\apolar}(x^*)}\in S^\apolar$ and let
    $s\in S$ be such that $\agauge_S(s)=
    \inner{\varfrac{x^*}{\agauge_{S^\apolar}(x^*)};s}$. Then
    \begin{align}
	    \hspace*{-6mm}0=\agauge_S(s) +
	       \inner{\varfrac{x^*}{\agauge_{S^\apolar}(x^*)};-s}
        &\iff \forall{y\in\bnch},\  \inner{\varfrac{x^*}{\agauge_{S^\apolar}(x^*)};y}  
        = \agauge_S(s) + \inner{\varfrac{x^*}{\agauge_{S^\apolar}(x^*)};y-s}.
        \label{eq:r3}
    \end{align}
    The reverse H\"{o}lder inequality \eqref{eq:holder} gives
    \begin{gather}
        \forall{y\in\bnch},\  \left[ \agauge_{S^\apolar}(x^*)\agauge_S(y)≤\inner{x^*;y} \iff 
	\agauge_{S}(y)≤\inner{\varfrac{x^*}{\agauge_{S^\apolar}(x^*)};y}\right]
	\label{eq:rvfnch}.
    \end{gather}
    Combining \eqref{eq:r3} with \eqref{eq:rvfnch} we have 
    \begin{gather}
       \mathrm{(B)} \implies \forall{y\in\bnch},\  \agauge_S(y)≤ \agauge_S(s) + \inner{\varfrac{x^*}{\agauge_{S^\apolar}(x^*)};y-s}.
    \end{gather}
    Thus $\varfrac{x^*}{\agauge_{S^\apolar}(x^*)}\in\subdiff\agauge_S(s)$, and \eqref{eq:partial-claim} is proved. 
    
    Let $x\in\dom(\agauge_S)$. Then
    $\agauge_S\g\big(\varfrac{x}{\agauge_S(x)}) = 1$. This follows since
    $\agauge_S$ is 1-homogenous. Since $S=\lev_{\ge 1}\agauge_S$, we have
    $\varfrac{x}{\agauge_S(x)} \in S$. Substituting $s =
    \varfrac{x}{\agauge_{S}(x)}$ in \eqref{eq:partial-claim} implies that
    for all $x^*\in\dom(\agauge_{S^\apolar})$
    \begin{gather}
        \varfrac{x^*}{\agauge_{S^\apolar}(x^*)}\in \subdiff\agauge_S(x) \iff 
           \g(\varfrac{x^*}{\agauge_{S^\apolar}(x^*)}\in S^\apolar\mbox{\
	   and\ }
                       \oneon{\agauge_S(x)}·\agauge_S\g(x)=\inner{\varfrac{x^*}{\agauge_{S^\apolar}(x^*)};\varfrac{x}{\agauge_{S}(x)}}),\label{eq:semifinal1}
    \end{gather}
    where we used the 0-homogeneity of $\subdiff\agauge_S$ (Proposition 
    \ref{prop:subdiff-zero-homog}) to obtain
    $\subdiff\agauge_S(\varfrac{x}{\agauge_{S}(x)}) =
    \subdiff\agauge_S(x)$. Since $S$ is shady we can apply the bipolar
    theorem \eqref{eq:bipolar_theorem} to obtain the equivalent condition
    for all $x^*\in\dom(\agauge_{S^\apolar})$:
    \begin{gather}
        \varfrac{x}{\agauge_{S}(x)}\in \subdiff\agauge_S^\apolar(x^*) \iff 
        \g(
	\varfrac{x}{\agauge_{S}(x)}\in S\mbox{\ and\ }
            \oneon{\agauge_{S^\apolar}\g(x^*)}·\agauge_{S^\apolar}\g(x^*)=\inner{\varfrac{x^*}{\agauge_{S^\apolar}(x^*)};\varfrac{x}{\agauge_{S}(x)}}
        ).\label{eq:semifinal2}
    \end{gather}
    Finally observe
    \begin{gather}
        1 =\inner{\varfrac{x^*}{\agauge_{S^\apolar}(x^*)};\varfrac{x}{\agauge_{S}(x)}}
        \iff \agauge_{S^\apolar}(x^*)\agauge_{S}(x) = \inner{x^*;x},
    \end{gather}
    which concludes the proof.
\end{proof}

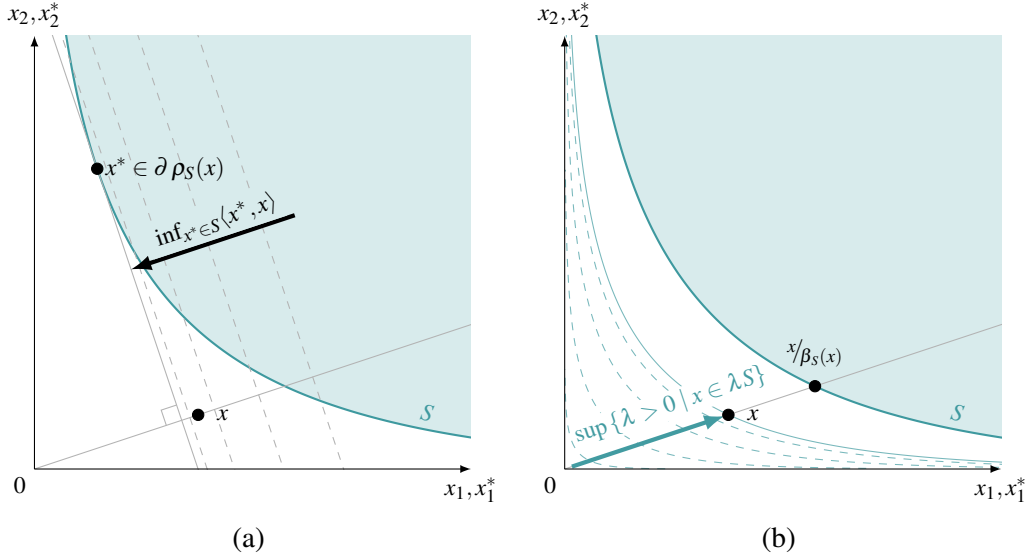
\begin{figure}[t]
    \centering
    \tikzsetnextfilename{support_functions}
    \subcaptionbox{}{
    \begin{tikzpicture}
        \begin{axis}[domain=0.05:0.95, xmin=0, xmax=2, ymin=0, ymax=2, width=0.5\textwidth, height=0.5\textwidth,
            xlabel={$x_1, x^*_1$}, ylabel={$x_2, x^*_2$}, xtick={0},ytick={0}]
            \coordinate (o) at (axis cs: 0,0);
            \path[name path=upper_axis] (axis cs:0,5) -- (axis cs:5,0);

            \addplot[set, color=Cyan, name path global=l] ({-ln(x)},{-ln(1-x)}) 
                node[pos=0.25, anchor=south, sloped] {$S$}; 
            \addplot[draw=none, name path global=m] ({-x/(x*ln(x) + (1-x)*ln(1-x))},{(x-1)/(x*ln(x) + (1-x)*ln(1-x))});

            \addplot[fill=LightCyan] fill between[of=l and upper_axis];

            \coordinate (p) at (axis cs:0.75,0.25);
            \coordinate (lp) at (axis cs:0.2876820725,1.3862943611);

            \addplot[domain=0:1.8, name path global=p_ray,draw=none] ({0.75*x},{0.25*x});
            
            \draw[name intersections={of=p_ray and m, name=llp}];
            \draw[name intersections={of=p_ray and l, name=lpp}]; 
            \draw[shorten >=-4cm, shorten <=-2cm,faint] (o) -- ($5*(p)$);
            
            \coordinate (xmin) at ($(o)!(lp)!(llp-1)$);

            \draw[shorten >=-4cm, shorten <=-2cm,faint] (xmin) -- (lp);
            
            \draw[shorten >=-4cm, shorten <=-2cm,faint, dashed] ($(xmin) + 0.8*(p)$) -- ($(lp) + 0.8*(p)$);
            \draw[shorten >=-4cm, shorten <=-2cm,faint, dashed] ($(xmin) + 0.47*(p)$) -- ($(lp) + 0.47*(p)$);
            \draw[shorten >=-4cm, shorten <=-2cm,faint, dashed] ($(xmin) + 0.2*(p)$) -- ($(lp) + 0.2*(p)$);
            \draw[shorten >=-4cm, shorten <=-2cm,faint, dashed] ($(xmin) + 0.05*(p)$) -- ($(lp) + 0.05*(p)$);
           plit
            \draw[ultra thick, latex-] ($(xmin) + 0.6*(lp) - 0.6*(xmin)$)
	    -- ($(xmin) + (p) + 0.6*(lp) - 0.6*(xmin)$) node[pos = 0.55,
	    inner sep=2pt, sloped, anchor=center, above, fill=LightCyan]
	    {$\:\inf_{x^*\in S}\inner{x^*;x}$};
            
            \draw[right angle quadrant=2, right angle symbol={o}{llp-1}{lp},faint];    
            
            \draw (p) node[dot] {} node[anchor=west, xshift=1ex] {$x$};
            \draw (lp) node[dot] {} node[anchor=west] {$x^*\in\subdiff\cvsprt_S(x)$};
        \end{axis}
        \draw (o) node[anchor = north east] {$0$};
    \end{tikzpicture}}
    \tikzsetnextfilename{gauge_functions}
    \subcaptionbox{}{
    \begin{tikzpicture}
        \begin{axis}[domain=0.05:0.95, xmin=0, xmax=2, ymin=0, ymax=2, width=0.5\textwidth, height=0.5\textwidth,
            xlabel={$x_1, x^*_1$}, ylabel={$x_2, x^*_2$}, xtick={0},ytick={0}]
            \coordinate (o) at (axis cs: 0,0);
            \path[name path=upper_axis] (axis cs:0,5) -- (axis cs:5,0);
            
            \addplot[color=Cyan!70, name path global=l, domain=0.01:0.99] ({-0.655*ln(x)},{-0.655*ln(1-x)});
            \addplot[color=Cyan!70, name path global=l, dashed, domain=0.01:0.99] ({-0.6*ln(x)},{-0.6*ln(1-x)});
            \addplot[color=Cyan!70, name path global=l, dashed, domain=0.01:0.99] ({-0.48*ln(x)},{-0.48*ln(1-x)});
            \addplot[color=Cyan!70, name path global=l, dashed, domain=0.01:0.99] ({-0.3*ln(x)},{-0.3*ln(1-x)});
            \addplot[color=Cyan!70, name path global=l, dashed, domain=0.01:0.99] ({-0.1*ln(x)},{-0.1*ln(1-x)});

            \addplot[set, color=Cyan, name path global=l] ({-ln(x)},{-ln(1-x)}) 
                node[pos=0.25, anchor=south, sloped] {$S$}; 
            \addplot[draw=none, name path global=m] ({-x/(x*ln(x) + (1-x)*ln(1-x))},{(x-1)/(x*ln(x) + (1-x)*ln(1-x))});

            \addplot[fill=LightCyan] fill between[of=l and upper_axis];

            \coordinate (p) at (axis cs:0.75,0.25);
            \coordinate (lp) at (axis cs:0.2876820725,1.3862943611);

            \addplot[domain=0:1.8, name path global=p_ray,draw=none] ({0.75*x},{0.25*x});
            
            \draw[name intersections={of=p_ray and m, name=llp}];
            \draw[name intersections={of=p_ray and l, name=lpp}]; 
            \draw[shorten >=-4cm, shorten <=-2cm,faint] (o) -- ($5*(p)$);
            
            \coordinate (xmin) at ($(o)!(lp)!(llp-1)$);

            
            
            \draw[faint, ultra thick, -latex, color=Cyan!98, shorten <=0.1cm, shorten >=1.5pt]  ($0.0*(lp) - 0.0*(p)$) -- ($(p) + 0.0*(lp) - 0.0*(p)$) node[pos=0.7, sloped, anchor=center, above, fill=white, inner sep=1pt, above=1ex] {$\sup\set{\lambda > 0; x\in \lambda S}$};
            
            
            \draw (p) node[dot] {} node[anchor=west, xshift=1ex] {$x$};
            \draw (lpp-1) node[dot] {} node[anchor=south, yshift=1ex] {$\varfrac{x}{\agauge_S(x)}$};
        \end{axis}
        \draw (o) node[anchor = north east] {$0$};
    \end{tikzpicture}
    }
    \caption{ Illustration of concave support function $\rho_S$
	    and concave gauge function $\beta_S$.  In (a) the set $S$ is
	    supported by the hyperplane normal to $x$ at the point $x^*$
	    with offset equal to $\rho_S(x)$.
	    In (b), picking a point $x$, we consider scaled versions
	    $\lambda S$ of $S$
	    such that $x$ ends up on the boundary of $\lambda S$, thus
	    giving the value of $\beta_S(x)$. The ray
	    passing through $x$ intersects the boundary of $S$ at
	    $x/\beta_S(x)$.
    }
    \label{fig:support_functions}
\end{figure}

\begin{lemma}\label{lem:shady_strict_agauge}
    Suppose $S\in\shdy(\bnch)$. Then $\agauge_S$ is strictly concave if and only if $S$ is strictly convex.
\end{lemma}
\begin{proof}
    We show the sufficient condition using a proof by contradiction. Assume
    (for the contradiction)
    $\agauge_S$ is strictly convex. If $S$ is not strictly convex there
    exists $x,y\in S$ and $t\in(0,1)$ such that $tx + (1-t)y\in\bd(S)$.
    Since $S$ is convex it follows that $\relint(S)$ is convex. And so it
    must be the case that $x,y\in\bd(S)$. Since $S$ is shady,
    and by hypothesis $\agauge_S$ is strictly convex, it follows that
    \begin{gather}
        \underbrace{\agauge_S(tx + (1-t)y)}_1 < t\underbrace{\agauge_S(x)}_1 + (1-t)\underbrace{\agauge_S(y)}_1 = 1,
    \end{gather}
    which is absurd. Thus $S$ is strictly convex.
    
    For necessity, choose arbitrary $x,y\in S$ with $x\neq y$. Then
    $\varfrac{x}{\agauge_S(x)},\varfrac{y}{\agauge_S(y)}\in\bd(S)$ and
    $\agauge_S(\varfrac{x}{\agauge_S(x)})=\agauge_S(\varfrac{y}{\agauge_S(y)})=1$.
    Since $S$ is strictly convex $t\varfrac{x}{\agauge_S(x)} +
    (1-t)\varfrac{y}{\agauge_S(y)}\in\relint(S)$ for all $t\in(0,1)$. Thus
    \begin{align}
       \MoveEqLeft[0]\agauge_S\g(\varfrac{t}{\agauge_S(x)}x + \varfrac{(1-t)}{\agauge_S(y)}y) > 
       1 = t\agauge_S\g(\varfrac{x}{\agauge_S(x)})+(1-t)\agauge_S\g(\varfrac{y}{\agauge_S(y)})\\
       \iff\ \ \ \ \  & 
       \agauge_S\g(\varfrac{t}{\agauge_S(x)}x + \varfrac{(1-t)}{\agauge_S(y)}y) > 
       \varfrac{t}{\agauge_S(x)}\agauge_S\g(x)+\varfrac{(1-t)}{\agauge_S(y)}\agauge_S\g(y),
    \end{align}
    where in the second line we exploited the 1-homogeneity of $\agauge_S$. 
    Multiplying both sides by $\frac{1}{\varfrac{t}{\agauge_S(x)} +
    \varfrac{(1-t)}{\agauge_S(y)}}$ and exploiting 1-homogeneity again
    gives $\forall t\in(0,1)$
    \begin{gather}
       \begin{aligned}
        & \frac{1}{\varfrac{t}{\agauge_S(x)} +
	\varfrac{(1-t)}{\agauge_S(y)}}\agauge_S\g(\varfrac{t}{\agauge_S(x)}x
	+ \varfrac{(1-t)}{\agauge_S(y)}y) >
\\& \hspace{12em}
       \frac{\varfrac{t}{\agauge_S(x)}}{\varfrac{t}{\agauge_S(x)} + \varfrac{(1-t)}{\agauge_S(y)}}\agauge_S\g(x) + 
       \frac{\varfrac{(1-t)}{\agauge_S(y)}}{\varfrac{t}{\agauge_S(x)} + \varfrac{(1-t)}{\agauge_S(y)}}\agauge_S\g(y)\\
           \iff\ 
           &\agauge_S\g\bigg(\frac{\varfrac{t}{\agauge_S(x)}}{\varfrac{t}{\agauge_S(x)} + \varfrac{(1-t)}{\agauge_S(y)}}x + \frac{\varfrac{(1-t)}{\agauge_S(y)}}{\varfrac{t}{\agauge_S(x)} + \varfrac{(1-t)}{\agauge_S(y)}}y) > \\
           &\hspace{12em}\frac{\varfrac{t}{\agauge_S(x)}}{\varfrac{t}{\agauge_S(x)} + \varfrac{(1-t)}{\agauge_S(y)}}\agauge_S\g(x) + 
           \frac{\varfrac{(1-t)}{\agauge_S(y)}}{\varfrac{t}{\agauge_S(x)} + \varfrac{(1-t)}{\agauge_S(y)}}\agauge_S\g(y),
       \end{aligned}
    \end{gather}
    and thus $\agauge_S$ is strictly concave.
\end{proof}
\begin{corollary}\label{lem:apolar_strictly_convex}
    Let $S\in\shdy(\bnch)$. Then $\cvsprt_S$ strictly concave if and only if $S^\apolar$ is strictly convex.
\end{corollary}
\begin{proof}
    Apply Lemma \ref{lem:shady_strict_agauge} to
    $\cvsprt_S=\agauge_{S^\apolar}$.
\end{proof}

\begin{lemma}\label{lem:sequences_in_cone}
    Let $(s_i^n)_{n\in\N}$ with $s_i^n \in \pcone\setminus{0}$ for $i\in[m]$. 
    Assume $\left(\sum_{i\in[m]} s_i^n\right)_{n\in\N}$ is a convergent sequence. 
    Then each sequence $(s_i^n)_{n\in\N}$ for $i\in[m]$ has a convergent
    subsequence.
\end{lemma}
\begin{proof}
    If each of the $m$ sequences $(s_i^n)_{n\in\N}$ is bounded the proof is
    trivial. Assume that for $1\leq i\leq l$  the sequences
    $(s_i^n)_{n\in\N}$ are unbounded. We now show there exists a linear
    functional $z^*$ such that with $x_n\coloneqq \sum_{i=1}^m s_i^n$,  
    \begin{gather}
        \underbrace{\inner{z^*;x_n}}_{\text{bounded}} = \underbrace{\inner{z^*;s^n_1} + \dots + \inner{z^*;s^n_l}}_{\text{unbounded}} + \underbrace{\inner{z^*;s^n_{l+1}}+ \dots + \inner{z^*;s^n_{m}}}_{\text{bounded}},\label{eq:craziness}
    \end{gather}
    producing a contradiction.
    
    To show the existence of $z^*$ define the set
    \begin{gather}
	    U\coloneqq\cl\conv\left(\union_{i\in[m]}
	    \set{\varfrac{s_i^n}{\norm{s_i^n}};
	    n\in\N}\right)\subset\pcone.
    \end{gather}
    Observe that since $\pcone$ is salient, closed and convex, we have 
    $U\cap \single{0}=\varnothing$. Furthermore, we trivially have that
    $\single{0}$ is compact and $U$ is closed.
    Then by the Hahn--Banach separation theorem 
    \citep[Theorem~1.79, p.~55]{Penot2012Calculus} there exists 
    $z^*\in\bnch^*$ such that $\inner{z^*;u}\geq\delta>0$ for all $u\in U$.
    To see why the first $l$ terms must be unbounded with this choice 
    of $z^*$ note that we can write
    \begin{gather}
        \forall{i\in[l]},\ 
        \inner{z^*;s_i^n} = \inner{z^*;\norm{s_i^n}\varfrac{s_i^n}{\norm{s_i^n}}} = \norm{s_i^n}\inner{z^*;\varfrac{s_i^n}{\norm{s_i^n}}},
    \end{gather}
    where $\norm{s_i^n}\to\infty$ and
    $\inner{z^*;\varfrac{s_i^n}{\norm{s_i^n}}}\geq\delta$ for every
    $n\in\N$ since $\varfrac{s_i^n}{\norm{s_i^n}}\in U$. This shows 
    \eqref{eq:craziness}, which is absurd.
\end{proof}

\subsection{Comparing the Convex and Concave Versions}
Table \ref{table:comparison} tabulates the convex and concave versions of
the key mathematical objects we make use of in this paper.   
As can be seen, for every
standard convex version, there is a corresponding concave version.
\begin{table}[t]
	\small
	    \arrayrulecolor{lightgray}
    \centering
    \begin{tabular}{r C{6.5em} C{6.5em} l}\toprule
         & Convex case $R\in\rdnt(\bnch)$ $f$~convex & 
	    Concave case $S\in\shdy(\bnch)$ $f$~concave & Definition
	    \\ \midrule
         Support function & $\cxsprt_R$ & $\cvsprt_R$ &
		 \eqref{eq:convex-support-function-def},
		 \eqref{eq:concave-support-function-def}\\
         Gauge function  & $\gauge_R$ & $\agauge_S$ &
		 \eqref{eq:gauge_defn}, \eqref{eq:agauge_defn} \\
		 Conjugate function  & $f^*$ & $f_*$ &
		 \eqref{eq:convex_conjugate}, \eqref{eq:concave_conjugate} \\
         Polar set & $R^\polar$ & $S^\apolar$
		 &\eqref{eq:polar_antipolar_dfn}\\
         Polar function & $f^\polar$ & $f^\apolar$
		 &\eqref{eq:function-polar-def}\\
         \bottomrule
    \end{tabular}
    \caption{Summary of the convex case and the (less known) concave case of
    support functions, gauge functions, polar sets, polar functions and
    Legendre-Fenchel conjugates.}
    \label{table:comparison}
\end{table}

\section{Loss Functions} 
\label{sec:loss_functions}

In this section we will introduce proper losses; first in the traditional
way, and then in terms of the superprediction set. We will then show some
of the implications of the latter approach. 
A loss function is an ``outcome contingent disutility'': that is, for a given
outcome $y$, it provides a measure of (dis)utility of a prediction as a
function $-u(\cdot,y)$ \citep{Berger1985}.
We introduce loss functions more formally by first introducing some
concepts from statistical decision theory, to which we apply some of the
geometric concepts introduced in \S\ref{sec:preliminaries}. Let $\Def{\rv
Z}$, $\Def{\rv Y}$ be random variables taking values in the spaces
$\Def{\cal Z}$ and $\Def{\cal Y}$. We assume $\cal Y$ is finite with
$\Def{n}{\abs{\cal Y}}$, and therefore distributions that assign
probability to every state of $\cal Y$ are isomorphic to probability
vectors from $\Def{\probm}{\relint(\set{p\in\Rp^n; \textstyle\sum_{i=1}^n
p_i = 1})},$ the relative interior of the \Def{$n$-simplex}, that is
$\cl(\probm) \simeq \smplx(\cal Y)$. Its dual cone, $\Rp^n$, is the
associated collection of loss vectors. We use $\Def{\pcone}\subseq\R^n$ and
$\Def{\fcone}\subseq\R^n$ to denote an arbitrary pair of salient, closed,
convex cones, dual to one another. The reader might find it helpful to
identify $\pcone$ with $\cl\cone(\probm)$, and $\fcone$ with its dual,
however most of our theorems merely depend on a dual pair of closed convex
cones.

One can understand the effect of choice of loss in terms of the ``conditional
perspective'' which allows one to ignore the distribution of $\rv Z$, which
is typically unknown.\footnote{See \citep{Steinwart2008,Reid2011} for a
discussion of this conditional perspective.} We call mappings $\ell\colon
\probm\times\cal Y\to\Rp$ \Def{loss functions}, and $\ell(p;\rv y)$  is
the penalty from predicting $p\in\probm$ upon observing $\rv y \in \cal Y$.
(It will sometimes be convenient to consider the extension of $\ell$
defined as $\ell\colon\cl(\probm)\times\cal Y\rightarrow\Rx_{\ge 0}$, which
now needs to map to $\Rx_{\ge 0}$ to allow for infinite values;
see Remark \ref{rem:bounded-to-unbounded}). It will be
convenient to stack loss functions into a function over the second
argument:
\begin{gather}
   \forall{p\in\probm},\  \Def{\ell(p)}{\rv y \mapsto \ell(p; \rv y)} \in \Rp^{\cal Y},
\end{gather}
or equivalently we have a vector
\begin{gather}
   \forall{p\in\probm},\  \ell(p) \simeq \g(\ell(p; \rv y_1),\dots,\ell(p; \rv
   y_n))' \in \Rp^n.
\end{gather}
For each fixed $\rv y\in \cal Y$, the functions $\ell(\marg;\rv y)$ are
called \Def{partial losses}. If for some norm (the choice does not
matter) $\|\ell(p)\|<\infty$ for all $p\in\probm$ we say the loss is
\Def{bounded}.

The \Def{conditional risk associated with $\ell$}  is defined via
\begin{gather}
    L\colon\probm\times\probm\ni (p,q)\mapsto \Def{L(p,q)}{\E_{p}[\ell(q)]} = \inner{\ell(q);p} \in\Rp.
\end{gather}
The \Def{conditional Bayes risk} $\Def{\minL}{\probm\ni p \mapsto
	\inf_{q\in\probm} L(p,q)}$ is always concave. For the next two
	definitions let $f\colon\cal Z \to \probm$, and $g(\rv z) \coloneqq
	\Pr(\rv Y| \rv Z=\rv z)$. That is, for $\rv z \in \cal Z$, $f(\rv
	z)$ is a distribution over $\rv Y$ conditioned on $\rv Z = \rv z$
	and $g(\rv Z)$ is the true conditional distribution. The \Def{full
	risk} is
\begin{gather}
    \E_{\rv Z} \E_{\rv Y \vert \rv Z}[\ell\circ f(\rv Z)] = \E_{\rv Z} \inner{\ell\circ f(\rv Z); g(\rv Z)}.\label{eq:full_risk_defn}
\end{gather}
The most general framing of a supervised machine learning problem is to
minimise \eqref{eq:full_risk_defn} by choosing an appropriate function $f$.
If we fix $\ell$ and $g$, the minimal value of the full risk
\eqref{eq:full_risk_defn} is bounded below by the \Def{Bayes
risk}\footnote{See \citep{Williamson:2022aa} for a further discussion of
	the Bayes risk, its generalisation to restricted model classes
	$\mathcal{F}$, and the relationship to measures of information such as
	$f$-divergences and integral probability metrics.}:
\begin{gather}
    \inf_{f\colon\cal X \to \probm}\E_{\rv Z} \inner{\ell\circ f(\rv Z); g(\rv Z)}.\label{eq:bayes_risk_defn}
\end{gather}

The \Def{superprediction set} \citep{KalnishkanVovkVyugin2004,
KalnishkanVyugin2002:colt, Dawid2007} of a loss function
$\ell\colon\probm\to\Rp^n$ is
\begin{gather}
    \Def{\super(\ell)}{\bigcup_{l\in\ell(\probm)}\set{x\in \R^n ;  x \succeq_{\Rp^n} l}}\subseq\Rp^n.
    \label{eq:superprediction}
\end{gather}

The set $\super(\ell)$ (we write $\super \ell$ when there is no ambiguity)
consists of all the points $x$ that incur no less loss than some point $l
\in\ell(\probm)$. In the parlance of game theory, this is the union of the
points $p$ that are weakly dominated by some other point, together with the
dominating points.  Equivalently \eqref{eq:superprediction} can be written
\begin{gather}
    \super \ell = \union_{l\in{\ell(\probm)}}(l + \Rp^n) = 
    \ell(\probm) + \Rp^n.\label{eq:super2}
\end{gather}
In the next section we will be interested in the closed convex hull of the
superprediction set mapping, for which it is useful to note 
\begin{gather}
    \cl\conv(\super\ell) \overset{\eqref{eq:super2}}{=} \cl\conv(\ell(\probm) + \Rp^n) = \cl\conv(\ell(\probm)) + \Rp^n = \union_{l\in{\cl\conv(\ell(\probm))}}(l + \Rp^n).
    \label{eq:super3}
\end{gather}


Lemma \ref{thm:choquet_closure} is a special case of a result due to
\citet{choquet1962cones}, but as the original is in French we include a
proof for our setting below.
\begin{lemma}\label{thm:choquet_closure}
    Let $S_1,\dots,S_m \subseq\pcone\setminus{0}$ each be closed. Then the set $S_1 + \dots + S_m$ is closed.
\end{lemma}
\begin{proof}
    Let $S\coloneqq S_1 + \dots + S_m$, and take a convergent sequence
    $(x_n)_{n\in\N}\to x$ with $x_n \in S$. Then for each $x_n$ there
    exists $(s_i^n)_{i\in[m]}$ with $s_i^n \in S_i$ for $i\in[m]$ and
    $n\in\N$. By Lemma \ref{lem:sequences_in_cone} each of the sequences $(s_i^n)_{n\in\N}$ has a convergent subsequence. And as $S_i$ is closed $\lim_{n\to\infty} s_i^n \in S$ for each $i\in[m]$ and 
    $x = \sum_{i\in[m]}\lim_{n\to\infty} s_i^n\in S$ (taking subsequences
    if need be).
\end{proof}

\subsection{Proper Losses} 
\label{sub:the_family_of_proper_losses}

A natural requirement to impose upon $\ell$ is that it is
\Def{proper}\footnote{See \citep{GneitingRaftery2007,Reid2011} for  
an elaboration of the notion of properness, which dates back at least to
work of \citet{Shuford:1966to} and \citet{Holstein1970};  early particular
examples are due to \citet{Brier:1950aa} and \citet{Good:1952aa}.}
\citep{Hendrickson1971}, which means that 
\begin{gather}
    \left[\forall{p,q\in\probm},\ \inner{\ell(q);q}≤
    \inner{\ell(p);q}\right]
    \iff \left[\forall{q\in\probm},\  
    q\in\arginf_{p\in\probm}\inner{\ell(p);q}\right].\label{eq:proper_defn}
\end{gather}
That is, predicting the true probability minimises the expected loss.  We
say $\ell$ is \Def{strictly proper} when the above inequality is strict for
$p≠q$, that is, $\single{q}=\arginf_{p\in\probm}\inner{\ell(p);q}$ for all
$q\in\probm$.  If $\ell$ is proper, $\minL(p)=L(p,p)=\inner{\ell(p);p}$. 
The superprediction set $\super(\ell)$ of a proper loss $\ell$ has some
useful properties: Theorem \ref{thm:representation} makes explicit the
link between the superprediction set and the convex geometry in
\S\ref{sec:preliminaries}. Proposition \ref{prop:subdifferential_representation}
below
justifies that from a superprediction set we can construct a loss function.
This motivates a shift in focus of analysis from loss
functions $\ell$  to families of convex superprediction sets of proper
losses.

\begin{theorem}[Representation]\label{thm:representation}
    Let $\ell\colon\probm\to\R^n$ be a loss function with the associated 
    conditional Bayes risk $\minL$. 
    Then 
    \begin{enumerate}
	    \item $\minL = {\cvsprt_{\cl\conv(\super\ell)}}_{|\probm}$ (the
    restriction of $\cvsprt_{\cl\conv(\super\ell)}$ to $\probm$); 
    \item $ [\forall{p\in\probm},\ \ell(p) \in 
	    \subdiff \cvsprt_{\cl\conv(\super\ell)}(p)] 
	    \ \Longleftrightarrow\ \ell\ \mbox{is proper}.$
	\end{enumerate}
\end{theorem}
\begin{proof}
    The claim that $\minL = \cvsprt_{\cl\conv(\super\ell)}$ on $\probm$ for all loss functions $\ell$ is straightforward:
    \begin{gather}
    \forall{p\in\probm},\ 
        \minL(p) 
        = \inf_{q\in\probm} \inner{\ell(q);p}
        = \inf_{l\in\overline{\ell(\probm)}} \inner{l;p}
        = \inf_{l \in\overline{\ell(\probm)} + \Rp^n} \inner{l;p}
        \overset{\eqref{eq:super3}}{=} \cvsprt_{\cl\conv(\super\ell)}(p).\label{eq:cond_bayes_risk_rep}
    \end{gather}
    We now prove the second claim.
    Assume $\ell$ is proper. Then for $p\in \probm$  we have
    $\cvsprt_{\super(\ell)}(p) = L(p,p) = \inner{\ell(p);p}$ and
    $\forall{p\in\probm},\ \forall q\in\R^n$, 
    \begin{align}
	    \inner{\ell(p);q} ≥ \inner{\ell(q);q}
        &\iff \inner{\ell(p);q} - \inner{\ell(p);p} ≥  \inner{\ell(q);q}- \inner{\ell(p);p}\\
        &\overset{\eqref{eq:cond_bayes_risk_rep}}{\iff} \inner{\ell(p);q-p} 
	    ≥ \cvsprt_{\cl\conv(\super\ell)}(q) - \cvsprt_{\cl\conv(\super\ell)}(p).
    \end{align}
    Thus $\ell(p)\in \subdiff\cvsprt_{\cl\conv(\super\ell)}(p)$ for $p\in\probm$.
    
     We use a proof by contradiction to show the reverse implication. Assume $\ell$ enjoys the subgradient representation $\ell(p)\in\subdiff\cvsprt_{\cl\conv(\super\ell)}(p)$ for $p\in\probm$, but is not proper. Since $\ell$ is not proper, by \eqref{eq:proper_defn} there exists $p,q\in\probm$ with
    \begin{alignat}{2}
        \inner{\ell(p);q} < \inner{\ell(q);q}
        &\iff\ \ & \inner{\ell(p);q} - \inner{\ell(p);p} &< \inner{\ell(q);q} - \inner{\ell(p);p}\\
        &\iff\ \ & \inner{\ell(p);q-p} &< \inner{\ell(q);q} - \inner{\ell(p);p}.\label{eq:proper_just_1}
    \end{alignat}
    Since $\ell(p')\in\subdiff\cvsprt_{\cl\conv(\super\ell)}(p')$ for $p'\in\probm$, from \eqref{eq:superdiff_defn} we have
        $\forall{p',q'\in\Rp^n}$,
    \begin{alignat}{1}
	&
        \inner{\ell(p');q'-p'} ≥ \cvsprt_{\cl\conv(\super\ell)}(q') - 
	     \cvsprt_{\cl\conv(\super\ell)}(p')\\
            \mathrel{\mathrslap{\implies}{\iff}}\ \ &
            -\inner{\ell(p');p'} ≥- \cvsprt_{\cl\conv(\super\ell)}(p')  \\
            \iff\ \  & 
             \inner{\ell(p');p'} ≤ \cvsprt_{\cl\conv(\super\ell)}(p'),\label{eq:proper_just_2}
    \end{alignat}
    where in the implication we take $q'=0$. This gives us, for our choice
    of $p,q$,
    \begin{align}
        \inner{\ell(p);q-p} 
        &\overset{\eqref{eq:proper_just_1}}{<} \inner{\ell(q);q} - \inner{\ell(p);p} \\
        &\overset{\eqref{eq:proper_just_2}}{≤} \cvsprt_{\cl\conv(\super\ell)}(q) - \inner{\ell(p);p}\\
        &\overset{\hphantom{\eqref{eq:proper_just_2}}}{≤}\cvsprt_{\cl\conv(\super\ell)}(q) - \inf_{r\in\smplx}\inner{\ell(r);p}\\
        &\overset{\smash{\eqref{eq:cond_bayes_risk_rep}}}{=}
                \cvsprt_{\cl\conv(\super\ell)}(q) - \cvsprt_{\cl\conv(\super\ell)}(p),
    \end{align}
    which contradicts our assumption that
    $\ell(p)\in\subdiff\cvsprt_{\cl\conv(\super\ell)}(p)$ for all $p\in\probm$.
\end{proof}

Thus in order to build a geometry of loss functions in terms of convex sets, 
with Theorem \ref{thm:representation} we see that the propriety condition
of the losses cannot be discarded; see also the discussion in section 
\ref{sec:naturalness} below.

The definition of $\probm$ as the relative interior of the probability
simplex guarantees (via
Corollary \ref{cor:loss_function_domain}(\ref{cor:subdiffdomain}) below) that the
subdifferential $\subdiff \cvsprt_{\cl\conv(\super\ell)}(p)$ is nonempty
for all $p\in \probm$. This is analogous to the differentiable case, where
if one wishes to compute gradients of a differentiable function, the
natural area of analysis is the interior of its domain of definition. 

\begin{proposition}\label{prop:superc}
    Suppose $\ell\colon\probm\to\Rp^n$ is a loss function. Then $\cl\conv(\super\ell)$ is $\Rp^n$-oriented.
\end{proposition}
\begin{proof}
    By hypothesis $\ell\colon\probm\to\R^n$ and so from the definition of
    the superprediction set $\super(\ell)\subseq \Rp^n$ we have
    \begin{align}
        \rec(\Rp^n)
	& \!\!\!\overset{\mathrm{P}\ref{prop:rccalc} 
	(\ref{prop:rccalc_subset})}{\supseq}
		\rec(\cl\conv(\super\ell))\\
	& \!\overset{\eqref{eq:super3}}{=}
		\rec\left(\union_{l\in{\cl\conv(\ell(\probm))}}(l +
		\Rp^n)\right)\\
		&
		\!\!\!\overset{\mathrm{P}\ref{prop:rccalc}(\ref{prop:rccalc_union})}{\supseq} 
		\union_{l\in\cl\conv(\ell(\probm))}\rec(l + \Rp^n) \\
	& = \rec(\Rp^n)\\
	&\!\!\!\!\overset{\mathrm{P}\ref{prop:rccalc}(\ref{prop:rccalc_cone})}{=} \Rp^n
    \end{align}
    as desired.
\end{proof}
\begin{corollary}\label{cor:loss_function_domain}
        Suppose $\ell\colon\probm\to\Rp^n$ is a loss function. Then 
	\begin{enumerate}
        \item $\dom(\cvsprt_{\cl\conv(\super\ell)}) \supseq \Rp^n$, and
		\label{cor:bayesdomain}
        \item $\dom(\subdiff\cvsprt_{\cl\conv(\super\ell)})\supseq\Rpp^n$.
		\label{cor:subdiffdomain}
    \end{enumerate}
\end{corollary}
\begin{proof}
    Take Proposition \ref{prop:superc} and apply Lemma
    \ref{lem:sprt_domain}, this shows claim \ref{cor:bayesdomain}. Apply
    Lemma \ref{prop:subdiff_nonempty} to \eqref{cor:bayesdomain} to show 
    claim \ref{cor:subdiffdomain}.
\end{proof}
Theorem \ref{thm:representation} and Proposition \ref{prop:superc} motivate
the introduction of the following family of convex sets. (Recall that
$\bnch=\R^n$ for some $n$ and $X_+\subset\bnch$ denotes a salient closed
convex cone.) Take
$C\subseq\bnch$, let $\prop(C)$ be the collection of $C$-oriented, closed,
nonempty convex subsets of $C\setminus{0}$:
\begin{gather}
    \Def{\prop(C)}{\set{S \in \cvx(C\setminus{0}) ; 
    \rec(S) = C}}.\label{eq:prop_defn}
\end{gather}
The construction \eqref{eq:prop_defn} admits a lot of structure. In
particular, Lemma \ref{lem:prop_shady} justifies our interest in the shady
sets introduced in \S\ref{sec:gauge_functions_and_polar_duality}.
Recall that $\shdy(\bnch)$ denotes the collection of closed shady subsets
of $\bnch$.
\begin{lemma}\label{lem:prop_shady}
     $\prop(\pcone)\subseq\shdy(\bnch)$.
\end{lemma}
\begin{proof}
    Take $S\in\prop(\pcone)$. Since $S\subseq\rec(S)=\pcone$, we have 
    \begin{align}
	    \left(\forall{d\in \pcone},\  \forall{\alpha>0},\
	    S + \alpha d \subseq S \right)
        \implies&
        \left(\forall{d\in S},\ \forall{\alpha>0},\  S + \alpha d \subseq
		S\right)\\
        \implies& 
        \left(\forall{\alpha≥1},\  \alpha S \subseq S\right),    
    \end{align}
    and $S$ is co-star-shaped. From \eqref{eq:prop_defn}, $S$ is 
    also convex, thus $S$ is shady.
\end{proof}

\begin{lemma}\label{lem:nomempty_relint}
    Let $S\in\prop(\pcone)$. Then $\relint(S)≠\varnothing$.
\end{lemma}
\begin{proof}
    The result follows from a proof by contradiction. Suppose
    $\relint(S)=\varnothing$. Then from \eqref{eq:relint_defn} this means
    \begin{gather}
        \forall{x\in S},\ \forall{\lambda > 1},\ \exists{y\in S},\  
        \left[ \lambda x + (1-\lambda) y\notin S 
        \iff 
	\lambda x \notin S +  (\lambda- 1) y\right]. \label{eq:nomempty_relint}
    \end{gather}
    From the conditions, $\lambda - 1 > 0$ and $y\in S\subseq\rec(S)$. Thus
    $(\lambda -1) y \in \rec(S)$. It follows that $S+(\lambda - 1) y = S$
    because for any $x\in\rec S$, $S+x=S$ (Proposition 
    \ref{prop:rccalc} (2)).
    Therefore \eqref{eq:nomempty_relint} reduces to  
    \begin{gather}
        \forall{x\in S},\ \forall{\lambda > 1},\ 
        \lambda x \notin S,
    \end{gather}
    which contradicts the fact that $S$ is co-star-shaped (Lemma
    \ref{lem:prop_shady}).
\end{proof}
\begin{lemma}\label{lem:prop_intersect}
    Let $A_i\in\prop(\pcone)$ for $i\in[m]$. Then
    $\inter_{i\in[m]}A_i≠\varnothing.$
\end{lemma}
\begin{proof}
    Take an arbitrary sequence $(a_i)_{i\in[m]}$ with $a_i\in A_i$. Let
    $J_i \coloneqq [m]\setminus{i}$, for $i\in[m]$. Then $\sum_{j\in
    J_i}a_j + a_i = \sum_{k\in [m]}a_k$ for each $i\in[m]$. Since $\pcone$
    is a cone $\sum_{j\in J_i}a_j\in \pcone$ for each $i\in[m]$. By
    assumption $A_i\subseq\rec(A_i) = \pcone$, and from the definition of
    the recession cone \eqref{eq:rec_cone_defn} we always have $\sum_{k\in
    [m]} a_k = a_i + \sum_{j\in J_i}a_j \in  a_i + \pcone \subseq A_i$ for
    all $i\in[m]$, and thus $\inter_{i\in[m]} A_i \neq \varnothing$.
\end{proof}

Proposition \ref{prop:subdifferential_representation} completes the
connection between superprediction sets and loss functions.

\begin{proposition}\label{prop:subdifferential_representation}
    Take $S\in\prop(\Rp^n)$. There exists a 0-homogeneous selection
    $\ell\colon\Rpp^n\to\Rp^n$ of $\subdiff\cvsprt_S$ in the sense that 
    \begin{gather}
        \forall{p\in\Rpp^n},\ 
        \ell(p)\in\subdiff\cvsprt_S(p),\label{eq:subgrad_rep}
    \end{gather}
    and $\ell$ restricted to $\probm$ is a proper loss.
\end{proposition}
\begin{proof}
    From Lemma \ref{prop:subdiff_nonempty},
    $\dom(\subdiff\cvsprt_S)\supseq\relint(\Rp^n)=\Rpp^n$, and so there
    exists $\ell(p)\in\displaystyle\arginf_{\smash{x\in S}}\inner{x;p}$ for
    $p\in\Rpp^n$. Thus
    \begin{gather}
        \forall{p,q\in\probm\subseq\Rpp^n},\ 
        \inner*{\ell(q);q}
        ≤
        \inner*{\ell(p);q},\label{eq:other}
    \end{gather}
    and $\ell$ is proper. Since $\cvsprt_S$ is 1-homogeneous, $\ell$ is
    0-homogeneous (Proposition \ref{prop:subdiff-zero-homog}).
\end{proof}
\vspace*{-4mm}
 \begin{remark}[From bounded to unbounded]\label{rem:from_bounded_to_unbounded}
	 \normalfont
	 \label{rem:bounded-to-unbounded}
     In Theorem \ref{thm:representation} and Proposition
     \ref{prop:subdifferential_representation} we needed to be careful when
     talking about the domain of definition of a loss function $\ell$.
     This was to ensure that $\subdiff \cvsprt_{\super(\ell)}$ is nonempty in
     order to have the inclusion $\ell(p)\in\subdiff
     \cvsprt_{\super(\ell)}(p)$. Recall our definition of $\probm$ as the
     relative interior of the standard simplex.  In practice since if there
     exists $q\in\cl(\probm)$ with $\subdiff \cvsprt_{\super(\ell)}(q) =
     \varnothing$, we can define $\ell(q;\rv y)\coloneqq \lim_{(p_n)\to
     q}\ell(p_n;\rv y)$, where the sequence $(p_n)$ is chosen with $p_n \in
     \probm$ and by Lemma \ref{prop:subdiff_nonempty} we know we have
     $\subdiff\cvsprt_{\super(\ell)}(p_n)≠\varnothing$ (allowing us to take
     $\ell(p_n) \in \subdiff\cvsprt_{\super(\ell)}(p_n)$). This extends
     our loss function to a mapping
     $\ell\colon\cl(\probm)\to\Rx_{\ge 0}^n$.  Many of the results on
     bounded loss functions can be similarly extended to unbounded loss
     functions, by restricting analysis to $\probm$; for example the range
     of log loss is $\Rx_{\ge 0}^n$ when defined on $\cl(\probm)$ but
     $\R_{\ge 0}^n$ when restricted to
     $\probm$.  
 \end{remark}
In light of Remark \ref{rem:from_bounded_to_unbounded}, the distinction
between bounded and unbounded losses becomes less important. 
Some commonly used loss functions are listed in Table
\ref{tab:common_losses} along with their boundedness. 
\begin{table}
	    \arrayrulecolor{lightgray}
    \centering
    \small
    \begin{tabular}{L{2.5cm} l l l c c c} \toprule
	    {\scriptsize Name} &  & $\ell(p;\rv y)$ & $\minL(p)$ & {\scriptsize Propriety}  & 
	    {\scriptsize Bounded} &  {\scriptsize Ref.}\\ \midrule
0/1 
            & $\ooneloss$   
	    & $\co\{\eul^j\colon\! j\in \displaystyle\argmax_{\rv y\in\cal
	    Y} p_{\!\rv y}\}$
            & $\displaystyle\min_{\rv y \in \cal Y} p_{\rv y}$ 
            & proper 
	    & $\bullet$ 
	    & \S\ref{sub:the_family_of_proper_losses} \\[3mm] 
Logarithmic & $\logloss$    & $ - \log(p_{\rv y})$        
            & $\!\!\!\!\!\!-\displaystyle\sum_{\rv y \in\cal Y}\!\! p_{\rv y} \log p_{\rv y} $ 
            & strict 
	    & 
	    & \S\ref{sub:the_polar_loss}\\[0mm] 
Concave Norm {\scriptsize $a\in[-\infty, 1]\setminus{0}$}
            & $\ell_a$ & $\g(\displaystyle\frac{p_{\rv
	    y}}{\beta_{\!\!\frac{a}{a-1}}(p)})^{\frac{1}{a-1}}$
	    & $\beta_{\!\!\frac{a}{a-1}}(p)$ 
	    & strict
	    & 
	    & \S\ref{ssec:concave_norm_losses} \\[5mm]
Brier       & $\brierloss$  & $(1 + \norm2{p}^2)·1_n - 2p_{\rv y}$   
            & $1-\norm2{p}^2$ 
            & strict
	    & $\bullet$
	    & \S\ref{ssec:brier_loss} \\[3mm]
Cobb--Douglas {\scriptsize $a\in\Rp^n$}
            & $\cdloss$     & $\psi_a\g(\varfrac{p}{a})·\varfrac{a_{\rv y}}{p_{\rv y}}$ 
            & $\psi_a(p)$
            & strict 
	    &
            & \S\ref{ssec:cobb_douglas} \\ 
\bottomrule
    \end{tabular}
    \caption{Some common loss functions, their conditional Bayes risks, and
    their propriety and boundedness.}
    \label{tab:common_losses}
\end{table}

Corollary \ref{por:strictly_convex_proper} below follows from the proof of
Proposition \ref{prop:subdifferential_representation} together with
Corollary \ref{cor:cvsprt_differentiable}.
\begin{corollary}\label{por:strictly_convex_proper}
A loss function $\ell$ is strictly proper if and only if
$\cl\conv(\super\ell)$ is strictly convex.
\end{corollary}
\begin{corollary}\label{cor:bijection}
There is a bijection between the equivalence class of loss functions
$\ell\colon\probm\to\Rp^n$ which agree almost everywhere and the family of convex
sets $\prop(\Rp^n)$.
\end{corollary}
\begin{proof}
    There is a bijection between superlinear functions $\cvsprt_S$ and
    closed convex sets $S$ \citep[Theorem~C.2.2.2]{hiriarturruty2001fca},
    and the mapping $\subdiff\cvsprt_S$ is a singleton almost everywhere
    \citep[Theorem~B.4.2.3]{hiriarturruty2001fca}.  The connection to loss
    functions (Theorem \ref{thm:representation} and Proposition
    \ref{prop:subdifferential_representation}) completes the proof.
\end{proof}

\subsection{Starting with Sets}
The above development
motivates the key viewpoint of the present paper: \emph{start with a
set $S\in\prop(\Rp^n)$ and derive the loss (and other quantities) from it}.
We will thus sometimes explicitly parametrise the loss theoretic functions
as $\ell_S$, $L_S$ and $\minL_S$.  One immediate consequence of
using a set $\super(\ell)\in\prop(\Rp^n)$ to define a proper loss $\ell$ is
that it may be the case that for two different loss functions $\ell≠\emm$
we have $\cl\conv(\super\ell) = \cl\conv(\super\emm)$. This is the case
whenever the conditional Bayes risk functions for $\ell$ and $\emm$
coincide. However in such cases $\ell$ and $\emm$ differ only on a set of
measure zero \citep[cf.][Proposition~8]{Vernet:2016aa}. That is, for some
$S\coloneqq\super(\ell)=\super(\emm)$, both $\ell$ and $\emm$ satisfy
\eqref{eq:subgrad_rep}. However when $\ell$ is strictly proper,
$\super(\ell)$ is strictly convex and so for strictly convex
$\super(\ell)=\super(\emm)$ we always have $\ell =\emm$.

\begin{remark}[Misclassification loss]
	\normalfont
Misclassification loss $\ooneloss$ (also called $0/1$ loss)
\citep{Buja:2005,GneitingRaftery2007} assigns zero loss when predicting
correctly and a loss of $1$ when predicting incorrectly. This can be
extended to when one predicts with a distribution $p\in\Delta$, with
$\ooneloss(p)=\eul^j$ where $j=\argmax_{i\in[n]} p_i$ and $\eul^j$ is the
$j$th canonical basis vector in $\R^n$, under the assumption that
$\{p_i\colon i\in[n]\}$ has a unique maximum.  We can extend $\ooneloss$ to
all of $\probm$ in a manner that is consistent with Theorem
\ref{thm:representation} as follows: Define 
\begin{gather}
	\Def{\minL_{0/1}}\colon
	\Delta\ni p\mapsto \min_{i\in[n]}p_i,
\end{gather}
and let $\Def{\ooneloss(p)}{\subdiff^\prime\minL_{0/1}(p)}
=-\subdiff_\prime(-\minL_{0/1}(p)) =
-\subdiff_\prime(-\min_{i} p_i)=-\subdiff_\prime(\max_i (-p_i))$.
Using \citep[Example 3.4, page 182]{hiriarturruty2001fca} we have
\begin{gather}
\ooneloss(p)= -\subdiff_\prime(-\minL(p))= 
-\co\{-\eul^j\colon j\in\argmax_{i\in[n]} p_i\}= 
\co\{\eul^j\colon j\in \argmax_{i\in[n]}
p_i\}.
        \label{eq:oone_loss_defn}
\end{gather}
When the argmax is unique, this reduces to just $\eul^j$ as per the usual
definition.
\end{remark}

\begin{remark}[Naturally 0-Homogeneous]\label{rem:0_homogeneous}
	\normalfont
In Theorem \ref{thm:representation} we needed to make restrictions on the
domain of the concave support function and subdifferential in order for the
geometric functions from convex analysis to agree with the classical
concept of a loss function. However when loss functions are viewed instead
as the subgradient of a concave support function these restrictions are
indeed arbitrary. The subgradient representation of a proper loss function
$\eqref{eq:subgrad_rep}$ suggests via the 0-homogeneity result of
Proposition \ref{prop:subdifferential_representation} that it is more natural to
take a proper loss function $\ell$ defined in the conventional way on
$\probm$ and instead consider its 0-homogeneous extension:
$\Rpp^n\ni p\mapsto\ell'(p)\coloneqq\ell\g\big(p·\oneon{\norm1{p}})\in\Rp^n$. 
Defined like this,
$\ell'$ satisfies \eqref{eq:subgrad_rep} and agrees with $\ell$ on
$\probm$. Without loss of generality, for the remainder of our analysis we
will refer to a \emph{loss function} as such a 0-homogeneous mapping
$\Rpp^n\to\Rp^n$.\footnote{%
	\label{footnote:cosmic}
	A formal alternative would be to work with the
	{horizon} $\hzn\R^n$ of {directions}
	$\dir\colon\R^n\rightarrow\hzn\R^n$ which can be considered 
	pure (magnitudeless) direction vectors, so that for 
	$\alpha>0$, $\dir(\alpha x)=\dir(x)$
	\citep[Chapter 3]{RockafellarWets2004}. It amounts to
	the same thing since 0-homogeneity of $\ell$ means that the
	magnitude of any vector $x\in\R^n$ does not affect the value of
	$\ell(x)$; all that matters is the direction of $x$. Thus one could
	in fact \emph{define} $\tilde{\ell}\colon\hzn\Rp^n\rightarrow\Rp^n$ 
	via $\tilde{\ell}(\tilde{x})=\ell({x})$,
	where $x$ is the unique point of intersection of the
	``infinite magnitude'' direction vector $\tilde{x}$ with the unit sphere.}
\end{remark}

\subsection{Bregman Divergences, Semi Inner Products and Finslerian Geometry} 
\label{sub:bregman_divergences}

The relationship between $\ell$ and the conditional Bayes risk $\minL$
(which we know is equal to $\cvsprt_{\super(\ell)}$) is
usually credited to \citet{Savage:1971} and is intimately related to
Bregman divergences\footnote{See \citep{Reid2011} for more
context and background on Bregman divergences.}. 
Given a convex function $\phi$, the 
\Def{Bregman divergence} between $x,y\in \dom(\phi)$ is defined to be
\begin{equation}
	\Def{\breg_\phi(x,y)} \coloneqq \phi(x)-\phi(y)-\inner{g(y);x-y},
\end{equation}
where $g\in\subdiff \phi$ is a selection of the subgradient of $\phi$.
It is known that the \Def{regret} $\Def{L(p,q)-\minL(p)}$ is a Bregman
divergence with $\phi=-\minL$, which is not only convex but is also
1-homogeneous. The additional structure of 1-homogeneity offers a nice
simplification. 
\begin{proposition}
	Give a Bayes risk $\minL$, $\forall{p,q\in\Rpp^n}$,
	\begin{equation}
		\label{eq:simple_bregman}
	   \breg_{-\minL}(p,q) = L(p,q)-\minL(p)=\inner{\ell(q) - \ell(p);p}.
	   \end{equation}
\end{proposition}
\begin{proof}
We have
$
    \forall{p,q\in\Rpp^n},\  L(p,q)=\inner{\ell(q);p},
$
and thus if $\ell$ is proper by Theorem \ref{thm:representation},
the general form of the Bregman divergence simplifies:
\begin{align*}
    \breg_{-\minL}(p,q) 
    &= -\minL(p) + \minL(q) + \inner{\ell(q);p-q}\\
    & = -\inner{\ell(p);p} + \inner{\ell(q);q} +  \inner{\ell(q);p} -
	    \inner{\ell(q);q}\\
    &= \inner{\ell(q) - \ell(p);p}\\
    &= L(p,q)-\minL(p), 
\end{align*}
where we have used the fact that since $\ell$ is proper
$\minL(p)=\inner{\ell(p);p}$.
\end{proof}
The simpler form \eqref{eq:simple_bregman}
provides for a geometrical interpretation of the Bregman divergence as the
inner product of the vectors  $(\ell(q) - \ell(p))$ and $p$, which we
illustrate in Figure \ref{figure:bregman}.

\begin{figure}[t]
    \centering
    \tikzsetnextfilename{bregman_interpretation}
    \begin{tikzpicture}
        \begin{axis}[domain=0.05:0.95, xmin=-0.5, xmax=2, ymin=-0.2, ymax=2, width=0.625\textwidth, height=0.55\textwidth, xlabel={${p_1, \ell(p;\rv y_1)}$}, ylabel={$p_2, \ell(p;{\rv y}_2)$},ylabel style={anchor=south},
        after end axis/.code={\draw[-latex, shorten >=1.5pt] (axis cs:0,0) -- (axis cs:-0.2231,0.699) node[dot] {} node[anchor=east] {$\ell(q)-\ell(p)$}; }]
            \coordinate (o) at (axis cs: 0,0);
            \path[name path=upper_axis] (axis cs:0,5) -- (axis cs:5,0);

            \addplot[set, color=Blue, name path=l] ({-ln(x)},{-ln(1-x)}) 
                node[pos=0.25, anchor=south, sloped] {$\super(\logloss)$}; 
            \addplot[fill=LightBlue] fill between[of=l and upper_axis];

            \coordinate (p) at (axis cs:0.6,0.4);
            \coordinate (p_ray) at (axis cs:6,4);
            \coordinate (lp) at (axis cs:0.5108256238,0.9162907319);
            \coordinate (q) at (axis cs:0.8,0.2);
            \coordinate (q_ray) at (axis cs:8,2);
            \coordinate (lq) at (axis cs:0.2231435513,1.6094379124);
            
            \coordinate (breg) at (axis cs:-0.2231,0.699);
            \draw[faint,shorten >=-6cm, shorten <=-5cm] (breg) -- ($(o)!(breg)!(p_ray)$);
            \draw[faint, right angle quadrant=2, right angle symbol={o}{p_ray}{breg}];

            \draw[shorten >=-6cm, shorten <=-5cm, faint] ($(o)!(lp)!(p_ray)$) -- (lp);
            \draw[shorten >=-6cm, shorten <=-5cm, faint] ($(o)!(lq)!(q_ray)$) -- (lq);
            \draw[faint, right angle quadrant=2, right angle symbol={o}{p_ray}{lp}];
            \draw[faint, right angle quadrant=2, right angle symbol={o}{q_ray}{lq}];
            
            \draw[faint] ($-1*(p_ray)$) -- (o) -- (p_ray);
            \draw[faint] ($-1*(q_ray)$) -- (o) -- (q_ray);
            \draw[-latex, shorten >=1.5pt] (o) -- (p);
            \draw[-latex, shorten >=1.5pt] (o) -- (q); 
 
            \draw[decorate,decoration={brace,raise=6pt,amplitude=2pt,mirror}]
	    (o)  -- node[anchor=north west, yshift = -9pt, prominent]
            {$\breg_{-\minL}(p,q)·\varfrac{1}{\norm2{p}}$}
            ($(o)!(breg)!(p_ray)$) ;
            
            \addplot[set, domain=0:1] ({x},{1-x}) node[pos=0.95, anchor=south west, prominent] {$\probm$};
            
            \draw (p) node[dot] {} node[anchor=south, yshift=1ex] {$p$};
            \draw (lp) node[dot] {} node[anchor=west, yshift=0ex] {$\ell(p)$};
            \draw (q) node[dot] {} node[anchor=south, yshift=1ex] {$q$};
            \draw (lq) node[dot] {} node[anchor=west, yshift=0ex] {$\ell(q)$};
        \end{axis}
        \draw (o) node[anchor = north east] {$0$};
        \draw (o) node[anchor = north east] {$0$};
    \end{tikzpicture}
    \caption{Geometrical interpretation of regret as the Bregman divergence $\breg_{-\minL}(p,q)=\inner{\ell(q) - \ell(p);p}$. As $q\to p$, so $\ell(q)\to\ell(p)$ and the vectors $\ell(q)-\ell(p)$ and $p$ become orthogonal and $\breg_{-\minL}(p,q)\to 0$.
    \label{figure:bregman}}
\end{figure}

Considering loss functions as subgradients of concave support functions
provides an intriguing geometrical perspective which we now sketch. It is
based upon existing work on norm derivatives \citep{Alsina:2010aa} which we first
summarise.  Given a norm $\|\cdot\|$ on some vector space $V$, define  the
\Def{normalised norm derivative} for $x,y\in V$ via
\[
	{\tau'(x,y)} \coloneqq \lim_{\lambda\rightarrow 0} 
	\frac{\|x+\lambda y\|^2-\|x\|^2}{2\lambda} =
	\|x\|\lim_{\lambda\rightarrow 0} \frac{\|x+\lambda
	y\|-\|x\|}{\lambda} .
\]
The function $\tau'$ is of interest because if the norm is derived from an
inner product via $\|\cdot\|=\langle\cdot,\cdot\rangle^{\frac{1}{2}}$, then
$\tau'(x,y)= \langle x,y\rangle$. For norms that are \emph{not} derived
from an inner product, the normalised norm derivative can be claimed to be
``like'' an inner product. This claim can be formalised as follows.  First
we slightly generalise  $\tau'$ by writing it in terms of a gauge function
$\gamma$ with associated unit ball $S^\polar$ in $V$ which we henceforth
take to be $\R^n$ (recall every norm is a gauge function,
but a gauge function is only a norm if its unit ball is centrally symmetric
with respect to the origin).  Suppose $S^\polar$ is smooth and strictly convex and
thus $S$ is too,  and hence $\sigma_S$ (and $\sigma_{S^\polar}$) is
differentible everywhere since there is only one support point for a given
hyperplane.  We can generalise the definition of  $\tau'$ as
\begin{align}
	\label{eq:alternate-tau-dash}
	\Def{\tau_S'(x,y)}  &\coloneqq\,\gamma_{S^\polar}(x) \lim_{\lambda\rightarrow 0}
	\frac{\gamma_{S^\polar} (x+\lambda y) -
	\gamma_{S^\polar}(x)}{\lambda}\\
	&= \sigma_S(x)\lim_{\lambda\rightarrow 0} \frac{\sigma_S(x+\lambda
	y)-\sigma_S(x)}{\lambda}\\
	&=\Def{\sigma_S(x)  \Ds\sigma_S(x) \cdot  y} ,
\end{align}
where $\Def{\Ds f(x)=(\partial f(x)/\partial
	x_1,\ldots,\partial f(x)/\partial {x_n})}$, the Jacobian of $f$,
	is a row vector  \citep[page 99]{Magnus:1999}.

\citet{Lumer:1961aa} introduced a \Def{semi inner product}
on a real vector space $V$ as a real-valued two-place function $[\cdot,\cdot]$ 
satisfying the following axioms\footnote{
	Semi-inner-products have been used previously in machine learning; see
	e.g.~\citep{Zhang:2009aa,Der:2007aa}.}.
\begin{description}
	\item[SIP1] $[x+y,z] = [x,z] + [y,z]\ \ \forall x,y,z\in V$.
	\item[SIP2] $[\lambda x, y] = \lambda [x,y]\ \ 
		\forall\lambda\in\R,\ \forall x,y\in V$.
	\item[SIP3] $[x,x]\ge 0$\ \  $\forall x\in V,\ x\ne 0$.
	\item[SIP4] $|[x,y]|^2 \le [x,x] [y,y]\ \ \forall x,y\in V$.
	\item[SIP5] $[x,\lambda y] = \lambda [x,y]\ \ \forall\lambda\in\R,\
		\forall x,y\in V$.
	\item[SIP6]  $[y,x+\epsilon y] \rightarrow [y,x]$ for all real
		$\epsilon\rightarrow 0$, $\forall x,y\in V$.
\end{description}
Unlike the standard inner product, $[\cdot,\cdot]$ is
\emph{not} symmetric: in general $[x,y]\ne [y,x]$. 

Suppose $0\in S$ then  $0\in S^\polar$. Suppose further that
the principal curvatures of $S$ and $S^\polar$ are all non-zero ($\ell_S$
and $\ell_{S^\apolar}$ being strictly proper guarantee that). Then 
$\sigma_S$ and $\sigma_{S^\polar}$ are in $C^2$ \citep[page 115]{Schneider2014}.
Let $\Def{G_S\coloneqq\frac{1}{2} \Hs\gamma_{S^\polar}^2}$, where $\Hs$ denotes the hessian:
for $f\colon\reals^n\rightarrow\reals$, $\Def{\Hs f(x) \coloneqq
\Ds((\Ds(x))')}$ \citep{Magnus:1999}. 
Following \citet{Giles:1967aa} (who assumed $\gamma$ was additionally a
norm, and thus symmetric, an assumption which we drop) 
for $x,y\in\reals^n$ we define
\begin{gather}
	\Def{[y,x]_S \coloneqq y'\cdot G_S(x)\cdot x}.
\end{gather}
\vspace*{-6mm}
\begin{proposition}
	Suppose $S$ satisfies the assumptions above. Then $[\cdot,\cdot]_S$ is a
	semi inner product.
\end{proposition}
\begin{proof}
Since $\sigma_S$ is convex, we have $\Hs\sigma_S$ is
positive semidefinite  (in fact it has only one
zero eigenvalue  under the assumptions above \cite[page
118]{Schneider2014}).
By properties of support functions we have that for all $x$
\[
	\Ds \gamma_{S^\polar}(x) \cdot x=\Ds\sigma_S(x)\cdot x =
\sigma_S(x)=\gamma_{S^\polar}(x).
\]
Furthermore, for all $x$
\[
	\Hs\gamma_{S^\polar}(x)\cdot x=\Hs\sigma_S(x)\cdot x= 0_n.
\]
By the product rule, we thus have for all $x$
\begin{equation}
	\label{eq:hessian-gamma-squared}
	\textstyle\frac{1}{2}\Hs \gamma_{S^\polar}^2(x)  = \frac{1}{2} \Ds((\Ds
	\gamma_{S^\polar}^2(x))') 
	=\Ds(\Ds\gamma_{S^\polar}(x)' \cdot \gamma_{S^\polar}(x))
	=\Ds\gamma_{S^\polar}(x)'\cdot\Ds\gamma_{S^\polar}(x) +
	\gamma_{S^\polar}(x)\Hs\gamma_{S^\polar}(x).
\end{equation}
Hence  for all $x$
\begin{align}
	\label{eq:H-gamma-squared-dot-x}
	\textstyle\frac{1}{2}\Hs\gamma_{S^\polar}^2(x)\cdot x
	&=(\Ds\gamma_{S^\polar}(x)'\cdot\Ds\gamma_{S^\polar}(x) +
	\gamma_{S^\polar}(x)\Hs\gamma_{S^\polar}(x))\cdot x\\
	&= \Ds\gamma_{S^\polar}(x)'\cdot\Ds\gamma_{S^\polar}(x) \cdot x\\
	&=\Ds\gamma_{S^\polar}(x)'\cdot\gamma_{S^\polar}(x),
\end{align}
and consequently for all $x$
\[
	x'\cdot \frac{1}{2}\Hs\gamma_{S^\polar}^2(x)\cdot x  =
	x'\cdot\Ds\gamma_{S^\polar}(x)'\cdot \gamma_{S^\polar}(x)
	= \gamma_{S^\polar}(x)'\cdot\gamma_{S^\polar}(x)
	=\gamma_{S^\polar}^2(x).
\]
Since $0\in S^\polar$,  $\gamma_{S^\polar}(x)\ge 0$ for all $x$. Thus from
\eqref{eq:hessian-gamma-squared} we see that for all $x$, $G_S(x)$ is the
sum of two  positive semidefinite matrices
$\gamma_{S^\polar}(x)\Hs\gamma_{S^\polar}(x)$ and a
$\Ds\gamma_{S^\polar}(x)'\cdot\Ds\gamma_{S^\polar}(x)$. The
positive definiteness of the rank one second term follows from the fact its only
non-zero eigenvalue is
$\Ds\gamma_{S^\polar}(x)\cdot\Ds\gamma_{S^\polar}(x)'$ and since
$\gamma_{S^\polar}(0)=\sigma_S(0)=0$, by positive homogeneity, and by
convexity, 
for all $x$, 
$\Ds\gamma_{S^\polar}(x) \ge 0$ (elementwise); consequently
$\Ds\gamma_{S^\polar}(x)\cdot\Ds\gamma_{S^\polar}(x)' \ge 0$. Since
$\lambda_{\mathrm{min}}(A+B) \ge \lambda_{\mathrm{min}}(A)$ for positive
semidefinite $A$ and $B$ \cite[9.12.2(7)]{Lutkepohl:1996aa}, we conclude
that $G_S(x)= \frac{1}{2}\Hs\gamma_{S^\polar}^2(x)$ is positive 
semidefinite for all $x$.

For arbitrary $x,y,z\in\reals^n$, $\lambda\in\reals$ and
$\epsilon\rightarrow 0$,  we have
\begin{align*}
	[x+y,z]_S &=(x+y)'\cdot G_S(z)\cdot z =x'\cdot G_S(z)\cdot z + y'\cdot
		G_S(z)\cdot z=[x,z]_S+[y,z]_S\\
	[\lambda x,y]_S &=(\lambda x)'\cdot G_S(y)\cdot y=\lambda(x'\cdot
		G_S(y)\cdot y)= \lambda [x,y]_S\\
		[x,x]_S & =x'\cdot G_S(x)\cdot x \ge 0 \mbox{\ since\ }
		G_S(x) \mbox{\ is positive semidefinite for all\ } x\\
		[y,x]_S & = y'\cdot G_S(x)\cdot x = y'\cdot \Ds\gamma_{S^\polar}(x)'
		\gamma_{S^\polar}(x)
		\le \gamma_{S^\polar}(y) \gamma_{S^\polar}(x) = 
		    [x,x]_S^\frac{1}{2} [y,y]_S^\frac{1}{2}\\
	[x,\lambda y]_S &= x'\cdot G_S(\lambda y)\cdot \lambda y= x'\cdot
		G_S(y)\cdot\lambda y = \lambda [x,y]_S\\
	[y,x+\epsilon y]_S &= x'\cdot G_S(x+\epsilon y) \cdot
		(x+\epsilon y) \rightarrow y'\cdot G_S(x)\cdot x \mbox{\ \
		since\ } \gamma_{S^\polar}\in C^2, 
\end{align*}
demonstrating that $[\cdot,\cdot]_S$ satisfies axioms SIP1--SIP6,
where the antepenultimate line used the
fact that $y'\cdot \Ds \gamma_{S^\polar}(x)'\le \gamma(y)$, and 
the penultimate line follows from
$\gamma_{S^\polar}$ being 1-homogeneous, implying $\gamma_{S^\polar}^2$
is 2-homogeneous and so by
Euler's theorem, $\Hs\gamma_{S^\polar}^2$ is 0-homogeneous. 
\end{proof}

Observe that by \eqref{eq:H-gamma-squared-dot-x}, we have
\begin{gather}
	[y,x]_S=y'\cdot \Ds\gamma_{S^\polar}(x)' \gamma_{S^\polar}(x) =
	\gamma_{S^\polar}(x) \Ds\gamma_{S^\polar}(x)\cdot y =
	\sigma_S(x) \Ds\sigma_S(x)\cdot y=\tau_S'(x,y),
	\label{eq:sip-via-tau}
\end{gather}
by \eqref{eq:alternate-tau-dash}.
One can reparametrise $[\cdot,\cdot]_S$ as follows.  By \citep[Page
55]{Schneider2014}, we have 
\[
	\left(\frac{1}{2}\gamma_{S}^2\right)^* =
	\frac{1}{2}\gamma_{S^\polar}^2 ,
\]
where $(\cdot)^*$ is the Legendre-Fenchel conjugate
\eqref{eq:convex_conjugate}. Combining this with the result from
\citep{Seeger:1992aa} that $(\Hs f(x))^{-1}=\Hs f^*(y)$, where $y=(\Ds
f)(x)$ and $x=(\Ds f)^{-1}(y)$, we can write
\begin{gather}
	\frac{1}{2} \Hs \gamma_{S^\polar}^2(u) = \frac{1}{2}(\Hs\gamma_S^2(x))^{-1},
	\label{eq:H-gamma-S-polar}
\end{gather}
where $u=(\frac{1}{2}\Ds \gamma_S^2(x))$ and
$x=(\frac{1}{2}\Ds\gamma_S^2)^{-1}(u)$.   Observe that if $A=\Hs\gamma_S^2$
is 0-homogeneous, then it is easy to see that $A^{-1}$ is too.   
Combining \eqref{eq:sip-via-tau} and \eqref{eq:H-gamma-squared-dot-x}
we have
\begin{gather}
[y,x]_S= y'\cdot  \frac{1}{2} \Hs \gamma_S^2 (x)^{-1} \cdot x.
\end{gather}

The semi-inner product $[x,y]_S$ induces a Finslerian geometry, a
generalisation of Riemannanian geometry, where the norm varies throughout
the space (as in the Riemannian case) but the unit balls are not
necessarily ellipsoids (as in the Riemannian case). 
\citet[page 16]{Rund:1959aa} notes that the metric of a Finsler space may
be regarded as being locally Minkowskian --- just like Riemannian geometry
without the quadratic restriction \citep{Chern:1996aa}.

For the situation of interest in the present paper (where we work with concave, rather
than convex, gauge and support functions), we can mimic the above
development to consider \Def{anti semi inner products} $\llangle
\cdot,\cdot\rrangle$.
We merely need replace the convex gauge $\gamma$ by the concave gauge $\beta$,
and the convex support function $\sigma$ by the concave support function $\rho$
and to reverse the inequality in the fourth axiom to read
\begin{description}
	\item[SIP$\widehat{\mathbf{4}}$] $| \llangle x,y \rrangle|^2 \ge \llangle
		x,x \rrangle \llangle y,y\rrangle,\ \ \  \forall x,y\in V$ .
\end{description}
We keep the other axioms the same.  Define 
\[
	\Def{\llangle y,x\rrangle_S \coloneqq y'\cdot \widehat{G}_S(x)\cdot
		x,}
\]
where $\Def{\widehat{G}_S\coloneqq \frac{1}{2}\Hs\beta_{S^\apolar}^2}$.  Then by a
similar argument to the above, we have that $\llangle y,x\rrangle_S$ is indeed an
anti semi inner product. The only change is working with negative
semidefiniteness instead of positive, and the  use of the ``reverse''
inequality
$\Ds\rho_{S^\apolar} \cdot y \ge \rho_{S^\apolar}(x)$, which is a
rephrasing of the result with loss functions that $L(x,y)\ge\minL(x)$.

Following the argument in \citep{Lumer:1961aa}, but reversing the
inequalities, we have that $\llangle x,x\rrangle^{\frac{1}{2}}$ is a concave
gauge function.
Translating to our loss notation we can write
\begin{equation}
	\label{eq:anti-semi-inner-product-S}
	\llangle y,x\rrangle_S = \minL_S(x) L_S(y,x),
\end{equation}
which can be seen to be a weighted conditional risk.   Utilising the
formula for Bregman divergences \eqref{eq:simple_bregman} 
and by analogy  with \eqref{eq:sip-via-tau}, we have  (writing
$\breg_S\coloneqq\breg_{\minL_S}$) that
	$\llangle y,x\rrangle_S = \rho_S(x) \Ds\rho(x)\cdot y
	=\minL_S(x) \langle \ell_S(x),y\rangle
	= \minL_S(x) L_S(y,x),
	$
	and thus
\begin{align*}
	\minL(x)\breg_S(y,x) &= \minL_S(x)[L_S(y,x)-\minL_S(y)]\\
	&=\minL_S(x) L_S(y,x)-\minL_S(x) \minL_S(y)\\
	&= \llangle y,x\rrangle_S - \sqrt{\llangle
	x,x\rrangle_S \llangle y,y\rrangle_S}.
\end{align*}
Expressing loss-theoretic quantities in terms of $\llangle
\cdot,\cdot\rrangle_S$ and hence  $\widehat{G}_S$ may also be conceptually
valuable, but we defer further investigation of this.  The normalisation
that naturally arises in \eqref{eq:anti-semi-inner-product-S} has not, to
our knowledge, arisen previously.  This does suggest that Riemannian
geometry, the traditional foundation of ``information geometry''
\citep{Amari:2016aa}, is not quite the right fit for the geometry of
losses, and we instead need the richer, locally Minkowskian
\citep{Chern:1996aa}, notion of Finslerian geometry \citep{Rund:1959aa}.

\subsection{The Antipolar Loss} 
\label{sub:the_polar_loss}

A loss function $\ell$ maps a distribution $p\in\Rpp^n$ to a loss vector
$\ell(p)\in\Rp^n$. Given $\ell(p)$, one might ask if we can recover $p$?
This problem arises naturally in a variety of settings
\citep{Vovk:2001,Gneiting2014}.  In practice it might be difficult to find
or even show the existence of an inverse loss, but in light of Remark
\ref{rem:0_homogeneous} we can show the existence of, and suggest several
ways to calculate, a pseudoinverse, $\ell^\apolar$. For reasons that will
become clear, we call the function $\ell^\apolar$ the \Def{antipolar loss}. 

The antipolar loss provides a universal substitution function
\citep{Kamalaruban:2015aa} for the Aggregating Algorithm
\citep{Vovk:2001,Vovk:1995,Vovk:1990}. The substitution function needs to
map an arbitrary superprediction $x\in \super(\ell)$ to a prediction
$p\in\probm$ such that $l\coloneqq\ell(p)$ dominates $x$ in the sense that
$l\preceq_{\Rp^n} x$ (that is, pointwise inequality not incurring more loss
under each $\rv y\in\cal Y$). Explicitly stating a substitution function is
the primary difference between the (unrealisable) aggregating
``pseudo-algorithm'' and the aggregating ``algorithm'' \citep{Vovk:2001}.
Determining the substitution function even for simple cases can be
difficult \citep{Zhdanov2011}.  We show below that by making use of
antipolars, we can determine such substitution functions; see
\citep{Friedlander:2014aa, Aravkin:2018aa} for further uses of
antipolarity.

\begin{proposition}\label{prop:the_polar_loss} 
    Let $\ell\colon\Rpp^n \to \Rp^n$ be a proper loss. There exists
    $\ell^\apolar\colon\Rpp^n \to \Rp^n$ with $\ell^\apolar \in \subdiff \cvsprt_{(\super\ell)^\apolar}$ that satisfies
    \begin{gather}
        \forall{p\in\Rpp^n},\  \ell(p) = (\ell\circ\ell^\apolar\circ\ell)(p)
	\mbox{\ \ \ and\ \ \ }  \forall{x\in\Rpp^n},\  \ell^\apolar(x) = (\ell^\apolar\circ\ell\circ\ell^\apolar)(x).
        \label{eq:pseudinverse_property}
    \end{gather}
    Furthermore $\ell^\apolar$ is a proper loss.
\end{proposition}
\begin{proof}
    Let $S\coloneqq\super(\ell)$.
    Since the subdifferential of a support function is 0-homogenous
    (Proposition \ref{prop:subdiff-zero-homog}) applying Lemma 
    \ref{lem:polars_and_inverses} we have
        $\forall{p\in\Rpp^n}$,
    \begin{align}
	\ell(p) \in \subdiff\cvsprt_{S}(p)
        &\iff
        \frac{\ell(p)·\cvsprt_{S^\apolar}(\ell(p))}{\cvsprt_{S^\apolar}(\ell(p))} 
        \in \subdiff\cvsprt_{S}(p)\\
	&\overset{\mathrm{L}\ref{lem:polars_and_inverses}}{\iff}
            \frac{p}{\cvsprt_{S}(p)} 
            \in \subdiff \cvsprt_{S^\apolar}(\ell(p)·\cvsprt_{S^\apolar}(\ell(p)))= 
            \subdiff\cvsprt_{S^\apolar}(\ell(p)).
    \end{align}
    Let us choose $\ell^\apolar$ to satisfy 
    $\ell^\apolar(\ell(p)) = p·\oneon{\cvsprt_{S}(p)}$. 
    This defines $\ell^\apolar\colon\R_{> 0}^n \rightarrow\R_{\ge 0}^n$. 
    We can then extend $\ell^\apolar$ to $\R_{\ge 0}^n\rightarrow\R_{\ge
    0}^n$ using the argument of Remark \ref{rem:bounded-to-unbounded}.
    Note that
    $\ell(\ell^\apolar(\ell(p)))=\ell\g\big(p·\oneon{\cvsprt_{S}(p)} ) =
    \ell(p)$. For the second claim regarding $\ell^\apolar$, exchange the
    roles of $\ell$ and $\ell^\apolar$ in the above argument and apply
    Proposition \ref{prop:subdiff-zero-homog}. 

    We now argue that $\ell^\apolar$ is proper. By construction
    $\ell^\apolar(\ell(p))=p/\rho_S(p)\in\subdiff\rho_{S^\apolar}(\ell(p))$.
    Let $q=\ell(p)$. Then $\ell^\apolar(q)\in\subdiff\rho_{S^\apolar}(q)$.
    Proposition \ref{prop:subdifferential_representation} then implies that
    $\ell^\apolar |_\probm$ is proper as long as 
    $S^\apolar\in\prop(\R_{\ge 0}^n)$ which we will now show. We have 
    \begin{gather}
	    \prop(\R_{\ge 0}^n)=\{S\in\cvx(\R_{\ge 0}^n\setminus\{0\}) \st
	    \rec(S)=\R_{\ge 0}^n\}.
    \end{gather}
    By the observation following \eqref{eq:polar_antipolar_dfn}, the map $S\mapsto
    S^\apolar$ takes closed shady sets to closed shady sets and
    $S\in\shdy(X_+)\Rightarrow S^\apolar\in\shdy(X_+^*)$. Furthermore
    $S^\apolar$ is convex. (Suppose $x_0^*,x_1^*\in S^\apolar$ and for some
    $\lambda\in(0,1)$, let $x_\lambda^*\coloneqq \lambda x_1^*
    +(1-\lambda) x_0^*$, then it is straightforward to check that $x_\lambda^*\in
    S^\apolar$ from the definition of $S^\apolar$.)
    Finally we have that 
    \begin{align*}
	    \rec(S^\apolar) & = \{d\in \R^n\st \forall x\in S,\ \langle
		    x^*,x\rangle \ge 1\Rightarrow \langle x^*+d,x\rangle
		    \ge 1\}\\
	    &=\{d\in\R^n\st\forall x\in S,\  \langle
		    x^*,x\rangle \ge 1\Rightarrow \langle x^*,x\rangle +
		    \langle d,x\rangle \ge 1\}\\
	    &=\{d\in\R^n\st\forall x\in S,\ \langle d,x\rangle \ge 1\}\\
	    &= \{d\in\R^n\st\forall x\in S,\ \langle d,x\rangle \ge 0\}\\
	    &=\bigcap_{x\in S} \{d\in\R^n\st \langle d,x\rangle \ge 0\}\\
	    &=\R_{\ge 0}^n,
    \end{align*}
    where the last line follows from the fact that $\rec(S)=\R_{\ge 0}^n$.
\end{proof}
\begin{figure}
    \centering
    \tikzsetnextfilename{polars_inverses}
    \begin{tikzpicture}
        \begin{axis}[domain=0.05:0.95, xmin=0, xmax=2, ymin=0, ymax=2, width=0.5\textwidth, height=0.5\textwidth,
            xlabel={${p_1, \ell(p;\rv y_1)}$}, ylabel={$p_2, \ell(p;{\rv y}_2)$}]
            \coordinate (o) at (axis cs: 0,0);
            \path[name path=upper_axis] (axis cs:0,5) -- (axis cs:5,0);

            \addplot[set, color=Blue, name path global=l] ({-ln(x)},{-ln(1-x)}) 
                node[pos=0.25, anchor=south, sloped] {$\super(\logloss)$}; 
            \addplot[set, color=Magenta, name path global=m] ({-x/(x*ln(x) + (1-x)*ln(1-x))},{(x-1)/(x*ln(x) + (1-x)*ln(1-x))})
                node[pos=0.64, anchor=south, sloped] {$\super\big(\logloss^\apolar)$};

            \addplot[fill=LightBlue] fill between[of=m and l];
            \addplot[fill=LightMagenta] fill between[of=m and upper_axis];

            \addplot[set, domain=0:1] ({x},{1-x}) node[pos=0.95, anchor=south west] {$\probm$};

            \coordinate (p) at (axis cs:0.75,0.25);
            \coordinate (lp) at (axis cs:0.2876820725,1.3862943611);

            \addplot[domain=0:1.8, name path global=p_ray,draw=none] ({0.75*x},{0.25*x});

            \draw[name intersections={of=p_ray and m, name=llp}] (llp-1) node[anchor=south,yshift=1ex] {$(\logloss^\apolar\circ\logloss)(p)$}; 
            \draw[shorten >=0cm, shorten >=1.5pt, -latex, name path global=lp_ray] (o) -- (lp);
            \draw[shorten >=0cm, shorten >=1.5pt, -latex, name path global=lp_ray] (o) -- (llp-1);

            \draw[shorten >=-4cm, shorten <=-2cm,faint] ($(o)!(lp)!(llp-1)$) -- (lp);
            \draw[right angle quadrant=2, right angle symbol={o}{lp}{llp-1},faint];
            \draw[shorten >=-3cm, shorten <=-5cm,faint] (llp-1) -- ($(o)!(llp-1)!(lp)$);
            \draw[right angle quadrant=2, right angle symbol={o}{llp-1}{lp},faint];    
            
            \draw (p) node[dot] {} node[anchor=south, yshift=1ex] {$p$};
            \draw (lp) node[dot] {} node[anchor=west] {$\logloss(p)$} ;
            \draw (llp-1) node[dot] {};
        \end{axis}
        \draw (o) node[anchor = north east] {$0$};
        \draw (o) node[anchor = north east] {$0$};
    \end{tikzpicture}
    \caption{
        Illustration of Proposition \ref{prop:the_polar_loss} using
	$\ell_{\mathrm{log}}$ (Table \ref{tab:common_losses}) with $\cal Y
	= [2]$. Upon predicting $p\in\probm$, we receive the loss
	vector $\logloss(p)$. Evaluating the antipolar loss
	$\logloss^\apolar$ at $\logloss(p)$ we have the point
	$(\logloss^\apolar\circ\logloss)(p)=\alpha p$ for some $\alpha>0$.
	However, $\logloss$ is $0$-homogeneous (Proposition 
	\ref{prop:subdifferential_representation}) and so 
	$(\logloss\circ\logloss^\apolar\circ\logloss)(p) = \logloss(p)$ for
	all $p\in\Rpp^2$. 
    }
    \label{fig:polars_inverses}
\end{figure}

Proposition \ref{prop:the_polar_loss} is illustrated with $\logloss$ in
Figure \ref{fig:polars_inverses}, which should be compared with the work of
\citet[pg.~23]{Shephard1953} which was the inspiration for the argument
regarding antipolar losses in the present paper\footnote{%
	\label{footnote:economic-duality}
	This has become known as ``Shephard's duality theorem'' in the
	economics literature \citep{Shephard:1970aa,Jacobsen1972,
	McFadden1978, Hanoch1978, Cornes:1992aa, Fare1994, Fare1995,
	Penot:2005aa, Zualinescu2013} and appears in standard microeconomics texts
	\citep{Varian1978}.  Shephard's development of dual theory in economics in
	his 1953 book \citep{Shephard1953} was described as 
	``one of the most original contributions to
	economic theory of all time'' \citep{Jorgenson:1981aa} due to three key
	ideas: 
	\begin{enumerate} 
	\item  The duality between ``cost'' and ``production'' functions
		(essentially polar duality of concave gauge functions); 
	\item Shephard's lemma \cite[page 74]{Varian1978}, \citep[Page
		141]{Mas-Collel:1995aa}: essentially the result
		\citep{Schneider2014} that the subgradient of a support
		function (in economics terminology, ``cost function''
		evaluated at some fixed price of input vectors)
		$\partial\sigma(x)$ is the support set (``conditional
		factor demand correspondence'' evaluated at the same price
		vector), and furthermore if $\sigma$ is differentiable at
		$x$, the support set is a singleton;  
	\item Homotheticity: essentially that the key
		functions of the theory are a composition of a positive
		monotone increasing scalar function and a positively
		homogeneous function of several variables.  
	\end{enumerate}
	The economic theory tends to obscure the simplicity of
	concave gauge duality because of the need to parametrise
	families of sets (either by the vector of inputs available
	to a firm or the vector of outputs) and the adoption of
	convoluted terminology (``conditional factor demand
	correspondence'' instead of ``support set''). The geometry
	in all cases is simply that of concave gauge duality, a
	point explicitly recognised by \citet{Hasenkamp1978} in the
	context of aggregation problems arising in production economics.
}.

More generally, it follows from Lemma \ref{lem:polars_and_inverses} that if
$\ell$ is a mapping $\relint(\pcone)\to\fcone$ with
$\super(\ell)\in\prop(\pcone)$ then $\ell^\apolar$ is a mapping
$\relint(\fcone)\to\pcone$ with $\super(\ell^\apolar)\in\prop(\fcone)$.
The pseudoinverse property of \eqref{eq:pseudinverse_property} can be
expressed using the notion of the direction of a vector as in footnote
\ref{footnote:cosmic}, allowing us to write  for all $p\in\Rpp^n$, $\dir
(\ell^\apolar\circ\ell)(p)=\dir p$.

\begin{proposition}\label{prop:proper_polar_loss}
Let $\ell\colon\Rpp^n \to \Rp^n$ be a proper loss. Then
$\cvsprt_{\super(\ell)}$ is strictly concave if and only if $\ell^\apolar$
is strictly proper.
\end{proposition}
\begin{proof}
	This follows immediately from Corollaries \ref{lem:apolar_strictly_convex} 
	and \ref{por:strictly_convex_proper}.
\end{proof}
\begin{corollary}\label{prop:unique_polar_loss}
If $\cvsprt_{\super(\ell)}$ is strictly concave then for any function
$\emm$ that satisfies $\emm \in \subdiff \cvsprt_{(\super\ell)^\apolar}$ we
have $\emm=\ell^\apolar$.
\end{corollary}
\begin{proof}
    It follows that  $\super(\ell)^\apolar$ is strictly convex (Corollary
    \ref{lem:apolar_strictly_convex}), thus
    $\subdiff\cvsprt_{\super(\ell)^\apolar}$ is always a singleton
    (Corollary \ref{cor:cvsprt_differentiable}). Thus there is only one
    possible selection $\emm$ with
    $\emm\in\subdiff\cvsprt_{(\super\ell)^\apolar}$.
\end{proof}

\begin{remark}\label{rem:polars_closed_under_scalar}
	\normalfont
    If $\ell^\apolar$ is the antipolar of the loss $\ell$ then clearly so
    is the family $\g(\alpha\ell^\apolar)_{\alpha>0}$ since as $\ell$ is
    0-homogeneous $\ell\circ(\alpha\ell^\apolar)= \ell\circ\ell^\apolar$.
\end{remark}

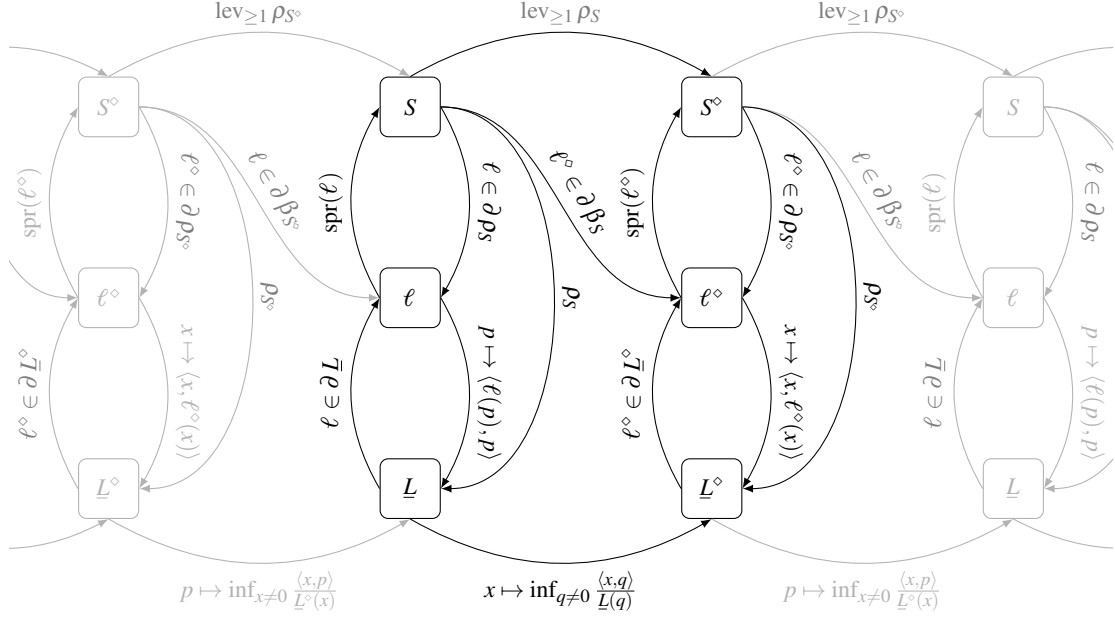
\begin{figure}
    \centering
    \tikzsetnextfilename{polar_relationships}
    \begin{tikzpicture}[every node/.style = {rounded corners=3pt, minimum size=2.5em}]
        \clip (-0.495\textwidth,-7.0) rectangle (0.495\textwidth,1.45);
        \coordinate (o) at (0,0);
        \node [draw,left = 5em of o] (s) {$S$};
        \node [draw,below = 8em of s.west, anchor = west] (l) {$\ell$};
        \node [draw,below = 8em of l.west, anchor = west] (L) {$\minL$};
        \node [draw,right = 5em of o] (sp) {$S^\apolar$};
        \node [draw,below = 8em of sp.west, anchor = west] (ll) {$\ell^\apolar$};
        \node [draw,below = 8em of ll.west, anchor = west] (LL) {$\minL^\apolar$};
        \node [draw,faint, right = 10em of sp] (spp) {$S$};
        \node [draw,faint, below = 8em of spp.west, anchor = west] (lll) {$\ell$};
        \node [draw,faint, below = 8em of lll.west, anchor = west] (LLL) {$\minL$};
        \node [draw,faint, left = 10em of s] (sspp) {$S^\apolar$};
        \node [draw,faint, below = 8em of sspp.west, anchor = west] (llll) {$\ell^\apolar$};
        \node [draw,faint, below = 8em of llll.west, anchor = west] (LLLL) {$\minL^\apolar$};
        
        \draw node [left = 8em of sspp] {} %
            edge [faint,bend left, -latex, sloped] (sspp.north);
        \draw node [right = 8em of spp] {} %
            edge [faint,bend right, latex-, sloped] (spp.north);
        \draw node [left = 8em of LLLL] {} %
            edge [faint,bend right, -latex, sloped] (LLLL.south);
        \draw node [right = 8em of LLL] {} %
            edge [faint,bend left, latex-, sloped] (LLL.south);
        \draw node [left = 8em of sspp] {} %
            edge [faint,out=0,in=-180, -latex, sloped] 
            node[draw=none, midway, above, minimum size=0] 
            {} (llll.west);
        \node [right = 8em of lll] (z) {};
        \draw (spp)
            edge [faint,out=0,in=-180, -latex, sloped] 
            node[draw=none, midway, above, minimum size=0] {} (z);
            
        \path (s.east) edge [out=0,in=-180, -latex, sloped] 
            node[draw=none, midway, above, minimum size=0] 
            {$\ell^\apolar\in\subdiff\agauge_S$} (ll.west);
        \path (sp.east) edge [faint,out=0,in=-180, -latex, sloped] 
            node[draw=none, midway, above, minimum size=0] 
            {$\textcolor{gray}{\ell\in\subdiff\agauge_{S^\apolar}}$} (lll.west);
        \path (sspp.east) edge [faint, out=0,in=-180, -latex, sloped] 
            node[draw=none, midway, above, minimum size=0] 
            {$\textcolor{gray}{\ell\in\subdiff\agauge_{S^\apolar}}$} (l.west);
                
        \path (s.north) edge [bend left, -latex, sloped] 
            node[draw=none, midway, above, minimum size=0] 
            {$\textcolor{gray}{\lev_{≥1}\cvsprt_{S}}$} (sp.north);    
        \path (sp.north) edge [faint,bend left, -latex, sloped] 
            node[draw=none, midway, above, minimum size=0] 
            {$\textcolor{gray}{\lev_{≥1}\cvsprt_{S^\apolar}}$} (spp.north);             
        \path (sspp.north) edge [faint,bend left, -latex, sloped] 
            node[draw=none, midway, above, minimum size=0] 
            {$\textcolor{gray}{\lev_{≥1}\cvsprt_{S^\apolar}}$} (s.north);    
        
        \path (L.south) edge [bend right, -latex, sloped] 
            node[draw=none, midway, below, minimum size=0] 
            {$x\mapsto\inf_{q≠0}\frac{\inner{x;q}}{\minL(q)}$} (LL.south); 
        \path (LL.south) edge [faint,bend right, -latex, sloped] 
            node[draw=none, midway, below, minimum size=0] 
            {$p\mapsto\inf_{x≠0}\frac{\inner{x;p}}{\minL^\apolar(x)}$} (LLL.south);
        \path (LLLL.south) edge [faint,bend right, -latex, sloped] 
            node[draw=none, midway, below, minimum size=0] 
            {$p\mapsto\inf_{x≠0}\frac{\inner{x;p}}{\minL^\apolar(x)}$} (L.south); 
        
        \path (llll.west) edge [faint, bend left, -latex, sloped] 
            node[draw=none, midway, above, minimum size=0] 
            {$\super(\ell^\apolar)$} (sspp.west);
        \path (LLLL.west) edge [faint, bend left, -latex, sloped] 
            node[draw=none, midway, above, minimum size=0] 
            {$\textcolor{gray}{\ell^\apolar\in\subdiff\minL^\apolar}$} (llll.west);
        \path (sspp.east) edge [faint, bend left, -latex, sloped] 
            node[draw=none, midway, above, minimum size=0] 
            {$\textcolor{gray}{\ell^\apolar\in\subdiff\cvsprt_{S^\apolar}}$} (llll.east);
        \path (llll.east) edge [faint, bend left, -latex, sloped] 
            node[draw=none, midway, above, minimum size=0] 
            {$x\mapsto\inner{x;\ell^\apolar(x)}$} (LLLL.east);
        \path ($(sspp.east)$) edge [faint, bend left=90, -latex, sloped] 
            node[draw=none, midway, above, minimum size=0] 
            {$\textcolor{gray}{\cvsprt_{S^\apolar}}$} ($(LLLL.east)$);
            
        \path (l.west) edge [bend left, -latex, sloped] 
            node[draw=none, midway, above, minimum size=0] 
            {$\super(\ell)$} (s.west);
        \path (L.west) edge [bend left, -latex, sloped] 
            node[draw=none, midway, above, minimum size=0] 
            {$\ell\in\subdiff\minL$} (l.west);
        \path (s.east) edge [bend left, -latex, sloped] 
            node[draw=none, midway, above, minimum size=0] 
            {$\ell\in\subdiff\cvsprt_{S}$} (l.east);
        \path (l.east) edge [bend left, -latex, sloped] 
            node[draw=none, midway, above, minimum size=0] 
            {$p\mapsto\inner{\ell(p);p}$} (L.east);
        \path ($(s.east)$) edge [bend left=90, -latex, sloped] 
            node[draw=none, midway, above, minimum size=0] 
            {$\cvsprt_S$} ($(L.east)$);
            
        \path (ll.west) edge [bend left, -latex, sloped] 
            node[draw=none, midway, above, minimum size=0] 
            {$\super(\ell^\apolar)$} (sp.west);
        \path (LL.west) edge [bend left, -latex, sloped] 
            node[draw=none, midway, above, minimum size=0] 
            {$\ell^\apolar\in\subdiff\minL^\apolar$} (ll.west);
        \path (sp.east) edge [bend left, -latex, sloped] 
            node[draw=none, midway, above, minimum size=0] 
            {$\ell^\apolar\in\subdiff\cvsprt_{S^\apolar}$} (ll.east);
        \path (ll.east) edge [bend left, -latex, sloped] 
            node[draw=none, midway, above, minimum size=0] 
            {$x\mapsto\inner{x;\ell^\apolar(x)}$} (LL.east);
        \path ($(sp.east)$) edge [bend left=90, -latex, sloped] 
            node[draw=none, midway, above, minimum size=0] 
            {$\cvsprt_{S^\apolar}$} ($(LL.east)$);
            
        \path (lll.west) edge [faint, bend left, -latex, sloped] 
            node[draw=none, midway, above, minimum size=0] 
            {$\super(\ell)$} (spp.west);
        \path (spp.east) edge [faint, bend left, -latex, sloped] 
            node[draw=none, midway, above, minimum size=0] 
            {$\textcolor{gray}{\ell\in\subdiff\cvsprt_{S}}$} (lll.east);   
        \path (LLL.west) edge [faint, bend left, -latex, sloped] 
            node[draw=none, midway, above, minimum size=0] 
            {$\textcolor{gray}{\ell\in\subdiff\minL}$} (lll.west);
        \path (lll.east) edge [faint,, bend left, -latex, sloped] 
            node[draw=none, midway, above, minimum size=0] 
            {$p\mapsto\inner{\ell(p);p}$} (LLL.east);
        \path ($(spp.east)$) edge [faint,bend left=90, -latex, sloped] 
            node[draw=none, midway, above, minimum size=0] {} ($(LLL.east)$);
    \end{tikzpicture}
    \caption{Illustration of the relationships superprediction sets $S$,
    associated loss functions $\ell$, and conditional Bayes' risks $\minL$,
    along with their antipolar counterparts. (The diagram, which is a finite graph,
    has been ``unrolled'' to make reading easier.)}
    \label{fig:loss_set_duality}
\end{figure}

There are three ways the antipolar loss $\ell^\apolar$ can be computed given a
proper loss $\ell$ (cf.\ \S\ref{sec:gauge_functions_and_polar_duality}):
we can \begin{enumerate}
    \item take the superdifferential of the concave support function 
	    of the antipolar superprediction set
	    $\ell\mapsto\subdiff\cvsprt_{(\super\ell)^\apolar}\ni
	    \ell^\apolar$;
    \item compute the antipolar of the associated conditional Bayes risk
	    function and superdifferentiate $L\mapsto\subdiff L^\apolar \ni
	    \ell^\apolar$; or
    \item solve the optimisation problem in
\end{enumerate}
\begin{gather}
    \ell^\apolar(p) \in \arginf\set{ x\in \Rp^n; \inner{\ell(p);x}≥1}.
\end{gather}
The complete set of antipolar loss function relationships is presented in
Figure \ref{fig:loss_set_duality}. The notion of the antipolar loss and its
relationship to the concave polar of the superprediction set provides
conceptual insight. Furthermore, at least in some cases one can determine
the inverse in closed  form (see equation \ref{eq:gauge_polar} in
\S\ref{ssec:concave_norm_losses}, as well as the other examples in
\S\ref{ssec:brier_loss} and \S\ref{ssec:cobb_douglas}).


\subsection{Convexifying Proper Losses: The Canonical Link}

All proper losses $\ell$ have convex superprediction sets, but that does not
imply  that the partial functions $\ell_i=\ell(\cdot;i)$ are convex for all
$i\in[m]$
\citep{Reid:2010,Vernet:2016aa}.  However, such proper losses with non-convex
partial losses can be made convex by reparametrisation.

A \Def{composite proper loss} $\ell\circ\ppt^{-1}$ is the composition of a
proper loss $\ell$ and an (inverse) ``link function'' $\ppt^{-1}$ that
reparametrizes the loss \citep{Reid:2010,Vernet:2016aa}.  The
aforementioned papers studied such links using the tools of differential
calculus.  We will now show that the geometric perspective of the present paper,
along with the properties of antipolar losses, allows a simpler proof of the
fact that there is always a special link function which ensures the
resulting composite loss is a convex function.  This link function is
called the  ``canonical link'' in \citep{Reid:2010} (binary case) and
\citep{Vernet:2016aa} (general multiclass case); as we shall see below, the
canonical loss is indeed the composition of the loss with its associated 
antipolar loss. 

We first need some additional notions
\citep{Pennanen:1999aa,Gissler:2022aa}.  Suppose $X$ and $Y$ are sets, and
$K\subset Y$ is a convex cone, and $f\colon X\rightarrow Y$ with $\dom f$
convex. Recalling from \S\ref{sec:cones} the ordering $\succeq_K$, we say
$f$ is \Def{$K$-convex} if for all $x_0,x_1\in X$, and all
$\alpha\in(0,1)$, 
\begin{gather}
	f(\alpha x_1+(1-\alpha)x_0) \preceq_K \alpha
	f(x_1)+(1-\alpha)f(x_0).
\end{gather}
That is, $f(\alpha x_1+(1-\alpha)x_0)- \alpha f(x_1)+(1-\alpha)f(x_0)\in -K$.
The \Def{$K$-epigraph} of a function $f\colon
X\rightarrow Y$ is 
\begin{gather}
	\Def{\epi_K f}{\left\{(x,y)\in X\times Y\st f(x) \preceq_K y\right\}}.
\end{gather}
For functions $f\colon X\rightarrow\R$, the traditional epigraph $\epi f$
corresponds to $\epi_K f$ with $K=\R_{\le 0}$.  Analogous to the result for
the traditional epigraph that $f$ is convex iff $\epi f$ is, we have 
\citep[Lemma 14.8]{Jahn:2011aa}:
\begin{lemma}
	\label{lemma:K-convex-characterisation}
	Suppose $f\colon X\rightarrow Y$ and $\dom f$ is convex, and
	$K\subset Y$ is a cone. Then $f$ is $K$-convex if and only if
	$\epi_K f$ is a convex subset of $X\times Y$.
\end{lemma}
Suppose $f\colon\R^n\rightarrow\R^n$, and write
$f(x)=(f_1(x),\ldots,f_n(x))$. Let $K=\R_{\ge 0}^n$. Then $f$ is
$K$-convex if 
\begin{align*}
	& f(\alpha x_1 +(1-\alpha)x_0) - \alpha f(x_1)-(1-\alpha)f(x_0) \in
	\R_{\le 0}^n\\
	\Rightarrow\ \  & \forall i\in [n],\ \  f_i(\alpha x_1 +(1-\alpha)x_0) 
- \alpha f_i(x_1)-(1-\alpha)f_i(x_0) \in (\infty,0]\\
\Rightarrow\ \  &\forall i\in[n],\ \  f_i(\alpha x_1 +(1-\alpha)x_0) \le \alpha
f_i(x_1)+(1-\alpha)f(x_0)\\
\Rightarrow\ \ & \forall i\in[n],\ \ f_i\ \mbox{is convex}.
\end{align*}
Hence (confer \citep[Section 4.4.4]{Gissler:2022aa}) $f$ is
\Def{component-wise convex}:
\begin{lemma}
	\label{lemma:componentwise-convex}
	$f\colon\R^n\rightarrow\R^n$ is $\R_{\ge 0}^n$-convex iff $f_i\colon
	\R^n\rightarrow \R$ is convex for all $i\in[n]$.
\end{lemma}

Let $\Def{\tilde{\ell}}{\ell\circ\ell^\apolar}$ denote the \Def{canonical
loss} induced by composing an arbitrary proper loss $\ell$ with its
associated antipolar loss $\ell^\apolar$. 
Proposition \ref{prop:the_polar_loss} implies there exists
a function $\gamma_{\tilde{\ell}}\colon\R_{\ge 0}^n\rightarrow\R_{>0}$ such that for all
$x\in\R_{\ge 0}^n$, $\tilde{\ell}(x)=\gamma_{\tilde{\ell}}(x) x$.

We can now prove the following result directly, without the 
need for differential calculus as used in 
\citep[Corollary 32 \emph{et seq.}]{Vernet:2016aa}.

\begin{theorem}\label{thm:canonical_link}
	Suppose $\ell$ is a proper loss. Then the canonical loss
	$\tilde{\ell}$ is component-wise convex.
\end{theorem}
\begin{proof}
	By Lemmas \ref{lemma:K-convex-characterisation} and 
	\ref{lemma:componentwise-convex}, $\tilde{\ell}$ is component-wise
	convex if $\epi_K\tilde{\ell}$ is convex, with $K=\R_{\ge 0}^n$,
	which we now proceed to show. Write $\gamma=\gamma_{\tilde{\ell}}$.
	Then
	\begin{align}
	\epi_K \tilde{\ell} & = \{(x,y)\in\R_{\ge 0}^n
			\times\R_{\ge 0}^n\st \tilde{\ell}(x) \preceq_K y\}\\
		&= \bigcup_{x\in\R_{\ge 0}^n} \{(x,y)\st
			y\in\R_{\ge 0}^n,\ \tilde{\ell}(x) \preceq_K y\}\\
		&= \bigcup_{x\in\R_{\ge 0}^n} \{(x,y)\st
			y\in\R_{\ge 0}^n,\ \gamma(x) x\preceq_K y\}\\
		&= \bigcup_{x\in\R_{\ge 0}^n} \{(x,y)\st
			y\in\R_{\ge 0}^n,\  x\preceq_K y/\gamma(x)\}\\
	        &= \bigcup_{x\in\R_{\ge 0}^n} \{(x,\gamma(x) y')\st
			y'\in\R_{\ge 0}^n,\  x\preceq_K y'\}\\
		&= \bigcup_{x\in\R_{\ge 0}^n} \{x\}\times
			\gamma(x)(\{x\}+\R_{\ge 0}^n)\\
		&= \R_{\ge 0}^n \times
			\bigcup_{x\in\R_{\ge 0}^n} \gamma(x)(\{x\}+\R_{\ge 0}^n)\\
		&= \R_{\ge 0}^n \times \bigcup_{x\in\R_{\ge 0}^n} 
		        \gamma(x)\{y\in\R_{\ge 0}^n\st x\preceq_K y\}\\
		&= \R_{\ge 0}^n \times \bigcup_{x\in\R_{\ge 0}^n} 
			\{y\in\R_{\ge 0}^n\st x\preceq_K y/\gamma(x)\}\\
		&= \R_{\ge 0}^n \times \bigcup_{x\in\R_{\ge 0}^n} 
			\{y\in\R_{\ge 0}^n\st \gamma(x) x\preceq_K y\}\\
		&= \R_{\ge 0}^n \times \bigcup_{x\in\R_{\ge 0}^n} 
		\{y\in\R_{\ge 0}^n\st \tilde{\ell}(x) \preceq_K y\}\\
		&=\R_{\ge 0}^n \times \bigcup_{l\in\ell(\Delta)}
			\{y\in\R_{\ge 0}^n\st l \preceq_K y\}\\
		&=\R_{\ge 0}^n\times \super \tilde{\ell},
	\end{align}
	by  \eqref{eq:super2}.
	By \citep[Proposition A.1.2.3]{hiriarturruty2001fca}, the Cartesian
	product of convex sets is convex.  Since $\R_{\ge 0}^n$ and
	$\super\tilde{\ell}$ are both convex, we have thus shown that
	$\epi_K\tilde{\ell}$ is also convex, concluding the proof.
\end{proof}

\subsection{The Naturalness of our Setup, and its Advantages}
\label{sec:naturalness}

The above development presumes a particular form for $S$, namely that it is
in the class $\prop(\R_{\ge 0}^n)$. This assumption is necessary for our
proofs, but is it reasonable? Furthermore, the definition of a loss
function as the subgradient of $\rho_S$ is rather unusual. What advantage
does it have? Finally, the introduction of the link function seems to just
complicate matters even further. What additionality does it bring? In this
brief subsection we provide succinct answers to these natural questions.

The choice of  $\prop(\R_{\ge 0}^n)$ is indeed simply justified by the
results obtainable: if $S$ is not convex, then $\rho_S=\rho_{\co S}$ so
there is nothing lost in assuming convexity. Since we only use the loss via
its average (the risk) we are thus only interested in supporting
hyperplanes with normals in the positive orthant; thus the assumption on
the recession cone simply allows unbounded losses to exist, and ensures
they are bounded on the relative interior of the simplex. 
Some advantages of defining the
loss as $\subdiff\rho_S$ will be elaborated in the remainder of the paper
(in terms of designing losses), but it also allows a direct connection to
the question of mixability of a loss, which is defined in terms of the
geometry of $S$ \citep{van-Erven:2012vz}, as well as to the dual theory of
production economics (as elaborated in footnote
\ref{footnote:economic-duality}). Without the inherent geometrical
structure, it seems unlikely one would have stumbled across the notion
of an antipolar loss.  Finally, the representation of a general
(not necessarily proper) loss $\ell$, such that $\super
\ell\in\prop(\R_{\ge 0}^n)$, as $\ell = \lambda\circ\psi^{-1}$ (where
$\lambda$ is proper, and $\psi^{-1}$ is an inverse link function) allows a
very clean separation of concerns: the statistical properties of the loss
are controlled by $\lambda$, and the convexity of the partial losses
$\ell_i$, $i\in[n]$ is controlled by $\psi$; confer \citep{Vernet:2016aa}.

\section{Examples of the Antipolar Loss} 
\label{sec:polar_applications}

We now present some examples of the antipolar loss for some 
well-known standard loss functions. 

\subsection{Concave Norm Losses} 
\label{ssec:concave_norm_losses}
In general the calculation of antipolars of superprediction sets, or
equivalently the antipolar loss may be difficult to achieve in closed form.
However, analogous to the case of classical $l_p$ gauges (norms) with the
duality property that $\ball_p^\polar=\ball_q$ with
$\frac{1}{p}+\frac{1}{q}=1$, 
there is a parametric
family of antigauges which has an attractive self-closure property with
respect to taking antipolars. Following \cite{Barbara1994}, we define
$\beta_a\colon\Rp^n\to\Rx$ for $a\in[-\infty,1]\setminus{0}$ as follows
\begin{align}
         \beta_{-\infty}(p) &\coloneqq 
            \min_{\rv y\in\cal Y} p_{\rv y} \\
         \beta_a(p) &\coloneqq \begin{cases}
            \g(\sum_{\rv y \in \cal Y} p_{\rv y}^a)^{\oneon{a}} & p\in\Rpp^n\\
            0                   & p\in\bd\Rp^n,
	    \end{cases}\hspace*{2cm} \forall{a\in{(-\infty,0)}} \\
        \beta_a(p) &\coloneqq 
            \g(\sum_{\rv y \in \cal Y} p_{\rv y}^a)^{\oneon{a}} 
	 \hspace*{4.7cm} \forall{a\in{(0,1].}}
    \label{eq:concave_norm_losses_defn}
\end{align}
\citet{Barbara1994} showed for all $a\in[-\infty,1]\setminus{0}$  
that $\beta_a$ is indeed an antigauge and furthermore
\begin{gather}
   \oneon{a}+\oneon{b}=1 \iff \beta_{a}^\apolar = \beta_{b}.\label{eq:gauge_polar}
\end{gather}
Note that if $a\in(0,1]$ then $b\in[-\infty,0)$, and if $a\in[-\infty,0)$
then $b\in(0,1]$. Thus no antigauge in the family 
$\{\beta_a\st a\in [-\infty,1]\setminus\{0\}\}$
can be its
own antipolar (unlike the classical result that $\ball_2^\polar=\ball_2$).
The family of antigauges 
$\{\beta_a\st a\in [-\infty,1]\setminus\{0\}\}$
can be used to define
a family of proper losses on $n$ outcomes. Since $\beta_a$ is an antigauge,
it can be written either as the antigauge of the set $\lev_{≥1}(\beta_a)$,
or using polar duality as the concave support function of the set
$\lev_{≥1}(\beta_a^\apolar)$, which is the convention we follow below.

In order to find the associated loss function we need $\ell_a$ such that
$\super(\ell_a) = \lev_{≥1}(\beta_a^\apolar)$. The self antipolar property
\eqref{eq:gauge_polar} makes this easy. Since $\beta_{\!\frac{a}{a-1}}$ is
differentiable on the interior of its domain we have
\begin{align}
    \forall a\in (-\infty,1)\setminus{0},\forall p\in\Rpp^n,\ 
    \ell_a(p) =\nabla\beta_{\!\frac{a}{a-1}}(p)
    =
    \exp\left(\textstyle\frac{1}{\beta_{\!\frac{a}{a-1}}(p)}·p ;\,
    {\frac{1}{a-1}}\right),
\end{align}
where the exponentiation of vectors is defined component wise,
$\Def{\exp\left( p;\, a\right)}\coloneqq
(p_1^a,\dots,p_n^a)$. Applying the same procedure to the antipolar loss we
have
\begin{align}
	 \forall a\in (-\infty,1)\setminus \{0\},\ \forall p\in\Rpp^n,\ 
    \ell_a^\apolar(p) = \nabla\beta_{a}(p) = 
    \exp\left(\textstyle\frac{1}{\beta_{a}(p)}·p;\, {a-1}\right).
    \label{eq:concave_norm_loss}
\end{align}

\begin{figure}
    \centering
    \tikzsetnextfilename{concave_norm_losses}
    \subcaptionbox{The functions $\g(\beta_a)_{a\in{[-\infty,0)}}$ are
    coloured from blue to white, and their antipolar counterparts,
    $\g(\beta_a)_{a\in{(0,1]}}\simeq\g(\beta_a^\apolar)_{a\in{[-\infty,0)}}$,
	    are coloured from white to green. As $a\nearrow 1$, $\ell_a\to
	    \ooneloss$, and as $a\to-\infty$, $\ell_\alpha\to 1_2$, the
	    constant loss.
	    \label{fig:concave_norm_losses_cond_bayes_risks}}
	    {\hspace*{-1ex}\begin{tikzpicture}
        \begin{axis}[domain=0:1, xmin=0, xmax=1, ymin=0, ymax=2, width=0.5\textwidth, height=0.5\textwidth, xlabel={$p_1$}, ylabel={$\beta_\alpha(p_1,1-p_1)$},
        xtick={0,0.25,0.5,0.75,1},
        ytick=,
        extra x ticks={0},
        extra y ticks={0, 0.5},
        inner axis line style={-},
        xtick style={draw},
        ytick style={draw},
        xlabel style={anchor=west},
        ylabel style={anchor=south},
        ]
            \pgfplotsinvokeforeach{1,...,40}{
                \pgfmathsetmacro\a{0.9525^(41-#1)}
                \pgfmathtruncatemacro{\pcnt}{100*0.9^(41-#1)}
                \edef\temp{\noexpand%
                   \addplot[color=Blue!\pcnt, domain=0.0001:1] {
                        ((1-x)^(\a) + x^(\a))^(1/\a)
                    };
                }\temp
            }
            \addplot[color=Blue]{1};
            \addplot[color=Green, domain=0:0.5] {x};
            \addplot[color=Green, domain=0.5:1] {-x + 1};
            \pgfplotsinvokeforeach{1,...,25}{
                \pgfmathsetmacro\a{0.925^(26-#1)}
                \pgfmathtruncatemacro{\pcnt}{100*0.9^(26-#1)}
                \edef\temp{\noexpand%
                    \addplot[color=Green!\pcnt, domain=0.0001:1] {
                        ((1-x)^(\a/(\a-1)) + x^(\a/(\a-1)))^((\a-1)/\a)
                    };
                }\temp
            }
        \end{axis}
    \end{tikzpicture}}
    \hspace{1ex}
    \tikzsetnextfilename{concave_norm_losses_superprediction}
    \subcaptionbox{The loss $\ell_{\varfrac{3}{4}}$ and its antipolar
    $\ell_{-3}$ acting on the vector $p\coloneqq(\oneon{3},\varfrac{2}{3})$.}
    {\hspace*{-6ex}\begin{tikzpicture}
           \begin{axis}[ xmin=0, xmax=2, ymin=0, ymax=2, width=0.5\textwidth, height=0.5\textwidth, xlabel={${p_1, \ell(p;\rv y_1)}$}, ylabel={$p_2, \ell(p;{\rv y}_2)$}]
               \coordinate (o) at (axis cs: 0,0);
               \path[name path=upper_axis] (axis cs:5,0) -- (axis cs:5,5) -- (axis cs:0,5);
               \pgfmathsetmacro\a{3/4};
               \addplot[set, color=Green, domain=0:1, name path=lpp] 
                   ({
                       ((1-x)^(1/(\a-1)))*((1 - x)^(\a/(\a-1)) + x^(\a/(\a-1)))^(-1/\a)
                   }, {
                       (x^(1/(\a-1)))*((1 - x)^(\a/(\a-1)) + x^(\a/(\a-1)))^(-1/\a)
                   });
               \addplot[fill=LightGreen] fill between[of=lpp and upper_axis];
               \addplot[set, color=Blue, domain=0.01:0.99, name path=lp] 
                   ({
                       x^(\a-1)*((1-x)^\a + x^\a)^(1/\a-1)
                   }, {
                       ((1-x)^(\a-1))*((1-x)^\a + x^\a)^(1/\a-1)
                   }) node[pos=0.36, anchor=south, sloped] {$\super(\ell_{\varfrac{3}{4}})$};
               \addplot[fill=LightBlue] fill between[of=lp and upper_axis];
               \draw (axis cs:2,0) node[anchor=south east, color=Green] {$\super(\ell_{-3})$};
               
               \addplot[set, domain=0:1] ({x},{1-x}) node[pos=0.95, anchor=south west] {$\probm$};

               \coordinate (p) at (axis cs:0.333,0.666);
               \coordinate (lp) at (axis cs:1.389,1.168);
               \coordinate (lpp) at (axis cs:0.268387, 0.536784);

               \draw[faint, name path global=p_ray] (o) -- ($5*(p)$);
               \draw[faint] (o) -- ($5*(lp)$);
               
               \draw[shorten >=-3cm, shorten <=-4cm,faint] ($(o)!(lp)!($5*(p)$)$) -- (lp);
               \draw[right angle quadrant=2, right angle symbol={o}{lpp}{lp},faint];
               \draw[shorten >=-3cm, shorten <=-4cm,faint] ($(o)!(lpp)!(lp)$) -- (lpp);
               \draw[right angle quadrant=2, right angle symbol={o}{lp}{lpp},faint];
               
               \draw[shorten >=0cm, shorten >=1.5pt, -latex] (o) -- (lp);
               \draw[shorten >=0cm, shorten >=1.5pt, -latex] (o) -- (p);
               
               \draw (p) node[dot] {} node[anchor=south, yshift=1ex] {$p$};
               \draw (lp) node[dot] {} node[anchor=south, yshift=1ex] {$\ell_{\varfrac{3}{4}}(p)$} ;
           \end{axis}
           \draw (o) node[anchor = north east] {$0$};
           \draw (lpp) node[dot] {} node[anchor=east] {$\ell_{-3}°\ell_{\varfrac{3}{4}}(p)$} ;
       \end{tikzpicture}}
    \caption{The concave norm conditional Bayes risk functions $\beta_a$
    and losses $\ell_a$, and an illustration of self-polarity of the family
    $\{\ell_a\}_{a\in[-\infty,1]\setminus\{0\}}$.}
    \label{fig:concave_norm_losses}
    \end{figure}
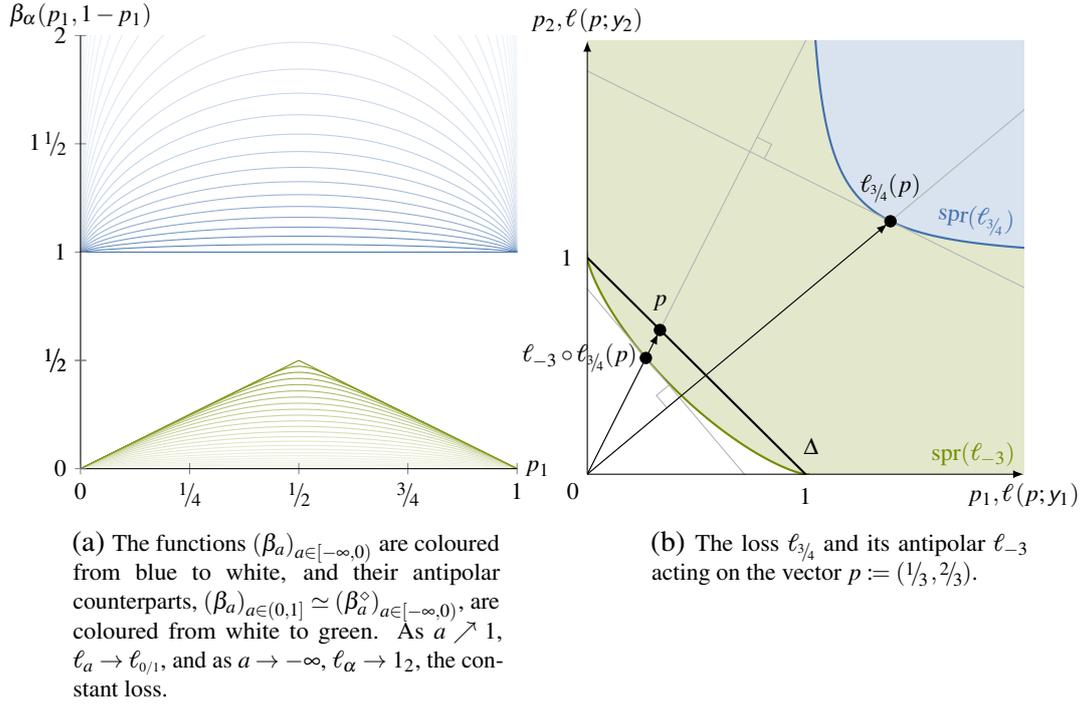
   
There are two special values of $a\in[-\infty,1]\setminus{0}$ worth mentioning: 
When $a=1$, $b=-\infty$ and  $\beta_b$ is no longer differentiable, 
but it is superdifferentiable, with
\begin{gather}
    \forall{p\in \Rpp^n},\ \ooneloss(p) \in \subdiff\beta_{-\infty}(p).
\end{gather}
When $a  = -\infty$, $b=1$ and $\nabla \beta_1(p)=1_n$. Thus
\begin{gather}
    \forall{p\in \Rpp^n},\ \ell_{-\infty}(p) \coloneqq 1_n \in \subdiff\beta_{1}(p),
    \label{eq:concave_norm_loss_polar}
\end{gather}
the \emph{constant loss}. Note that $\ell_{-\infty} = \ooneloss^\apolar$.
The closure of the family $(\ell_a)_{a\in[-\infty,1]\setminus{0}}$ under
the antipolar operation is illustrated in Figure \ref{fig:concave_norm_losses}.

\begin{example}\label{ex:approximating_oone_polar}
	\normalfont
	Misclassification loss is not \emph{strictly} proper and so
    $\subdiff\cvsprt_{\super(\ooneloss)}(p)$ will not be a singleton for
    all $p\in\probm$. This poses a problem for calculating
    $\ooneloss^\apolar$, since the antipolar superprediction set
    subdifferential $\subdiff\cvsprt_{\super(\ooneloss^\apolar)}(p)$ will
    not be a singleton. However, we can use the family
    $(\ell_a)_{a\in[-\infty,1]\setminus{0}}$ to approximate $\ooneloss$,
    and therefore approximate the antipolar. That is we can come arbitrarily
    close to obtaining the pair $(\ooneloss(p),\ooneloss^\apolar(p))$ with
    the sequence $(\ell_a(p),\ell_{\!\frac{a}{a-1}}(p))_{a<0}$ for
    $p\in\Rpp^n$. The pointwise limit is illustrated explicitly in
    Figure \ref{fig:approximating_oone_polar}.
\end{example}
 \begin{figure}
	    \centering
    \tikzsetnextfilename{approximating_oone_polar}
    \hspace*{-4ex}
    \begin{tikzpicture}
            \begin{axis}[ xmin=0, xmax=3.5, ymin=0, ymax=3.5, width=0.625\textwidth, height=0.625\textwidth, xlabel={${p_1, \ell(p;\rv y_1)}$}, ylabel={$p_2, \ell(p;{\rv y}_2)$}]
                \coordinate (o) at (axis cs: 0,0);
                \addplot[set, domain=0:1, name path = simplex] ({x},{1-x}) node[pos=0.05, anchor=south west] {$\probm$};
                
                \path[name path=upper_axis] (axis cs:0,10) -- (axis cs:10,10) -- (axis cs:10,0);
                \addplot[set, color=Green, domain=0:1, name path = oonesuper] {-x + 1};
                \addplot[fill=LightGreen] fill between[of=oonesuper and upper_axis];
                
                \pgfmathsetmacro\a{0.45};
                \addplot[set, color=Green!25, domain=0.001:0.999, name path=lp45sup] ({((1-x)^(1/(\a-1)))*((1 - x)^(\a/(\a-1)) + x^(\a/(\a-1)))^(-1/\a) }, {(x^(1/(\a-1)))*((1 - x)^(\a/(\a-1)) + x^(\a/(\a-1)))^(-1/\a) }); 
                \addplot[fill=LightGreen!25] fill between[of=lp45sup and simplex];
                \addplot[set, color=Blue!25, domain=0.0001:0.999, name path=lpp45sup] ({x^(\a-1)*((1-x)^\a + x^\a)^(1/\a-1) }, {((1-x)^(\a-1))*((1-x)^\a + x^\a)^(1/\a-1) }); 
                \addplot[fill=LightBlue!25] fill between[of=lpp45sup and upper_axis];
                
                \pgfmathsetmacro\a{0.6};
                \addplot[set, color=Green!50, domain=0.001:0.999, name path=lp] ({((1-x)^(1/(\a-1)))*((1 - x)^(\a/(\a-1)) + x^(\a/(\a-1)))^(-1/\a) }, {(x^(1/(\a-1)))*((1 - x)^(\a/(\a-1)) + x^(\a/(\a-1)))^(-1/\a) }); 
                \addplot[fill=LightGreen!50] fill between[of=lp and simplex];
                \addplot[set, color=Blue!50, domain=0.0001:0.999, name path = lpp6sup] ({x^(\a-1)*((1-x)^\a + x^\a)^(1/\a-1) }, {((1-x)^(\a-1))*((1-x)^\a + x^\a)^(1/\a-1) }); 
                \addplot[fill=LightBlue!50] fill between[of=lpp6sup and lpp45sup];
                
                \pgfmathsetmacro\a{0.75};
                \addplot[set, color=Green!75, domain=0.001:0.999, name path=lp] ({((1-x)^(1/(\a-1)))*((1 - x)^(\a/(\a-1)) + x^(\a/(\a-1)))^(-1/\a) }, {(x^(1/(\a-1)))*((1 - x)^(\a/(\a-1)) + x^(\a/(\a-1)))^(-1/\a) }); 
                \addplot[fill=LightGreen!75] fill between[of=lp and simplex];
                \addplot[set, color=Blue!75, domain=0.0001:0.999, name path=lpp75sup] ({x^(\a-1)*((1-x)^\a + x^\a)^(1/\a-1) }, {((1-x)^(\a-1))*((1-x)^\a + x^\a)^(1/\a-1) }); 
                \addplot[fill=LightBlue!75] fill between[of=lpp75sup and lpp6sup];
                
                \draw [set, color=Blue , name path = constantsuper] (axis cs:5,1) -- (axis cs:1,1) -- (axis cs: 1,5);
                \addplot[fill=LightBlue] fill between[of=constantsuper and lpp75sup];
                
                \coordinate (p) at (axis cs:0.33333,0.666666);
                \coordinate (lp) at (axis cs:1,0);
                \coordinate (lpp) at (axis cs:1,2);
                
                \coordinate (lp45) at (axis cs:0.368483, 0.104494);
                \coordinate (lp6) at (axis cs:0.603774, 0.106733);
                \coordinate (lp75) at (axis cs:0.854666, 0.0534167);
                
                \coordinate (lpp45) at (axis cs:1.73169, 3.46338);
                \coordinate (lpp6) at (axis cs:1.22363, 2.44726);
                \coordinate (lpp75) at (axis cs:1.04004, 2.08008);
                
                \coordinate (lp45_ray) at ($12*(lp45)$);
                \coordinate (lp6_ray)  at ($12*(lp6)$);
                \coordinate (lp75_ray) at ($12*(lp75)$);

                \draw[faint, name path global=p_ray] (o) -- ($8*(p)$);
                \draw[faint, name path global=p_ray] (o) -- ($12*(lp45)$);
                \draw[faint, name path global=p_ray] (o) -- ($10*(lp6)$);
                \draw[faint, name path global=p_ray] (o) -- ($5*(lp75)$);
                
                \draw[shorten >=-3cm, shorten <=-4cm,faint] ($(o)!(lp45)!(p)$) -- (lp45);
                \draw[shorten >=-3cm, shorten <=-4cm,faint] ($(o)!(lp6)!(p)$) -- (lp6);
                \draw[shorten >=-3cm, shorten <=-4cm,faint] ($(o)!(lp75)!(p)$) -- (lp75);
                \draw[right angle quadrant=2, right angle symbol={o}{p}{lp45},faint];
                \draw[right angle quadrant=2, right angle symbol={o}{p}{lp6},faint];
                \draw[right angle quadrant=2, right angle symbol={o}{p}{lp75},faint];

                \draw[shorten >=-5cm, shorten <=-4cm,faint] ($(o)!(lpp45)!($5*(lp45)$)$) -- (lpp45);
                \draw[shorten >=-5cm, shorten <=-4cm,faint] ($(o)!(lpp6)!($5*(lp6)$)$) -- (lpp6);
                \draw[shorten >=-5cm, shorten <=-4cm,faint] ($(o)!(lpp75)!($5*(lp75)$)$) -- (lpp75);
                \draw[right angle quadrant=2, right angle symbol={o}{lp45_ray}{lpp45},faint];
                \draw[right angle quadrant=2, right angle symbol={o}{lp6_ray}{lpp6},faint];
                \draw[right angle quadrant=2, right angle symbol={o}{lp75_ray}{lpp75},faint];
                
                \draw[shorten >=0cm, shorten >=1.5pt, -latex] (o) -- (p);
                
                \draw (p) node[dot] {} node[anchor=south, yshift=1ex] {$p$};
                \draw (lpp) node[dot] {} node[anchor=east] {$\ooneloss^\apolar°\ooneloss(p)$};
                \draw (lp45) node[dot, color=black!25] {};
                \draw (lp6)  node[dot, color=black!50] {};
                \draw (lp75) node[dot, color=black!75] {};
                \draw (lpp45) node[dot, color=black!25] {};
                \draw (lpp6)  node[dot, color=black!50] {};
                \draw (lpp75) node[dot, color=black!75] {};
            \end{axis}
            \draw (o) node[anchor = north east] {$0$};
            \draw (lp) node[dot] {} node[anchor=north west] {$\ooneloss(p)$};
            \draw[shorten >=0cm, shorten >=1.5pt, -latex] (o) -- (lp);
        \end{tikzpicture}
    \caption{Illustration of Example \ref{ex:approximating_oone_polar} with
    $p\coloneqq(\third, \varfrac{2}{3})$. We can simultaneously approximate
    misclassification loss, $\ooneloss$, and its antipolar, 
    $\ooneloss^\apolar$, over $\probm$ using the concave norm losses.}
    \label{fig:approximating_oone_polar}
\end{figure}
\subsection{Brier Loss} 
\label{ssec:brier_loss}

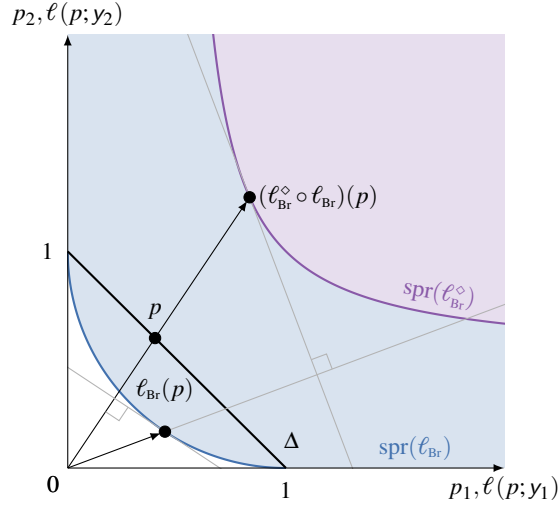
\begin{figure}
    \centering
    \tikzsetnextfilename{brier_loss_polar}
    \begin{tikzpicture}
        \begin{axis}[domain=0:1, xmin=0, xmax=2, ymin=0, ymax=2, width=0.5\textwidth, height=0.5\textwidth, xlabel={${p_1, \ell(p;\rv y_1)}$}, ylabel={$p_2, \ell(p;{\rv y}_2)$}]
            \coordinate (o) at (axis cs: 0,0);
            \path[name path=upper_axis] (axis cs:0,5) -- (axis cs:5,0);

            \addplot[set, color=Blue, name path=l] 
                ({1-x/sqrt(x^2 + (1-x)^2)}, {1-(1-x)/sqrt(x^2 + (1-x)^2)});
            \draw (axis cs:1.8,0) node[anchor=south east, color=Blue] {$\super(\brierloss)$};
            \addplot[fill=LightBlue] fill between[of=l and upper_axis];
            
            \addplot[set,domain=0.2:0.8, color=Magenta, name path global=m] 
                ({(1-x)/(1-(x^2+(1-x)^2))}, {x/(1-(x^2+(1-x)^2))})  
                node[pos=0.26, anchor=south, sloped] {$\super(\brierloss^\apolar)$};
            \addplot[fill=LightMagenta] fill between[of=m and upper_axis];

            \addplot[set, domain=0:1] ({x},{1-x}) node[pos=0.95, anchor=south west] {$\probm$};

            \coordinate (p) at (axis cs:0.4,0.6);
            \coordinate (lp) at (axis cs:0.4453,0.16795);

            \path[name path global=p_ray, draw=none] (o) -- ($5*(p)$);
            \draw[faint] (lp) -- ($5*(lp)$);

            \draw[name intersections={of=p_ray and m, name=llp}] (llp-1) 
                node[anchor=west] {$(\brierloss^\apolar\circ\brierloss)(p)$}; 
            \draw[shorten >=0cm, shorten >=1.5pt, -latex, name path global=lp_ray] (o) -- (lp);
            \draw[shorten >=0cm, shorten >=1.5pt, -latex, name path global=lp_ray] (o) -- (llp-1);

            \draw[shorten >=-4cm, shorten <=-2cm,faint] ($(o)!(lp)!(llp-1)$) -- (lp);
            \draw[right angle quadrant=2, right angle symbol={o}{lp}{llp-1},faint];
            \draw[shorten >=-3cm, shorten <=-5cm,faint] (llp-1) -- ($(o)!(llp-1)!(lp)$);
            \draw[right angle quadrant=2, right angle symbol={o}{llp-1}{lp},faint];    
            
            \draw (p) node[dot] {} node[anchor=south, yshift=1ex] {$p$};
            \draw (lp) node[dot] {} node[anchor=south, yshift=2ex] {$\brierloss(p)$} ;
            \draw (llp-1) node[dot] {};
        \end{axis}
        \draw (o) node[anchor = north east] {$0$};
        \draw (o) node[anchor = north east] {$0$};
    \end{tikzpicture}
    \caption{Brier loss and its antipolar acting on the vector $p\coloneqq(\varfrac{2}{5},\varfrac{3}{5})$.}
    \label{fig:brier}
\end{figure}

The Brier score \citep{Brier:1950aa} is usually defined for $p\in\smplx$ 
in terms of its conditional Bayes risk \citep[Section 5]{van-Erven:2012vz}, 
but for our purposes we 
need to work with the 1-homogeneous extension to $\Rp^n$:
\begin{gather}
    \forall{p\in\Rp^n},\ \cvsprt_{\super(\brierloss)}(p)=\begin{cases}
        \norm1{p}-\frac{\norm2{p}^2}{\norm1{p}} & p\in\Rp^n\setminus{0}\\
        0 & p = 0.
    \end{cases}
    \label{eq:brier_support}
\end{gather}
Indeed we have $\brierloss\in\subdiff\cvsprt_{\super(\brierloss)}$:
\begin{align}
    \forall{p\in\Rpp^n},\ \subdiff\cvsprt_{\super(\brierloss)}(p)
    &=\subdiff(\norm1{})(p)-\oneon{\norm1{p}^2}\g(\norm1{p}·(\nabla\norm2{}^2)(p) 
        + \subdiff(-\norm2{p}^2·\norm1{})(p))\\
    &=\subdiff(\norm1{})(p)
    -\oneon{\norm1{p}^2}\g(\norm1{p}·2p -\norm2{p}^2·\subdiff(\norm1{})(p))\\
    &=\oneon{\norm1{p}}\g(\norm1{p} + \frac{\norm2{p}^2}{\norm1{p}})·\subdiff(\norm1{})(p) -\frac{2p}{\norm1{p}}\\
    &\ni \g(1 + \frac{\norm2{p}^2}{\norm1{p}^2})1_n - 2\frac{p}{\norm1{p}},
\end{align}
where to compute the subdifferential of \eqref{eq:brier_support} we used the
concave subdifferential quotient rule \citep[Theorem~5.2]{mordukhovich1995nonconvex}.
Thus
\begin{gather}
	\forall{p\in\Rpp^n},\ \brierloss(p) = \g(1 +
	\frac{\norm2{p}^2}{\norm1{p}^2})1_n - 2\frac{p}{\norm1{p}}.
\end{gather}
\vspace*{-4mm}
\begin{example}
	\normalfont
	When $n=2$ we can determine the Brier loss antipolar explicitly. Since we
	know $\cvsprt_{\super(\brierloss)}^\apolar$ is 1-homogeneous, it suffices to
	evaluate it on the 2-simplex $\probm$ and then 1-homogeneously extend it.
	Parametrising an element $p\in\probm$ as $p=(p_1,1-p_1)$ it follows that
	\vspace*{-2mm}
    \begin{align}
        \cvsprt_{\super(\brierloss^\apolar)}(p) &=
        \inf_{q\ne 0}
        \frac{\inner{p;q}}{\cvsprt_{\super(\brierloss)}(q)}\label{eq:brier_polar}\\
        &=
        \inf_{q\in\cl{\probm}}
        \frac{q_1 p_1+q_2(1-p_1)}{1 -(q_1^2+q_2^2)} \\
        &=
        \inf_{0≤ q_1 ≤ 1}
        \frac{q_1 p_1+(1-q_1)(1-p_1)}{1 -q_1^2-(1-q_1)^2} .
    \end{align}
    This can be computed directly, resulting in
    \[
	    \forall{p\in\Delta},\ 
            \cvsprt_{\super(\brierloss^\apolar)}(p) = f(p_1)\coloneqq  \frac{(2p_1-1)^2
		    \sqrt{p_1(1-p_1)}}{4p_1^2 + 2\sqrt{p_1(1-p_1)} -4p_1},
    \]
    and thus for $p=\alpha(p_1,1-p_1)\in\Rp^2$, $
    \cvsprt_{\super(\brierloss^\apolar)}(p)=\alpha f(p_1)$\footnote{The 
	    explicit form of the loss itself can be obtained by differentiation
    of $\cvsprt_{\super(\brierloss^\apolar)}(p)$
    and restriction to the simplex. Although little insight seems gleanable
   from the formula, we present it for completeness. We have for all
   $q\in[0,1]$,
   \begin{align}    
	   {\brierloss^\apolar}_{,1}(q) & =\textstyle\frac{(2q-1)\left(2q^2\sqrt{q(1-q)} -2q^2(1-q)
	   +3q(1-q)-(q+1)\sqrt{q(1-q)}\right)}{2\left(2q^2+\sqrt{q(1-q)}
	   -2q\right)^2} \\
	 {\brierloss^\apolar}_{,2}(q) & =
	 \textstyle\frac{(2q-1)\left(2q^2\sqrt{q(1-q)} +2q^2(1-q) -3q\sqrt{q(1-q)} +q(1-q)\right)}{
		 2\left(2q^2+\sqrt{q(1-q)} -2q\right)^2}.
 \end{align}
 }.
    The Brier loss and its polar are illustrated in Figure \ref{fig:brier}.
\end{example}

It does not seem possible to find a closed form for $\brierloss^\apolar$
when $n>2$. However the objective function in \eqref{eq:brier_polar} can be
seen to be quasi-convex in $p$ (since $\cvsprt_{\super(\brierloss)}$ is
concave and positive and thus $\oneon{\cvsprt_{\super(\brierloss)}}$ is quasi-convex)
and therefore is amenable to numerical solution.

\subsection{Cobb-Douglas Loss} 
\label{ssec:cobb_douglas}
As a final example, let $a\in\Rp^n$ and consider the parametrised superlinear function 
\begin{gather}
	\R^n\ni p \mapsto \Def{\psi_a(p)}\coloneqq
    \begin{cases}
        \prod_{i\in [n]} p_i^{\varfrac{a_i}{\norm1{a}}}
                &p\in\Rp^n\\
        -\infty &\text{otherwise.}
    \end{cases}
\end{gather}
\citet{Barbara1994} 
show that $\psi_a$ is ``self-polar'' (cf.\ Remark \ref{rem:polars_closed_under_scalar}) in the sense that
\begin{gather}
	\forall a\in\R_{\ge 0}^n,\ \forall{p\in\Rpp^n},\ 
	\psi_a^\apolar(p) = \frac{\norm1{a}}{\psi_a(a)} \psi_a(p).
    \label{eq:cd_self_polar}
\end{gather}
The function $\psi_a$ can be seen to be the form of the Cobb-Douglas
production function \citep{Cobb:1928aa}, the self-duality of which 
has been an object of considerable interest in microeconomics 
\citep{Houthhakker:1965aa,Samuelson:1965aa,Sato:1976aa}\footnote{It would 
	be of interest to determine other self-dual losses using the
	results of \citep{Houthhakker:1965aa,Samuelson:1965aa,Sato:1976aa}
	and to ascertain the significance (if any) of the self-dual nature
	of the ``boosting loss'' (example \ref{ex:cd_loss}).  
	Observe that for all $a\in\R_{\ge 0}^n$ and all $p\in\Rpp^n$,
	$(\cdloss°\cdloss°\cdloss)(p)=\cdloss(p)$, a fact one can verify
	directly by using \eqref{eq:cd_self_polar}.}.
In order to find the associated loss function we need $\cdloss$ such that
$\super(\cdloss) = \lev_{≥1}(\psi_a^\apolar)$. The self polar property
\eqref{eq:cd_self_polar} makes this easy since 
\begin{gather}
    \cdloss \in \subdiff \psi_a^\apolar \iff \cdloss \in 
        \frac{\norm1{a}}{\psi_a(a)} \subdiff\psi_a.
\end{gather}
Writing the quotient of vectors componentwise, 
$\Def{\Tfrac{a}{p}}\coloneqq
(\varfrac{a_1}{p_1},\dots,\varfrac{a_n}{p_n})$,
since $\psi_a$ is differentiable on its domain we have
\begin{gather}
    \forall{p\in\Rp^n},\ 
    \cdloss(p) = \frac{\norm1{a}}{\psi_a(a)}(\nabla\psi_a)(p) 
    = \frac{\norm1{a}}{\psi_a(a)} \Dfrac{a}{p} \frac{\psi_a(p)}{\norm1{a}}
    = \frac{\psi_a(p)}{\psi_a(a)}\Dfrac{a}{p}. \label{eq:cd_loss}
\end{gather}
Applying the same procedure
to the antipolar loss we have
\begin{gather}
    \forall{p\in\Rp^n},\  \cdloss^\apolar(p) = (\nabla\psi_a)(p) =
    \frac{\psi_a(p)}{\norm1{a}} \Dfrac{a}{p} . \label{eq:cd_loss_apolar} 
\end{gather}

\begin{example}\label{ex:cd_loss}
	\normalfont
    We illustrate the self-duality of $\cdloss$ with a simple example. Set
    $n=2$ and $a_1=a_2=1$ and thus $\psi_a(x)=\sqrt{x_1 x_2}$,
    $\psi_a(a)=1$, and $\norm1{a}=2$. Restricting to $\probm([2])$, and writing
    $p=(p_1,p_2)\in\probm([2])$, from
    \eqref{eq:cd_loss} we have
    \begin{gather}
	    \forall{p\in\probm([2])},\ \cdloss(p) = \Dfrac{\sqrt{p_1 p_2}}{p} 
        = \g(\sqrt{\varfrac{p_2}{p_1}}, \sqrt{\varfrac{p_1}{p_2}}),
    \end{gather}
    which can be recognised as the ``boosting loss'' \citep{Buja:2005}.
    This loss has as its \Def{weight function} \citep{Reid2011}
    $\Def{w(p_1)}\coloneqq-\frac{\partial^2}{\partial p_1^2} \rho(p_1,1-p_1)$, 
    where $\rho=\psi_{a}$ is the concave support function of $\super
    \cdloss$.  We have
    \begin{gather}
        w(p_1)=\frac{1}{(p_1(1-p_1))^{3/2}}.
    \end{gather}
    Using \eqref{eq:cd_loss_apolar} to calculate the antipolar 
    loss $\cdloss^\apolar$ we have
    \begin{gather}
	    \cdloss^\apolar(p) = \frac{\sqrt{p_1p_2}}{2}\Dfrac{1}{p} = \half\cdloss(p).
    \end{gather}
    The superprediction sets associated with the loss $\cdloss$ and its
    antipolar are illustrated in Figure \ref{fig:cd_loss}.
\end{example}

\begin{figure}
    \centering
    \tikzsetnextfilename{cd_loss_polar}
    \begin{tikzpicture}
        \begin{axis}[domain=0.05:0.95, xmin=0, xmax=2, ymin=0, ymax=2, width=0.5\textwidth, height=0.5\textwidth,
            xlabel={${p_1, \ell(p;\rv y_1)}$}, ylabel={$p_2, \ell(p;{\rv y}_2)$}]
            \coordinate (o) at (axis cs: 0,0);
            \path[name path=upper_axis] (axis cs:0,5) -- (axis cs:5,0);

            \addplot[set, color=Blue, name path global=l] 
                ({sqrt(1/x -1)/2}, {sqrt(x/(1-x))/2}) 
                node[pos=0.12, anchor=south, sloped] {$\super(\cdloss)$};
            \addplot[fill=LightBlue] fill between[of=l and upper_axis];
            
            \addplot[set, color=Magenta, name path global=m] 
                ({sqrt(1/x -1)}, {sqrt(x/(1-x))})  
                node[pos=0.385, anchor=south, sloped] {$\super(\cdloss^\apolar)$};

            \addplot[fill=LightMagenta] fill between[of=m and upper_axis];

            \addplot[set, domain=0:1] ({x},{1-x}) node[pos=0.95, anchor=south west] {$\probm$};

            \coordinate (p) at (axis cs:0.7,0.3);
            \coordinate (lp) at (axis cs:0.327327,0.763763);

            \path[name path global=p_ray, draw=none] (o) -- ($5*(p)$);
            \draw[faint] (lp) -- ($5*(lp)$);

            \draw[name intersections={of=p_ray and m, name=llp}] (llp-1) 
                node[anchor=south, yshift=3ex] {$(\cdloss^\apolar\circ\cdloss)(p)$}; 
            \draw[shorten >=0cm, shorten >=1.5pt, -latex, name path global=lp_ray] (o) -- (lp);
            \draw[shorten >=0cm, shorten >=1.5pt, -latex, name path global=lp_ray] (o) -- (llp-1);

            \draw[shorten >=-4cm, shorten <=-2cm,faint] ($(o)!(lp)!(llp-1)$) -- (lp);
            \draw[right angle quadrant=2, right angle symbol={o}{lp}{llp-1},faint];
            \draw[shorten >=-3cm, shorten <=-5cm,faint] (llp-1) -- ($(o)!(llp-1)!(lp)$);
            \draw[right angle quadrant=2, right angle symbol={o}{llp-1}{lp},faint];    
            
            \draw (p) node[dot] {} node[anchor=south, yshift=1ex] {$p$};
            \draw (lp) node[dot] {} node[anchor=west] {$\cdloss(p)$} ;
            \draw (llp-1) node[dot] {};
        \end{axis}
        \draw (o) node[anchor = north east] {$0$};
    \end{tikzpicture}
    \caption{Illustration of the self-dual nature of the Cobb-Douglas loss
	    $\cdloss$
	    with $a \coloneqq (\half, \half)$ and $p\coloneqq(\varfrac{7}{10},\varfrac{3}{10})$.}
    \label{fig:cd_loss}
\end{figure}


\section{Designing Losses via their Superprediction Sets} 
\label{sec:designing_losses_via_their_superprediction_sets}

Loss functions are clearly essential for machine learning, but they are
often take for granted, their choice being primarily a consequence of
convenience or familiarity. We posit that this is due in part to a lack of
a tools for designing and tuning them. In this section we offer some
starting points for such a tuning exercise.

The conventional approach to working with loss functions is to focus on the
analytic form of the mappings $(p,\rv y) \mapsto \ell(p,\rv y)$. In this
section we show some examples of the power of instead working with the
family $\prop(\Rp^n)$ and deriving the associated loss functions via the
subdifferential of the concave support functions. 

\subsection{Canonical Normalisation} 
\label{sub:canonical_normalisation}

One problem that presents itself when working with the family
$\prop(\Rp^n)$ is that of normalisation. In
\S\ref{sec:polar_applications} the lack of consistency of normalisation
between $(\ell_a)$, $\brierloss$ and $\cdloss$ made it difficult to compare
these losses side by side. Our proposed normalisation for a loss $\ell$ is
to pick $p_\ell\in\Rpp^n$ and $\alpha>0$ such that
$\cvsprt_{\super(\alpha\ell)}(p_\ell) = 1$, a task that the superprediction
set machinery makes very simple. There are a couple of ways one might choose
$p_\ell$, the simplest is $p_\ell \coloneqq 1_n$ for all loss functions $\ell$.
However, a more robust choice is $p_\ell \in
\argmax_{p\in\probm}\cvsprt_{\super(\ell)}(p)$. We say a loss function $\ell$
is \Def{normalised} if its associated conditional Bayes risk function
attains a maximum value of 1.
For several of our results we need a non-compact version of the
\citet{Sion1958} minimax theorem. The following result attributed to
\citet{ha1981noncompact} is originally shown in a much more general setting
and so we state it for the space $\bnch$ below.

\begin{lemma}[\protect{\citealt[Theorem~2]{ha1981noncompact}}]\label{thm:minimax}
        Let $X$ and $Y$ each be nonempty convex subsets of $\bnch$. Let
	$f\colon X \times Y\to \R$ be such that 
        \begin{enumerate}
            \item For each $x\in X$, $y\mapsto f(x,y)$ is lower 
		    semi-continuous and quasi-convex;
            \item For each $y\in Y$, $x\mapsto f(x,y)$ is upper 
		    semi-continuous and quasi-concave.
        \end{enumerate}
	If there exists a nonempty convex set $X'\subseq X$ and a compact
	set $Y'\subseq Y$ such that
        \begin{gather}
            \inf_{y\in Y}\sup_{x\in X} f(x,y) ≤ \inf_{y\in Y'}\max_{x\in X'} 
	        f(x,y),\label{eq:minimax_cond}
            \shortintertext{then}
            \inf_{y\in Y}\sup_{x\in X} f(x,y) = \sup_{x\in X}\inf_{y\in Y} f(x,y).
        \end{gather}
\end{lemma}
\begin{theorem}\label{thm:cvsprt_maximum}
    Let $\ell$ be a proper loss and let $p^\star \in \argmax_{p\in\probm}
    \cvsprt_{\super(\ell)}(p)$. Then 
    \begin{gather}
	 \subdiff\cvsprt_{\super(\ell)}(p^\star) \ni
	 \frac{1_n}{\agauge_{\super(\ell)}(1_n)} 
	 = \frac{1_n}{\cvsprt_{\super(\ell^\apolar)}(1_n)}.
    \end{gather} 
\end{theorem}
\begin{proof}
    First we apply Lemma \ref{thm:minimax} to establish
    \begin{gather}
        \max_{p\in\probm}\inf_{z\in\super(\ell)}\inner{z;p} 
        = \inf_{z\in\super(\ell)}\max_{p\in\probm}\inner{z;p}.
        \label{eq:minimax_support}
    \end{gather}
    For $\alpha \geq \oneon{\agauge_{\super(\ell)}(1_n)}$ 
    we have $\alpha 1_n \in\super(\ell)$, and
    \begin{gather}
        \exists{\beta\geq 1},\ 
        \inf_{z\in\super(\ell)}\max_{p\in\probm}\inner{z;p} ≤ 
	  \inf_{z \in\single{ \beta \alpha 1_n}} \max_{p\in\probm}\inner{z;p},
    \end{gather}
    since the left hand side is finite. Thus we have demonstrated the
    sufficient condition for \eqref{eq:minimax_cond}. Since $\super(\ell)$
    and $\probm$ are convex, we have satisfied the conditions for Lemma
    \ref{thm:minimax} and shown \eqref{eq:minimax_support}. 
    
    Therefore 
    \begin{gather}
        \max_{p\in\probm}\cvsprt_{\super(\ell)}(p) = 
	  \max_{p\in\probm}\inf_{z\in\super(\ell)}\inner{z;p} 
        \overset{\eqref{eq:minimax_support}}{=} 
	  \inf_{z\in\super(\ell)}\max_{p\in\probm}\inner{z;p}.
    \end{gather}
    Observe $\max_{p\in\probm}\inner{z;p} = \max_{\rv y \in \cal Y} z_{\rv
    y}$. A proof by contradiction easily confirms that for $z \in
    \subdiff\cvsprt_{\super(\ell)}(p^\star)$ we have $z_{\rv y } = z_{\rv
    z}$ for all $\rv y,\rv z \in \cal Y$. Thus the minimising $z \in
    \subdiff\cvsprt_{\super(\ell)}(p^\star)$ is a multiple of $1_n$, more
    precisely
    \begin{gather}
        z = \inf\set{\lambda > 0 ; \lambda 1_n \in \super(\ell)} · 1_n =
	1_n · \oneon{\beta_{\super(\ell)}(1_n)},
    \end{gather}
    which completes the proof.
\end{proof}
By Theorem \ref{thm:cvsprt_maximum} we see that the maximum of
$\cvsprt_{\super(\ell)}$ occurs for $p^\star\in\probm$ such that
$\ell(p^\star) = \alpha 1_n$ where $\alpha \coloneqq
\oneon{\agauge_{\super(\ell)}(1_n)}$. If we normalise a proper loss
$\ell$ with the coefficient $c \coloneqq \agauge_{\super(\ell)}(1_n)$, then
evaluating the conditional Bayes risk at $p^\star\in\probm$ we have
\begin{gather}
    \cvsprt_{\super(c\ell)}(p^\star) = c \cvsprt_{\super(\ell)}(p^\star) 
    \overset{\mathrm{T}\ref{thm:cvsprt_maximum}}{=} 
    c \inner*{\varfrac{1_n}{\agauge_{\super(\ell)}(1_n)} ; p^\star}
     =  \frac{c}{\agauge_{\super(\ell)}(1_n)} \inner{ 1_n; p^\star}
     = 1.
\end{gather}

The following corollary demonstrates another application of the polar
loss, that is the uniform loss vector $1_n$ is achieved by
the prediction that maximises the conditional Bayes risk.
\begin{corollary}\label{cor:cvsprt_maximum}
    Let $\ell$ be a proper loss and  
    $p^\star\coloneqq\varfrac{\ell^\apolar(1_n)}{\norm1{\ell^\apolar(1_n)}}$. Then 
    $p^\star \in \displaystyle\argmax_{p\in\probm} \cvsprt_{\super(\ell)}(p)$.
\end{corollary}

\begin{corollary} A proper loss function $\ell$ is normalised if and only
	if $1_n\in\bd(\super\ell)$.
\end{corollary}




We give the normalisation coefficients for the common losses from Table
\ref{tab:common_losses} in Table \ref{tab:normalised_common_losses}, and
plot their conditional Bayes risk functions and superprediction sets in
Figure \ref{fig:normalised_common_losses} (the overbar denotes this
normalisation). With the normalised versions of these loss functions we can
now see that $\cdloss*_{1_n}$ is attained as the limit $\lim_{a\to
0}\bar\ell_a$.

\begin{table}
	    \arrayrulecolor{lightgray}
    \centering
    \small
    \def\unif{(\oneon n,\dots,\oneon n)}
    \begin{tabular}{C{2.5cm} c c c c c c} \toprule
Name        & Symbol        & Maximiser & Coefficient    & Normalised Loss \\ \midrule
Misclassification Error  (0/1) 
            & $\expandafter\bar\ooneloss$ 
                            & $\unif$  & $n$ & $n\ooneloss(p,\rv y)$&
			    \\[2mm]
Logarithmic & $\expandafter\bar\logloss$ 
                            & $\unif$  & $\oneon{\log(n)}$ &
			    $-\oneon*{\log(n)}\log(p_{\rv y})$   \\[1mm]
Concave Norm {\scriptsize$a\in[-\infty, 1]\setminus{0}$}
            & $\bar\ell_a$ &  $\unif$  & $\sqrt[a]{n}$         &
	    $\sqrt[a]{n}·\g(\frac{p_{\rv
	    y}}{\beta_{\!\!\frac{a}{a-1}}(p)})^{\frac{1}{a-1}}$ \\[2mm]
Brier       & $\expandafter\bar\brierloss$ 
                           &  $\unif$  & $\frac{n}{n-1}$       &
			   $\frac{n}{n-1}·\g((1 + \norm2{p}^2)·1_n -
			   2p_{\rv y})$ \\[1mm]
Cobb--Douglas {\scriptsize$a\in\Rp^n$}
            & $\cdloss*$    
                           &  $(a_1,\dots,a_n)$ & $\frac{\norm1{a}}{\psi_a(a)}$
                                                                & $\norm1{a}\psi_a\g(\varfrac{p}{a^2})·\frac{1}{p_{\rv y}}$  \\
\bottomrule
    \end{tabular}
    \caption{Canonical normalisations of the loss functions from
	    Table \ref{tab:common_losses}.}
    \label{tab:normalised_common_losses}
\end{table}

\begin{figure}
    \centering
    \subcaptionbox{Conditional Bayes risk for $\cdloss$:
	    graph of $\cvsprt_{\super(\bar{\ell}_a)}$ for
	    $a=(\alpha,\alpha)$, with  $\alpha>0$ and
	    $\alpha\in I\coloneqq\{0.9525^i\st
	    i\in[40]\}$ (blue); and $\alpha<0$  with 
	    $\alpha=\beta/(\beta-1)$, $\beta\in I$ (green).}{
    \tikzsetnextfilename{norm_ccv_conditional_bayes_risk}
    \hspace*{-4ex}
    \begin{tikzpicture}
        \begin{axis}[domain=0:1, xmin=0, xmax=1, ymin=0, ymax=1.01,
		width=0.5\textwidth, height=0.5\textwidth, xlabel={$p_1$}, ylabel={$\cvsprt_{\super(\bar\ell_a)}(p_1,1-p_1)$}, xtick={0,0.25,0.5,0.75,1}, ytick=, inner axis line style={-}, xtick style={draw}, ytick style={draw}, xlabel style={anchor=west}, ylabel style={anchor=south}]
            \pgfplotsinvokeforeach{1,...,40}{
                \pgfmathsetmacro\a{0.9525^(41-#1)}
                \pgfmathtruncatemacro{\pcnt}{100*0.9^(41-#1)}
                \edef\temp{\noexpand%
                   \addplot[color=Blue!\pcnt, domain=0.00001:1] {
                        (2^((\a-1)/\a))*((1 - x)^\a + x^\a)^(1/\a)
                    };
                }\temp
            }
            \pgfplotsinvokeforeach{1,...,40}{
                \pgfmathsetmacro\a{0.925^(41-#1)}
                \pgfmathtruncatemacro{\pcnt}{100*0.9^(41-#1)}
                \edef\temp{\noexpand%
                    \addplot[color=Green!\pcnt, domain=0.0001:1] {
                        (2^(1/\a))*((1-x)^(\a/(\a-1)) + x^(\a/(\a-1)))^((\a-1)/\a)
                    };
                }\temp
            }
            \addplot[color=Green, domain=0:0.5] {2*x};
            \addplot[color=Green, domain=0.5:1] {-2*x + 2};
            \addplot[color=Blue]{1};
        \end{axis}
    \end{tikzpicture}}\hspace*{-4ex}
    \subcaptionbox{Superprediction sets for $\cdloss$ for same parameter
    range as in (a).}{
    \tikzsetnextfilename{norm_ccv_superprediction_sets}
    \begin{tikzpicture}
        \begin{axis}[domain=0.01:0.99, xmin=0, xmax=3, ymin=0, ymax=3,
		width=0.5\textwidth, height=0.5\textwidth, xlabel={${p_1, \bar\ell(p;\rv y_1)}$}, ylabel={$p_2, \bar\ell(p;{\rv y}_2)$}]
            \path[name path global=upper_axis] (axis cs:0,8) -- (axis cs:8,8) -- (axis cs:8,0);
            \path[name path global=upper_axis2] (axis cs:8,0) -- (axis cs:8,8) -- (axis cs:0,8);
            \addplot[set, domain=0:1, color=Green, name path global=dual] ({2*2*x}, {2*(1-2*x)});
            \draw[set, color=Blue, name path global=primal] (axis cs:1,8) -- (axis cs:1,1) -- (axis cs:8,1);
            \pgfplotsinvokeforeach{1,...,15}{
                \pgfmathsetmacro\a{0.8^(16-#1)}
                \pgfmathtruncatemacro{\pcnt}{100*0.75^(15-#1)}
                \edef\temp{\noexpand%
                    \addplot[set, color=Blue!\pcnt, domain=0.0001:0.9999, name path global=primal_#1](
                        {exp( ((\a-1)/\a)*ln(2) + (\a-1)*ln(x)   + (1/\a-1)*ln((1-x)^\a + x^\a) )},
                        {exp( ((\a-1)/\a)*ln(2) + (\a-1)*ln(1-x) + (1/\a-1)*ln((1-x)^\a + x^\a) )}
                    );
                }\temp
            }
            \pgfplotsinvokeforeach{1,...,15}{
                \pgfmathsetmacro\a{0.8^(16-#1)}
                \pgfmathtruncatemacro{\pcnt}{100*0.75^(16-#1)}
                \edef\temp{\noexpand%
                    \addplot[set, color=Green!\pcnt, domain=0.001:0.99, name path global=dual_#1] (
                        {2^(1/\a)*x^(1/(-1 + \a))*((1 - x)^(\a/(-1 + \a)) + x^(\a/(-1 + \a)))^(-1/\a)},
                        {2^(1/\a)*(1 - x)^(1/(-1 + \a))*((1 - x)^(\a/(-1 + \a)) + x^(\a/(-1 + \a)))^(-1/\a)}
                    );
                }\temp
            }
            \addplot[fill=LightGreen] fill between [of=dual and upper_axis];
            \pgfplotsinvokeforeach{15,...,1}{
                \pgfmathtruncatemacro{\pcnt}{100*0.75^(16-#1)}
                \edef\temp{\noexpand%
                    \addplot[fill=LightGreen!\pcnt] fill between [of=dual_#1 and upper_axis2];
                }\temp
            }
            
            \addplot[fill=white] fill between [of=dual_1 and primal_1];
            
            \pgfplotsinvokeforeach{1,...,15}{
                \pgfmathtruncatemacro{\pcnt}{100*0.75^(16-#1)}
                \edef\temp{\noexpand%
                    \addplot[fill=LightBlue!\pcnt] fill between [of=primal_#1 and upper_axis2];
                }\temp
            }
            \addplot[fill=LightBlue] fill between [of=primal and upper_axis];
            
            \coordinate (o) at (axis cs: 0,0);e
            \draw[faint] (o) -- (axis cs: 5,5);
            \coordinate (un) at (axis cs: 1,1);
            \draw[dashed, faint] ($(o)!(un)!(axis cs:0,5)$) -- (un) -- ($(o)!(un)!(axis cs:5,0)$);
            \draw (un) node[dot] {};
        \end{axis}
        \draw (o) node[anchor = north east] {$0$};
    \end{tikzpicture}
    }
    
    \bigskip

    \tikzsetnextfilename{other_normalised_conditional_Bayes_risk}
    \subcaptionbox{Comparison of conditional Bayes risks for  
       $\cdloss*$ with $a=(1/2,1/2)$ (green), $\expandafter\bar\brierloss$ (red),
       and $\expandafter\bar\logloss$ (blue). }{\hspace{-4ex}
    \begin{tikzpicture}
        \begin{axis}[domain=0:1, xmin=0, xmax=1, ymin=0, ymax=1.01,
			width=0.5\textwidth, height=0.5\textwidth, xlabel={$p_1$}, 
        ylabel={$\cvsprt_{\super(\bar\ell)}(p_1,1-p_1)$}, xtick={0,0.25,0.5,0.75,1}, ytick=, inner axis line style={-}, xtick style={draw}, ytick style={draw}, xlabel style={anchor=west}, ylabel style={anchor=south, align=center}]
            \addplot[color=Blue, domain=0:1] {-(1/ln(2))*(x*ln(x) + (1-x)*ln(1-x))};
            \addplot[color=Green, domain=0:1] {2*(1-x^2-(1-x)^2)};
            \addplot[color=Red, domain=0:1] {2*sqrt(x*(1-x))};
        \end{axis}
    \end{tikzpicture}}
    \hspace{-4ex}
    \subcaptionbox{Comparison of superprediction sets for  
       $\cdloss*$ with $a=(1/2,1/2)$ (green), $\expandafter\bar\brierloss$ (red),
       and $\expandafter\bar\logloss$ (blue).}{
    \tikzsetnextfilename{other_normalised_superprediction_sets}
    \begin{tikzpicture}
        \begin{axis}[domain=0.01:0.99, xmin=0, xmax=3, ymin=0, ymax=3,
			width=0.5\textwidth, height=0.5\textwidth,
            xlabel={${p_1, \bar\ell(p;\rv y_1)}$}, ylabel={$p_2, \bar\ell(p;{\rv y}_2)$}]
            \coordinate (o) at (axis cs: 0,0);
            \path[name path=upper_axis] (axis cs:0,5) -- (axis cs:5,5) -- (axis cs:5,0);

            \addplot[set, color=Green, domain=0:1, name path=cd]       ({2*sqrt(1/x - 1)/2}, {2*sqrt(x/(1-x))/2});
            \addplot[set, color=Blue, domain=0.01:0.99, name path=log] ({-ln(x)/ln(2)}, {-ln(1-x)/ln(2)});
            \addplot[set, color=Red, domain=0.01:0.99, name path=brier] ({2*(1+x^2+(1-x)^2-2*x)}, {2*(1+x^2+(1-x)^2 - 2*(1-x))});
            \addplot[fill=LightRed]   fill between[of=brier and log];
            \addplot[fill=LightBlue]  fill between[of=log and cd];
            \addplot[fill=LightGreen] fill between[of=cd and upper_axis];

            \draw[faint] (o) -- (axis cs: 5,5);
            \coordinate (un) at (axis cs: 1,1);
            \draw[dashed, faint] ($(o)!(un)!(axis cs:0,5)$) -- (un) -- ($(o)!(un)!(axis cs:5,0)$);
            \draw (un) node[dot] {};
        \end{axis}
        \draw (o) node[anchor = north east] {$0$};
    \end{tikzpicture}
    } 
    \caption{Normalised loss functions. 
    }\label{fig:normalised_common_losses}
\end{figure}


\subsection[Shifting the Maximum]{Shifting the Maximum} 
\label{sec:shifting_the_max}
In \S\ref{sub:canonical_normalisation} we saw that the conditional Bayes
risk of a proper loss $\ell$ is maximised over the probability simplex at
$p^\star\coloneqq\varfrac{\ell^\apolar(1_n)}{\norm1{\ell^\apolar(1_n)}}$
(Corollary \ref{cor:cvsprt_maximum}). The question naturally arises then of how
one might one modify $\ell \mapsto \check\ell$ to reposition the maximum to
an arbitrary $p^0\in\probm$\footnote{The motivation for doing so arises
	from considering the cost-sensitive missclassification losses $\ell_c$,
	$c\in(0,1)$ \citep[Section 5.2]{Reid2011}, whose conditional Bayes
	risks are $\minL_c(p)=(1-p)c \wedge (1-c) p$. The maximum of
	$\minL_c(p)$ over $p$ occurs at $c$ (although the maximum value
	does not remain constant as $c$ varies). The corresponding losses $\ell_c$
	allow one to impose a different cost for false positives and false
	negatives. Thus shifting the maximum of a general loss allows one
	to reweight the costs for the different types of prediction error one might
	make.}. 
That is, $p^\star\mapsto
{\check\ell}^\apolar(1_n)=p^0$. If one is given only $\ell$ or $\minL$ this
is not obvious. However, the answer is simple in terms of
$S\coloneqq\super(\ell)$. Before proceeding we note that this problem is not well
posed since we have not defined what exactly we are hoping to retain of the
original function $\ell$. That said, our construction entails---more or
less---the minimum required perturbation of $S$ in order to endow
$\check\ell$ with the necessary desiderata.

To solve this question we construct a new super prediction set $\check S$
from $S$ and define $\check\ell \in \subdiff\cvsprt_{\check S}$. The family
$\prop(\Rp^n)$ is a cone since it is closed under positive scalar
multiplication and addition of sets from the family. Thus if we construct a
mapping $S\mapsto \check S \coloneqq \alpha S + x^*$, where $\alpha>0$ and
$x^*\in\Rp^n$ we can easily ensure $\check S \in \prop(\Rp^n)$.

In order to move the maximiser from $p^\star$ to $p^0$ it suffices to
translate the set $S$ by $\ell(p^\star)-\ell(p^0)$. However
$-\ell(p^0)\notin \Rp^n$, and so we ``push'' the vector $-\ell(p^0)$ into
$\Rp^n$ by adding just enough of the constant loss: $\alpha \cdot 1_n$,
where $\alpha\coloneqq \max_{\rv y \in
\cal Y}\ell(p^0,\rv y)$. This has a neutral effect on the $\argmax$. We
now have
\begin{gather}
	\check S \coloneqq \frac{1}{\agauge(\ell(p^0))} \left(
	S + \ell(p^\star) - \ell(p^0) + \alpha\cdot 1_n\right),
    \label{eq:shifting_the_maximum_s}
\end{gather}
where the term $\oneon{\agauge_S(\ell(p^0))}$ normalises $\check S$ so that
$\max_{p\in\probm}\agauge_{S}(p) = \max_{q\in\probm}\agauge_{\check S}(q)$.
The normalisation can easily be calculated using the identity $\ell(p^0) =
\agauge_S(\ell(p^0))·\ell(p^\star)$.

Using the calculus of support functions,
\begin{gather}
    \cvsprt_{\check S} = \oneon{\agauge_S(\ell(p^0))}\g(\cvsprt_{S}+
    \inner{\ell(p^\star) - \ell(p^0);\marg} + \alpha\norm1{}),
    \shortintertext{and}
    \check\ell(q)
    = \oneon{\agauge_S(\ell(p^0))}\g(\ell(q)+ \ell(p^\star) -
    \ell(p^0) + \alpha\cdot 1_n).
    \label{eq:shifting_the_maximum_ell}
\end{gather}
\begin{figure}  
    \centering
    \subcaptionbox{The effect of
	    $\logloss\mapsto\expandafter\check\logloss$ on the conditional
	    Bayes risk function. The original conditional Bayes risk is
	    dashed.\label{fig:shifting_the_maximum1}}
    {\tikzsetnextfilename{shifting_the_maximum1}\hspace*{-4ex}\begin{tikzpicture}
            \begin{axis}[domain=0:1, xmin=0, xmax=1, ymin=0, ymax=0.79, width=0.5\textwidth, height=0.5\textwidth, xlabel={$p_1$}, 
            ylabel={$\cvsprt_{S}(p_1,1-p_1)$,\\$\cvsprt_{\check S}(p_1,1-p_1)$},
            xtick={0,0.25,0.5,1.0},
            axis y line=middle, axis x line=bottom,
            ytick=,
            extra y ticks={0.693147},
            extra y tick label={$\log 2$},
            inner axis line style={-},
            xtick style={draw},
            ytick style={draw},
            xlabel style={anchor=west},
            ylabel style={anchor=south,align=center},
            ]
                \coordinate (ell_old) at (axis cs:0.5,.6931471806);
                \coordinate (ell_new) at (axis cs:0.25,0.6931471806);
                \addplot[dashed]{-x*ln(x) - (1-x)*ln(1-x)};
                \addplot{(x*ln(4/3) + ln(6) + (x-1)*ln(1 - x) - x*ln(4*x))/3};
                \draw (ell_old) node[dot] {};
                \draw (ell_new) node[dot] {};
            \end{axis}
        \end{tikzpicture}}
    \hspace{-4ex}
    \subcaptionbox{The corresponding superprediction set operation.\label{fig:shifting_the_maximum2}}
    {\tikzsetnextfilename{shifting_the_maximum2}\begin{tikzpicture}
            \begin{axis}[domain=0.1:0.9, xmin=0, xmax=1.5, ymin=0, ymax=1.5, width=0.5\textwidth, height=0.5\textwidth, xlabel={${p_1, \ell(p;\rv y_1)}$}, ylabel={$p_2, \ell(p;{\rv y}_2)$},
            xtick={0,1},
            ytick={0,1}]
                \coordinate (o) at (axis cs: 0,0);
                \path[name path=upper_axis] (axis cs:0,5) -- (axis cs:5,0);
                \path[name path=upper_axis2] (axis cs:0.707107,5) -- (axis cs:5,0.707107);
                
                \coordinate (max_old) at (axis cs:0.5,0.5);
                \coordinate (ell_max) at (axis cs:0.6931471806,0.6931471806);
                \draw[faint, name path=p_ray] (o) -- ($8*(max_old)$);
                
                \coordinate (max_new) at (axis cs:0.25,0.75);
                \coordinate (eight_mn) at (axis cs:2,6);
                \draw[faint] (o) -- (eight_mn);

                \addplot[set, color=Blue, name path=ll,domain=0.001:0.99]
		({(ln(2) - ln(x))/3}, {(ln(6) - ln(1-x))/3}) node[pos=0.35,
		anchor=south, sloped] {$\super(\expandafter\check\logloss)$}; ; 
                \addplot[fill=LightBlue] fill between[of=ll and upper_axis];
                
                \addplot[set, color=Blue, dashed, name path=l] ({-ln(x)},{-ln(1-x)})  node[pos=0.3, anchor=south, sloped] {$\bd(\super\logloss)$};

                \addplot[set, domain=0:1] ({x},{1-x}) node[pos=0.95, anchor=south west, xshift=1em, prominent] {$\probm$};
            
                \draw[shorten >=-8cm, shorten <=-4cm,faint] ($(o)!(ell_max)!(eight_mn)$) -- (ell_max);
                \draw[right angle quadrant=2, right angle symbol={o}{eight_mn}{ell_max}, faint]; 
                
                \draw (max_old) node[dot] {} node[anchor=east] {$p^\star$};
                \draw (ell_max) node[dot] {} node[anchor=west, xshift=2ex, yshift=2ex] {$\ell(p^\star) = \ell(p^0)$};
                \draw (max_new) node[dot] {} node[anchor=east, xshift=-1ex] {$p^0$}; 
            \end{axis}
            \draw (o) node[anchor = north east] {$0$};
        \end{tikzpicture}}
    
    \caption{Illustration of Example \ref{ex:shifting_the_max}, shifting
    the maximum of the conditional Bayes risk function associated with
    $\logloss$. The modified conditional Bayes risk
    $\cvsprt_{\super(\expandafter\check\logloss)}$ attains its maximum at $(\quarter,\varfrac{3}{4})$.}
\end{figure}

\begin{example}\label{ex:shifting_the_max}
	\normalfont
    We will now apply the operations \eqref{eq:shifting_the_maximum_s} and
    \eqref{eq:shifting_the_maximum_ell} to $\logloss$ with $\cal Y
    \coloneqq [2]$
    and $p^0 \coloneqq (\quarter, \varfrac{3}{4})$. We know
    $\cvsprt_{\super(\logloss)}$ achieves its maximum at the uniform
    prediction: $\ell^\apolar(1_n) = (\half,\half)$. To shift the maximum
    to $p^0$ we define the new loss
        $\forall{p\in\probm},$
    \begin{align}
        \expandafter\check\logloss(p) 
        &= 
        \frac{1}{\agauge_{\super(\logloss)}(p^0)}
            \g\bigg(\logloss(p) 
                + \logloss(p^\star) 
                - \logloss(p^0) 
                - (\max_{\rv y' \in \cal Y} \logloss(p^0;\rv y'))\cdot 1_n
                )\\
        &= 1.7783 \g\bigg(\logloss(p) 
                + \logloss(1_n/2) - \logloss(p^0) +\log(4)\cdot 1_n).
    \end{align}
    The effect of
    $\cvsprt_{\super(\logloss)}\mapsto\cvsprt_{\super(\expandafter\check\logloss)}$
    is illustrated in Figure \ref{fig:shifting_the_maximum1}. The
    corresponding superprediction set operation is illustrated in
    Figure \ref{fig:shifting_the_maximum2}.
\end{example}


\subsection{Building Losses From Norms} 
\label{sub:building_losses_from_norms}

In \S\ref{sec:gauge_functions_and_polar_duality} we looked at norms as
gauge functions, and saw that the antigauge functions naturally give rise
to the notion of an antinorm. In \S\ref{sec:loss_functions} we saw that
these antinorms are precisely the conditional Bayes risk functions. In this
section we will see that there is a natural injection---or family
thereof---between the symmetric radiant sets and  the shady sets. In doing
so we define a new family of bounded, proper loss functions: the
\emph{norm losses}. 

Recall (\S\ref{sec:cones}) $X_+\subset\bnch$ denotes a
salient, closed, convex cone, and $X_+^*$ denotes its dual cone.   The
following result provides a means to take a symmetric radiant set to generate a
superprediction set for a proper loss.

\begin{figure}
    \centering
    \tikzsetnextfilename{rad_to_shad}
    \begin{tikzpicture}
        \begin{axis}[xmin=-0.5, xmax=3.0, ymin=-0.5, ymax=3.0, width=0.625\textwidth, height=0.625\textwidth, xlabel={$p_1, \ell(p;\rv y_1)$}, ylabel={$p_2, \ell(p;{\rv y}_2)$},
        ]
            \coordinate (o) at (axis cs: 0,0);
            \coordinate (u) at (axis cs:1,1);
            \path[name path=upper_axis] (axis cs:0.707107,5) -- (axis cs:5,5) -- (axis cs:5,0.707107);
            
            \draw[set, Cyan, name path=left_s] 
                (axis cs:0.707107,1.707107) -- (axis cs:0.707107,3);
            \draw[set, Cyan, name path=right_s] 
                (axis cs:1.707107,0.707107) -- (axis cs:3,0.707107);
            \addplot[set, Cyan, name path=round_s, domain=pi:1.5*pi,samples=300]({cos(deg(x)) + 1.707107},{sin(deg(x)) + 1.707107});
            
            \addplot[fill=LightCyan] fill between[of=round_s and upper_axis];
            \addplot[fill=LightCyan] fill between[of=right_s and upper_axis];
            
            \addplot[set,color=Blue, fill=LightBlue, domain=0:2*pi,samples=300]({cos(deg(x)) + 1.707107},{sin(deg(x)) + 1.707107});
            
            \addplot[set, color=Blue, fill=LightBlue, domain=1.5*pi:4*pi,samples=600]({cos(deg(x))},{sin(deg(x))});
            
            \draw[dashed, faint] ($(o)!(u)!(axis cs:0,5)$) -- (u) -- ($(o)!(u)!(axis cs:5,0)$);
            \path[draw,faint] ($-1*(u)$) -- (o) -- ($5*(u)$);
            \draw (axis cs:0.707107,0.707107) node[dot] {} node[anchor=north, xshift=-2ex] {$ 2^{-\half}1_2$};
            \draw[Cyan] (axis cs:3,0.707107) node[anchor=north east] 
                {$S_2$};
            \draw[Blue, yshift=1em] (axis cs:1.707107,0.707107) node[anchor=south] 
                {$B_2 + (1 + 2^{-\half})1_2$};
            \draw[Blue] (axis cs:1,0) node[anchor=south east] {$\ball_2$};
            \path[draw] (u) node[dot] {} node[anchor=south, yshift=1ex] {$1_2$};
        \end{axis}
        \draw (o) node[anchor = north east] {$0$};
    \end{tikzpicture}
    \caption{Illustration of Theorem \ref{thm:rad_to_shad} using $\ball_2$. We
    translate $\ball_2$ ``northeast'' using the vector $x^*\coloneqq  (1 +
    2^{-\half})·1_2$. This ensures $S_2\coloneqq \ball_2 +  (1 + 2^{-\half})·1_2+\Rp^2\subseq\Rp^2$. }
    \label{fig:rad_to_shad}
\end{figure}

\begin{theorem}\label{thm:rad_to_shad}
    Let $R \in\rdnt(\bnch)$ be symmetric. Choose $x\in\pcone$ with 
    $x\in\inter_{r\in R\cap\pcone}(\pcone + r)$ and $x\notin R$. Then \begin{enumerate}
            \item $R+x + \pcone\subseq\pcone$, and 
             \label{thm:rad_to_shad_subseq}
            \item $R+x+\pcone\in\prop(\pcone)$.
             \label{thm:rad_to_shad_nz}
        \end{enumerate}
\end{theorem}

\begin{corollary}\label{cor:rad_to_shad}
   Let $x \in \inter_{r \in R}(\pcone + r)$ then $R+\alpha x+\pcone\in\prop(\pcone)$,
   where $\alpha>1$.
\end{corollary}
\begin{proof}
    From the definition of the dual cone, this means
    \begin{gather}
        R + x\subseq\pcone \iff \forall{x^*\in\fcone},\ \forall{r\in R},\
	\inner{r + x; x^*} ≥ 0. \label{eq:dual_set_inclusion}
    \end{gather}
    Minimising the inner product in \eqref{eq:dual_set_inclusion} we have
    \begin{align}
        \forall{x\in\pcone},\ \forall{r\in R},\  \inner{r + x; x^*} &≥ \inf_{z\in\fcone\setminus{0}}\min_{r\in R}\inner{r + x;z}\\
        &=
	\inf_{z\in\fcone\setminus{0}}\left(-\inner*{\varfrac{z}{\gauge_R(z)};z}
	+ \inner{x; z}\right)\\
        &= \inf_{z\in\fcone\setminus{0}}\inner{x-\varfrac{z}{\gauge_R(z)};z}.
        \label{eq:rad_to_shad1}
    \end{align}
    We exclude $0$ from $\fcone$ since $z=0$ satisfies \eqref{eq:dual_set_inclusion} trivially. And $\min_{r\in R}\inner{r; z} = -\inner{\varfrac{z}{\gauge_R(z)};z}$ follows because since $R$ is symmetric and convex, the maximum occurs at $\inner{\varfrac{z}{\gauge_R(z)};z}$.
    
    Let us now choose $x\in\pcone$ as described in the theorem. Then
    \begin{gather}
       \begin{alignat}{3}
           \smash{x\in\inter_{r\in R\cap\fcone}(\pcone + r)}
           &\iff&\forall{r\in R\cap\fcone},\ &&
           x - r &\in \pcone\\
           &\:\implies&\ \ \ \ \ \ \forall{r\in \bd(R)\cap\fcone},\ &&
           x - r &\in \pcone\\
           &\iff&\forall{z\in\fcone\setminus{0}},\ &&
           \ \ x -\varfrac{z}{\gauge_R(z)} &\in \pcone
       \end{alignat}
        \shortintertext{which gives} 
        \inf_{z\in\fcone\setminus{0}}\inner{x-\varfrac{z}{\gauge_R(z)};z}≥0.\label{eq:rad_to_shad2}
    \end{gather}
    Therefore
    \begin{gather}
        \forall{x\in\fcone},\ \forall{r\in R},\  \inner{r + x; x^*} 
        \overset{\eqref{eq:rad_to_shad1}}{≥} 
        \inf_{z\in\fcone\setminus{0}}\inner{x-\varfrac{z}{\gauge_R(z)};z}
        \overset{\eqref{eq:rad_to_shad2}}{≥} 0.\label{eq:rad_to_shad3}
    \end{gather}
    Thus by \eqref{eq:dual_set_inclusion}, $R + x\subseq\pcone$. A closed
    cone is its own recession cone (Proposition
    \ref{prop:rccalc}(\ref{prop:rccalc_cone})), which proves 
    claim \ref{thm:rad_to_shad_subseq}.
   
   The only condition for $R+x+\pcone\in\prop(\pcone)$ that is
   non-trivial to show is $R+x+\pcone\ni 0$. As before assume
   $x\in\pcone$ is chosen according to the conditions of the theorem.
   Then $R\ni x$. Since $x\in\pcone$, let $x_0\coloneqq
   x·\oneon{\gauge_R(x)}$ and $x_1\coloneqq x·(1-\oneon{\gauge_R(x)})$ 
   and $x_0,x_1\in\pcone$. Then $x=x_0 + x_1$ and $x_0\in \bd(R)$. 
   Since $R$ is radiant $x_0≠ 0$.
   \begin{alignat}{2}
       x_0\in \bd(R) 
       &\iff& 0     &\in \bd(R) - x_0.
   \end{alignat}
   By the symmetry of $R$, this is eqivalent to
   \begin{alignat}{2}
       0 \in \bd(R) + x_0
       &\iff& x_1 &\in \bd(R) + x_0 + x_1\\
       &\iff& 0     &\notin \bd(R) + x,
   \end{alignat}
   which implies $R+x+\pcone\ni 0$, and claim \ref{thm:rad_to_shad_nz} is proved.
\end{proof}

In light of Theorem \ref{thm:rad_to_shad} it might be surprising to note that
there is no obvious operation
$\shdy(\R^n)\to\rdnt(\R^n)$. The long flat portions of  the set
$S_2\in\shdy(\R^2)$  in Figure \ref{fig:rad_to_shad} make it easy to see how
we might reconstruct $B_2$ by translating $S_2$ and forcing the resulting
set to be symmetric. But there is no such simple answer for the
superprediction sets of unbounded losses. See, for example,
$\super(\logloss)$ (Figure \ref{fig:loss_set_duality})  and $\super(\cdloss)$
(Figure \ref{fig:cd_loss}).

\subsection{The Norm Losses}\label{ssub:the_norm_losses}
Let $(\ball_\alpha)_{\alpha\in[1,\infty]}$ be the family of closed unit
$\alpha$-norm balls in $\R^n$ with $\ball_\alpha \coloneqq  \set{x\in\R^n;
	\norm{x}_\alpha ≤ 1}$. The family $(\ball_\alpha)$ is increasing in the
	sense that
\begin{gather}
   \alpha ≤ \gamma \iff \ball_\alpha \subseq \ball_\gamma.
\end{gather}
The point $1_n$ satisfies $1_n \in \inter_{r \in \ball_\alpha}(\fcone + r)$
for all $\alpha\in[1,\infty]$. For each $\ball_\alpha$ take the point
$\g\big(1+\oneon{\gauge_{\ball_\alpha}(1_n)})1_n= (1 +
n^{-\oneon\alpha})1_n$ and build the set $S_\alpha\coloneqq  \ball_\alpha +  (1 +
n^{-\oneon\alpha})1_n + \Rp^n$. By Corollary \ref{cor:rad_to_shad} we have
$S_\alpha\in\prop(\R_{\ge 0}^n)$, guaranteeing properness of the associated loss
functions: $\normloss \in \subdiff \cvsprt_{S_\alpha}$. The set $\ball_2$
along with $S_2$ is shown in Figure \ref{fig:rad_to_shad}.

Our choice of construction of  $S_\alpha$ has another convenient property:
\begin{gather}
    \forall{\alpha\in[1,\infty]},\  
    \cvsprt_{S_\alpha}(\normloss^\apolar(1_n)) =1. \label{eq:unif_normalisation}
\end{gather}
That is, the family $(\normloss)_{\alpha\in[1,\infty]}$ has the
normalisation about the conditional Bayes risk from
\S\ref{sub:canonical_normalisation}.

\begin{figure}
    \centering
    \subcaptionbox{The family $(\cvsprt_{S_\alpha})_{\alpha\in[1,\infty]}$, coloured from blue to red.\label{fig:building_shady_from_radient_cond_bayes_risks}}
    {\tikzsetnextfilename{norm_losses}\hspace*{-4ex}\begin{tikzpicture}
            \begin{axis}[domain=0:1, xmin=0, xmax=1, ymin=0.5, ymax=1.02, width=0.5\textwidth, height=0.5\textwidth, xlabel={$p_1$}, ylabel={$\cvsprt_{S_\alpha}(p_1,1-p_1)$},
            xtick={0,0.25,0.5,0.75,1},
            ytick=,
            extra x ticks={0},
            extra y ticks={0.5},
            inner axis line style={-},
            xtick style={draw},
            ytick style={draw},
            xlabel style={anchor=west},
            ylabel style={anchor=south},
            ]
                \pgfplotsinvokeforeach{1,...,40}{
                    \pgfmathsetmacro\q{1.25^(#1)}
                    \pgfmathtruncatemacro{\pcnt}{100*1.04^(-#1)}
                    \pgfmathsetmacro\p{\q/(\q-1)}
                    \edef\temp{\noexpand%
                        \addplot[color=Blue!\pcnt!Red]{ 1+2^(-1/\q) - ( x^(\p) + (1-x)^(\p) )^((\q-1)/\q)};
                    }\temp
                }
                \addplot[color=Blue, domain=0:0.5] {x + 0.5};\addplot[color=Blue, domain=0.5:1] {-x + 1.5};
                \addplot[color=Red]{1};
            \end{axis}
        \end{tikzpicture}}
    \hspace{-4ex}
    \subcaptionbox{The loss functions $\normloss_1$, $\normloss_2$ and $\normloss_\infty$ acting on the vector $p=(\varfrac{7}{10},\varfrac{3}{10})\in\probm$.\label{fig:building_shady_from_radient}}{%
    \tikzsetnextfilename{building_shady_from_radient}
    \colorlet{Mixed}{Blue!50!Red}
    \hspace*{-1ex}
    \begin{tikzpicture}
            \begin{axis}[domain=0.05:0.95, xmin=0, xmax=2, ymin=0, ymax=2, width=0.5\textwidth, height=0.5\textwidth, xlabel={$p_1, \ell(p,{\rv y}_1)$}, ylabel={$p_2, \ell(p;{\rv y}_2)$},
            xtick={0,0.5,1},
            ytick={0,1.5,1}]
                \coordinate (o) at (axis cs: 0,0);
                \path[name path=upper_axis] (axis cs:0,5) -- (axis cs:5,0);
                \path[name path=upper_axis2] (axis cs:0.707107,5) -- (axis cs:5,0.707107);
                
                \coordinate (p) at (axis cs:0.7,0.3);
                \draw[faint] (o) -- ($5*(p)$);
                \draw[shorten >=0cm, shorten >=1.5pt, -latex, name path=lp_ray] (o) -- (p);
                
                
                \draw[set, color=Blue, name path=minl] 
                    (axis cs:0.5,3) -- (axis cs:0.5,1.5) -- (axis cs:1.5,0.5) -- (axis cs:3,0.5);
                \addplot[fill=LightBlue] fill between[of=minl and upper_axis];
                \draw[set, color=Blue] (2,0.5) node[anchor=south east] {$S_1$};
                
                \addplot[set, color=Mixed, name path=eucl, domain=pi:1.5*pi,samples=300]({cos(deg(x)) + 1.707107},{sin(deg(x)) + 1.707107});
                \path[draw, set, color=Mixed] (axis cs:0.707107,3)-- (axis cs:0.707107,1.707107);
                \path[draw, set, color=Mixed] (axis cs:1.707107,0.707107) -- (axis cs:3,0.707107);
                \addplot[fill=Mixed!20] fill between[of=eucl and upper_axis2];
                \draw[set, color=Mixed] (axis cs:2,0.707107) node[anchor=south east, inner sep=1pt, rounded corners=3pt] {$S_2$};
                
                \draw[set, color=Red, name path=cst] (axis cs:1,3) -- (axis cs:1,1) -- (axis cs:3,1);
                \draw[set, color=Red] (axis cs:2,1) node[anchor=south east] {$S_\infty$};  
                \addplot[fill=LightRed] fill between[of=cst and upper_axis];    
        
                \coordinate (minlp) at (axis cs:0.5,1.5);
                \draw[shorten >=-3cm, shorten <=-5cm,faint] (minlp) -- ($(o)!(minlp)!($4*(p)$)$);
                \draw[right angle quadrant=1, right angle symbol={o}{p}{minlp},faint];

                \coordinate (euclp) at (axis cs:0.7872,1.31319);
                \draw[shorten >=-3cm, shorten <=-5cm,faint] (euclp) -- ($(o)!(euclp)!($4*(p)$)$);
                \draw[right angle quadrant=1, right angle symbol={o}{p}{euclp},faint];
                
                \coordinate (cstlp) at (axis cs:1,1);
                \draw[shorten >=-3cm, shorten <=-5cm,faint] (cstlp) -- ($(o)!(cstlp)!($4*(p)$)$);
                \draw[right angle quadrant=1, right angle symbol={o}{p}{cstlp},faint];   
                
                \addplot[set, domain=0:1] ({x},{1-x}) node[pos=0.95, anchor=south west, xshift=0.5em, prominent] {$\probm$};
                
                \draw[dashed, faint] ($(o)!(cstlp)!(axis cs:0,5)$) -- (cstlp) -- ($(o)!(cstlp)!(axis cs:5,0)$);
                \draw[dashed, faint] ($(o)!(minlp)!(axis cs:0,5)$) -- (minlp) -- ($(o)!(minlp)!(axis cs:5,0)$);
                
                \draw (minlp) node[dot] {} node[anchor=north east] {$\normloss_1(p)$};
                \draw (euclp) node[dot] {} node[anchor=north east,xshift=-1ex,yshift=-1ex,prominent] {$\normloss_2(p)$};
                \draw (cstlp) node[dot] {} node[anchor=south west] {$\normloss_\infty(p)$};
                \draw (p) node[dot] {} node[anchor=south, yshift=1ex] {$p$};
            \end{axis}
            \draw (o) node[anchor = north east] {$0$};
        \end{tikzpicture}
    }
        \caption{The family $\g(\normloss)_{\alpha\in[1,\infty]}$ smoothly varies between $\ooneloss + \half·1_n$ and the constant loss function.}
\end{figure}
We can derive the closed form expression for the whole family on $\R_{\ge
0}^n$ as follows:
\begin{align}
    \cvsprt_{S_\alpha} 
    = \cvsprt_{\ball_\alpha} + (1 + n^{-\oneon\alpha})\inner{1_n; \marg} + \cvsprt_{\,\Rp^n}
    = \cvsprt_{\ball_\alpha} + (1  + n^{-\oneon\alpha})\norm1{} + \cvsprt_{\,\Rp^n}.
\end{align}
Using \eqref{eq:cvx_to_ccv_sprt} and the fact that $\ball_\alpha$ is symmetric we have
\begin{gather}
    \cvsprt_{\ball_\alpha} 
    = - \cxsprt_{-\ball_\alpha} 
    = - \cxsprt_{\ball_\alpha}
    = - \agauge_{\ball_{\gamma}} 
    = -\norm{}_{\gamma},
\end{gather}
where $\gamma$ is the Hölder conjugate of $\alpha$, that is $\oneon{\alpha}
+ \oneon{\gamma}= 1$. Therefore
\begin{gather}
    \cvsprt_{S_\alpha} = (1 + n^{-\oneon{\alpha}})\norm1{} -
    \norm{}_{\frac{\alpha}{\alpha - 1}} + \cvsprt_{\R_{\ge 0}^n}.
\end{gather}
The family $(\cvsprt_{S_\alpha})_{\alpha\in[1,\infty]}$ is plotted in
Figure \ref{fig:building_shady_from_radient_cond_bayes_risks}. We can compute
the  subdifferential of $\cvsprt_{S_\alpha}$ directly:
\begin{align}
   \forall{p\in\pcone\setminus{0}},\  \subdiff\cvsprt_{S_\alpha}(p) 
    =  (1 + n^{-\oneon\alpha})\subdiff\norm1{}(p) - \frac{p}{\norm{p}_{\frac{\alpha}{\alpha - 1}}},
\end{align} 
giving us a closed form expression for $\normloss$:
\begin{gather}
    \forall{p\in\pcone\setminus{0},\ \forall\rv y \in \cal Y},\  \normloss(p,\rv y)
    \coloneqq 1 + n^{-\oneon\alpha} - \frac{p_{\rv y}}{\norm{p}_{\frac{\alpha}{\alpha - 1}}}.
\end{gather}

Some special values of $\alpha$ include:
\begin{gather}
	\forall{p\in\pcone\setminus{0}},\ \forall \rv y \in \cal Y,\ 
	\normloss_1(p,\rv y) 
    = \ooneloss(p,\rv y) + \half \mbox{\ and\ }  \normloss_\infty(p,\rv y) = 
    \ell_{-\infty}(p;\rv y) = 1,
\end{gather}
where $\ooneloss$ is misclassification loss \eqref{eq:oone_loss_defn} and
$\ell_{-\infty}$ is the \emph{constant loss} (which we derived in a
completely different way in \S\ref{ssec:concave_norm_losses}). The
various intermediaries like $\normloss_2$ smoothly interpolate between
these two extremes as illustrated in
Figure \ref{fig:building_shady_from_radient}, where it is clear that the
condition \eqref{eq:unif_normalisation} is equivalent to the simpler geometric
property $1_n\in\inter_{\alpha\in[1,\infty]}\bd(S_\alpha)$. Finally we note
the family $\g\big(\normloss)_{\alpha\in[1,\infty]}$ is clearly bounded.

\begin{example}
	\normalfont
    We can now derive the antipolar result $\ooneloss^\apolar =
    \ell_{-\infty}=\normloss_\infty$ from
    \S\ref{ssec:concave_norm_losses} using a simpler geometrical
    argument: Consider the set $\super(\ell_{-\infty})$ which has the
    property that $\cvsprt_{\super(\ell_{-\infty})} = \norm1{}$ over
    $\Rp^n$. And so we have
        \begin{gather}
            \super(\ell_{-\infty})^\apolar = \lev_{≥1}\cvsprt_{\super(\ell_{-\infty})} 
            = \lev_{≥1}(\norm1{} +\cvsprt_{\Rp^n})
            = \g(\lev_{≥1}\norm1{})\cap\Rp^n.
        \end{gather}
    These sets are illustrated in Figure \ref{fig:polars_between_normlosses}.  
\end{example}
\begin{figure}
    \centering
    \tikzsetnextfilename{polar_superprediction_sets}
    \begin{tikzpicture}
        \begin{axis}[domain=0.05:0.95, xmin=0, xmax=2, ymin=0, ymax=2, width=0.5\textwidth, height=0.5\textwidth, xlabel={${p_1, \ell(p;\rv y_1)}$}, ylabel={$p_2, \ell(p;{\rv y}_2)$},
        xtick={0,1},
        ytick={0,1}]
            \coordinate (o) at (axis cs: 0,0);
            \path[name path=upper_axis] (axis cs:0,5) -- (axis cs:5,0);
            \path[name path=upper_axis2] (axis cs:0.707107,5) -- (axis cs:5,0.707107);
            
            \coordinate (p) at (axis cs:0.7,0.3);
            \draw[faint] (o) -- ($5*(p)$);
            \draw[shorten >=0cm, shorten >=1.5pt, -latex, name path=lp_ray] (o) -- (p);
            
            \draw[set, color=Magenta, name path=cst] (axis cs:0,3) -- (axis cs:0,1) -- (axis cs:1,0) -- (axis cs:3,0);
            \draw[set, color=Magenta] (axis cs:2,0) node[anchor=south east] {$\super\big(\normloss_\infty^\apolar)$};  
            \addplot[fill=LightMagenta] fill between[of=cst and upper_axis];   
            
            \draw[set, color=Red, name path=cst] (axis cs:1,3) -- (axis cs:1,1) -- (axis cs:3,1);
            \draw[set, color=Red] (axis cs:2,1) node[anchor=south east] {$\super\big(\normloss_\infty)$};  
            \addplot[fill=LightRed] fill between[of=cst and upper_axis];

            \coordinate (cstlp) at (axis cs:1,1);
            \draw[shorten >=-3cm, shorten <=-5cm,faint] (cstlp) -- ($(o)!(cstlp)!($4*(p)$)$);
            \draw[right angle quadrant=1, right angle symbol={o}{p}{cstlp},faint];   
            
            \coordinate (cstlp_polar) at (axis cs:0,1);
            \draw[shorten >=-3cm, shorten <=-5cm,faint] (cstlp_polar) -- ($(o)!(cstlp_polar)!($4*(p)$)$);
            \draw[right angle quadrant=1, right angle symbol={o}{p}{cstlp_polar},faint];   
            
            \addplot[set, domain=0:1] ({x},{1-x}) node[pos=0.95, xshift=0.5em, anchor=south west] {$\probm$};
            
            \draw[dashed, faint] ($(o)!(cstlp)!(axis cs:0,5)$) -- (cstlp) -- ($(o)!(cstlp)!(axis cs:5,0)$);
            
            \draw (cstlp) node[dot] {} node[anchor=south west] {$\normloss_\infty(p)$};
        \end{axis}
        \draw (o) node[anchor = north east] {$0$};
            \draw (cstlp_polar) node[dot] {} node[anchor=south west] {$\normloss_\infty^\apolar(p)$};
            \draw (p) node[dot] {} node[anchor=south, yshift=1ex] {$p$};
    \end{tikzpicture}
    \caption{Illustration of $\co\super(\ooneloss) = \super(\normloss_\infty^\apolar)$ and that $\inner*{\normloss_1^\apolar(p);p}=\min_{\rv y \in \cal Y} p_{\rv y}$.}
    \label{fig:polars_between_normlosses}
\end{figure}


\section{Combining Given Proper Losses to Form New Ones} 
\label{sec:new_losses_from_old}

Much machine learning practice works with a small family of loss functions
for the pragmatic reason that they are available, and have explicit
formulas. The above development shows there is an enormous range of
possible proper losses one could use, but offers no concrete way of
constructing them (with explicit formulas). In this section, we develop a
straightfroward way of constructing a larger usable set of loss functions
by finding ways of combining existing proper losses in a manner that
guarantees the result is also a proper loss, and which provides explicit
formulas for the resulting loss function and associated conditional Bayes
risk (concave support function).

In \S\ref{sec:designing_losses_via_their_superprediction_sets} we observed
the power of defining loss functions $\ell$ by directly building their
superprediction sets $\super(\ell)$. We also saw that the family
$\prop(\pcone)$ is a cone, and is therefore closed under the family of
operations
\begin{gather}
	\forall{T\subseq\pcone},\ \forall\alpha>0,\ \prop(\pcone) \ni S \mapsto \alpha S + T \in \prop(\pcone).\label{eq:prop_cone_closure}
\end{gather}
It's natural then to consider what other operations have a closure
property analogous to \eqref{eq:prop_cone_closure} for the family $\prop(\pcone)$. With
the relationship between proper losses and $S\in\prop(\pcone)$, this
amounts to asking whether one can combine multiple proper losses
non-additively and still be guaranteed that the result is a proper
loss; when working directly with $\ell$, it is not obvious how to ensure
the resulting loss is proper\footnote{This question is 
	obviously analogous to the question of
	``aggregation'' in economics; see for example 
	\citep[Chapter 6]{Shephard:1970aa}. In our case, restricting
	consideration to proper losses makes the problem situation simpler,
	and a rather more comprehensive answer can be given.
}.
The convex analysis literature has largely studied the closely related family
$\rdnt(\bnch)$ \citep{Seeger1990,Seeger1995,Gardner:2013aa,Gardner:2018aa},
with some results for the family $\shdy(\bnch)$
\citep{Barbara1994,Penot1997Duality,Penot2000}. 

We will present a general family of operations, called $M$-sums and dual
$M$-sums, \citep{Gardner:2013aa,Mesikepp:2016tf} which provide a 
general means by which to create new proper
losses from two or more given proper losses. These $M$-sums provide the
opportunity to smoothly interpolate between several proper losses in a
variety of ways (beyond merely taking the sum)\footnote{
	In applying the existing theory of $M$-sums we have needed to
	extend it in two ways: we have developed the concave version (which
	combines shady sets rather than radiant ones), and we have
	developed a comprehensive duality theory. These results may be of
	interest in their own right. They extend and generalise a range of
	results in the literature on the combination of convex bodies, including 
	\citep{Artstein-Avidan:2017aa, Penot2000,
	Seeger1995, Volle:1998aa, Luc:1997aa, Penot:2001aa,
	Barbara1994,Pallaschke:2013aa,Milman:2017aa, Slomka:2011aa, Milman:2017ab}.
}. 

In \S\ref{sec:m-sum_results} we introduce our approach, and then
successively demonstrate the preservation of the convexity of the
superprediction sets (\S\ref{sec:M-convexity}), their 
closure (\S\ref{sub:M-closure}), and orientation (\S\ref{sec:M-orientation}).
We then introduce the functional analog of our
combination rules (which serve to combine conditional Bayes risks)
(\S\ref{sec:the_functional_m-sum}), examine their properties in terms of
support functions (\S\ref{sub:support_functions}), and the effect of polar
operations (\S\ref{sub:the_m-sum_polar}). The argument is summarised and
tied together in \S\ref{sub:common_texorpdfstring}.


The \Def{epimultiplication} operation is 
\begin{gather}
  \R\times 2^{\bnch} \ni (\alpha,S) \mapsto \Def{\alpha \star S}{\begin{cases}
      \alpha S & \alpha ≠ 0,\\
      \rec(S)  & \alpha = 0.
  \end{cases}
  }
\end{gather}
Fix $M\subseq \R^m$. Then the \Def{$M$-sum} and \Def{dual $M$-sum} operations are
defined as
\begin{gather}
    \g\big(2^{\bnch})^m\ni (A_1,\dots,A_m)\mapsto \Def{\msum(A_1,\dots,A_m)}{\union_{\mu\in M}\sum_{i\in[m]} \mu_i \star A_i},
    \shortintertext{and}
    \g\big(2^{\bnch})^m\ni (A_1,\dots,A_m)\mapsto \Def{\dualmsum(A_1,\dots,A_m)}{\union_{\mu\in M}\inter_{i\in[m]} \mu_i \star A_i}.
\end{gather}

\subsection{\texorpdfstring{$M$}{M}-Composition of Losses} 
\label{ssec:m_composition_of_losses}

Throughout this section let $\ell_1,\dots,\ell_m$ be a sequence of proper
loss functions, paired with their conditional Bayes risk functions
$\minL_1,\dots,\minL_m\colon\pcone\to\Rx$. Let
$\emm\colon\Rpp^m\to\Rp^m$ be a proper loss function with the associated
conditional Bayes risk $\minM\colon\Rp^m\to\Rx$. We introduce the two functions
\begin{gather}
    \Def{\msum_{\minM}(\minL_1,\dots,\minL_m )}{p\mapsto\minM(\minL_1(p),\dots,\minL_m(p))}\label{eq:msum_loss}
    \shortintertext{and}
   \Def{\dualmsum_{\minM}(\minL_1,\dots,\minL_m )}{p\mapsto\sup\set{\minM(\minL_1(a_{1}),\dots, \minL_m(a_{m})); a_{1} + \dots + a_{m} = p}},\label{eq:dual_msum_loss}
\end{gather}
which we call the \Def{functional $M$-sum} and \Def{dual functional $M$-sum} respectively.

The functional $M$-sum encompasses a wide range of operations on losses.
For example using Corollary \ref{cor:m-sum_subdifferential} we can see the sum of
losses $\jay$ and $\kay$ can be written as an $M$-sum using the constant
loss $\ell_{-\infty}$ with $\minN\coloneqq  \cvsprt_{\super(\ell_{-\infty})}$:
\begin{gather}
    \subdiff \msum_{\minN}(\cvsprt_{\super(\jay)},\cvsprt_{\super(\kay)})\ni \jay + \kay.
\end{gather}

Note that $\jay$ defined in this way is not guaranteed to retain the
pseudo-inverse property of the antipolar $\ell^\apolar$ in the sense of
Proposition \ref{prop:the_polar_loss}.

\subsection{Constructing Superprediction Sets with \texorpdfstring{$M$}{M}-Sums} 
\label{sec:m-sum_results}

Our approach here is organised as follows: First
we show some general sufficient
conditions for the operations $\msum$ and $\dualmsum$ (introduced in
\S\ref{sec:new_losses_from_old}) to map $\prop(\pcone)$ to $\prop(\pcone)$.
Since we are ultimately interested in the support functions of these sets,
in \S\ref{sec:the_functional_m-sum} we compute the (convex and concave)
support functions of sets in the images of $\msum$ and $\dualmsum$. The
choices of name and notation for the dual $M$-sum are not accidental, in
\S\ref{sub:the_m-sum_polar} we characterise the duality relationship of
$\msum$ and $\dualmsum$ in terms of the polar and antipolar and present
closure results for the families $\rdnt(\bnch)$ and $\shdy(\pcone)$
(Theorem \ref{thm:radshad_to_radshad} and Theorem \ref{thm:dual_msum}),
which allows us to compute the gauge and antigauge functions of sets in the
images of $\msum$ and $\dualmsum$ (Corollary \ref{cor:gauge_msum}). 

\label{sub:m-sums_over_the_family_P(X_+)}
We first seek to establish closure (in the algebraic sense) of
the family $\prop(\pcone)$ with the operations $\msum$ and $\dualmsum$. In
order to show this requires a number of theorems, which culminate in the
dénouement Corollary \ref{cor:msum_finale}. For ease of exposition these
results are summarised in Table \ref{tab:msum_summary}.

It is necessary to introduce the \Def{Panlevé--Kuratowski} 
 notion of convergence for sequences of sets \citep{RockafellarWets2004}. 
 Define the following classes of subsets:
\begin{gather}
	\Def{\cal N}{\set{N\subseq \N ; \text{$\N\setminus N$ is finite}}}
  \mbox{\ and\ } 
  \Def{\cal N^{\#}}{2^{\N}}.
\end{gather}
Let $(S_n)_{n\in\N}$ with $S_n\subseq\bnch$ be a sequence of sets. Then the 
\Def{inner and outer limit} are
\begin{align}
	\Def{\liminf_{n\to\infty} S_n} &\coloneqq  \set{x \in \bnch ; \forall{N\in\cal{N}},\
  \forall  n\in N,\ \exists{x_n\in S_n},\  x_n \to x  }\\
  \shortintertext{and}
  \Def{\limsup_{n\to\infty} S_n} &\coloneqq  \set{x \in \bnch ;
	  \forall{N\in\cal{N}^{\#}},\ 
  \forall n\in N,\ \exists{x_n\in S_n},\  x_n \to x  }.
\end{align}
If $\liminf_{k\to\infty} C_k = \limsup_{k\to\infty} C_k$ then we say
$(C_k)_{k\in\N}$ converges with limit $\lim_{k\to\infty} C_k$. As one might
hope, since $\cal N \subseq \cal N^{\#}$ it follows that
$\liminf_{n\to\infty} S_n \subseq \limsup_{n\to\infty} S_n$.

\begin{proposition}\label{prop:epim_continuity}
    Let $S\in\cvx(\bnch)$, $(\mu_k)_{k\in\N} \to \mu$ with $\mu_k\in\Rpp$ for all $k\in\N$. Then $\mu\star S = \lim_{k \to \infty}\mu_k·S$.
\end{proposition}
\begin{proof}
    The only interesting case is when $\mu = 0$, which is immediate from
    Lemma \ref{prop:rccalc}\ref{prop:rccalc_seq}.
\end{proof}

\subsection{Convexity}
\label{sec:M-convexity}

The convexity of superprediction sets plays an essential role in our
theory, and thus if we wish to combine multiple proper losses by combining
their superprediction sets, we need to ensure the resulting set is
guaranteed convex. We first need some auxilliary lemmas. 
\begin{lemma}\label{lem:epim_rec_cone}
    Let $S\in\cvx(\bnch)$, and $\alpha,\beta \in [0,\infty)$.  Then
    \begin{enumerate}
        \item $\rec(\alpha\star S)=\rec(S)$; \label{lem:epim_rec_cone_eq}
        \item $\alpha \star S + \beta \star S = (\alpha + \beta)\star S$. \label{lem:epim_rec_cone_dist}
    \end{enumerate}
\end{lemma}
\begin{proof}  
    \ref{lem:epim_rec_cone_eq}. Let $\alpha = 0$. Then $\alpha\star S =
    \rec(S)$. Since $\rec(S)$ is a closed cone, it is easily verified
    (Proposition \ref{prop:rccalc}\ref{prop:rccalc_cone}) that
    $\rec(\rec(S))=\rec(S)$. For $\alpha>0$ we have $\rec(\alpha\star S) =
    \rec(S)$. This is an immediate consequence of Proposition
    \ref{prop:rccalc}(\ref{prop:rccalc_seq}). 
    
    Turning now to claim \ref{lem:epim_rec_cone_dist} there are three cases:
    neither $\alpha$ nor $\beta$ is zero,
    only one of $\alpha$ or $\beta$ is zero, or 
    both $\alpha$ and $\beta$ are zero.
    The only interesting case is the second. Let $\alpha≠0, \beta=0$ and we have
    \begin{align}
        \alpha\star S + \beta\star S &= \alpha\star S +  \rec(S)
        = \alpha \star S
        = (\alpha + \beta) \star S.
    \end{align}
   The second equality follows from Proposition
   \ref{prop:rccalc}(\ref{prop:rccalc_cvx}) since $\rec(S) = \rec(\alpha \star S)$. 
\end{proof}
\begin{lemma}\label{lem:intersect_minkowski}
    Let $I$ be an arbitrary index set. For families of subsets of $\bnch$, $(S_i)_{i\in I}$ and $(T_j)_{j\in I}$ we have $\inter_{i\in I}S_i + \inter_{j\in I}T_j \subseq \inter_{i\in I}(S_i + T_i)$.
\end{lemma}
\begin{proof}
    Let $x\in \inter_{i\in I}S_i + \inter_{j\in I}T_j$. Then $x = s + r$ for some points $s,r$ where $s$ is in every $S_i$, and $r$ is in every $T_j$. Thus $x \in S_i + T_j$ for all $i,j\in I$, including the pairs $(S_i,T_j)$ with $j = i$. 
    Consequently $x$ is in the intersection $\inter_{i\in I}(S_i + T_i)$.
\end{proof}
The main result of this subsection is the following. 
\begin{theorem}\label{thm:msum_cvx_to_cvx}
    Let $M\in\cvx(\R^m)$, and $A_i\in\cvx(\bnch)$ for  $i\in[m]$. 
    Then the sets $\msum(A_1,\dots,A_m)$ and $\dualmsum(A_1,\dots,A_m)$ are convex.
\end{theorem}
\begin{proof}
    Fix arbitrary $x,y\in\msum(A_1,\dots,A_m)$. Then there are $\mu,\nu\in M$, such that  
    \begin{gather}
	    x \in \sum_{i\in[m]} \mu_i\star A_i \mbox{\ and\ } y \in \sum_{j=1}^m\nu_j\star A_j.\label{eq:m_sum_cvx_constr}
    \end{gather}
    To show $\msum(A_1,\dots,A_m)$ is a convex set, we need to show $tx + (1-t)y \in \msum(A_1,\dots,A_m)$ for all $t\in(0,1)$. By virtue of \eqref{eq:m_sum_cvx_constr},
	    $\forall{t\in(0,1)},$
    \begin{align}
        t x +(1-t) y  &\in t \sum_{i\in[m]} \mu_i\star A_i + (1-t)\sum_{j=1}^m\nu_j\star A_j\\
        &=\sum_{i\in[m]}\left(t\mu_i\star A_i + (1-t)\nu_i\star A_i\right).
	    \label{eq:m_sum_cvx2}
    \end{align}
    Applying Lemma \ref{lem:epim_rec_cone} with $S=A_i$, $\alpha=t\mu_i$ and $\beta=(1-t)\nu_i$ implies
    \begin{gather}
        \forall{i\in[m]},\ 
        t\mu_i\star A_i + (1-t)\nu_i\star A_i = (t\mu_i + (1-t)\nu_i)\star A_i,
        \label{eq:m_sum_cvx2a}
        \shortintertext{and thus}
        \sum_{i\in[m]}\left(t\mu_i\star A_i + (1-t)\nu_i\star A_i\right) =
	  \sum_{i\in[m]}(t\mu_i + (1-t)\nu_i)\star A_i. \label{eq:m_sum_cvx3}
    \end{gather}
    Finally, convexity of $M$ guarantees $t\mu + (1-t)\nu \in M$, and therefore
        $\forall{t\in(0,1)},$
    \begin{align}
        tx + (1-t)y 
        &\overset{\eqref{eq:m_sum_cvx2}}{\in} 
          \sum_{i\in[m]}\left(t\mu_i\star A_i + (1-t)\nu_i\star A_i\right)\\
        &\overset{\eqref{eq:m_sum_cvx3}}{=}
          \sum_{i\in[m]}(t\mu_i + (1-t)\nu_i)\star A_i\\
        &\overset{\hphantom{\eqref{eq:m_sum_cvx3}}}{\subseq}
            \union_{\mu\in M}\sum_{i\in[m]} \mu_i \star A_i,
        \label{eq:m_sum_cvxfinal}
    \end{align}
    which concludes the proof that $\msum(A_1,\dots,A_m)$ is convex.
    
   The proof that $\dualmsum(A_1,\dots,A_m)$ is convex is  similar.
   Let $x,y\in \dualmsum(A_1,\dots,A_m)$. 
   Then there exists $\mu,\nu \in M$ such that $x
   \in\inter_{i\in[m]}\mu_i\star A_i$ and  $ y \in\inter_{j=1}^m\nu_j\star
   A_j$.  Therefore
        $\forall{t\in(0,1)}, $
    \begin{align}
        tx + (1-t) y 
        &\overset{\hphantom{\ref{lem:intersect_minkowski}}}{\in} t\g(
	\inter_{i\in[m]}\mu_i\star A_i) + (1-t)\g(\inter_{j\in[m]}\nu_j\star A_j)\\
        &\overset{\hphantom{\ref{lem:intersect_minkowski}}}{=}\g(\inter_{i\in[m]}t\mu_i\star
	A_i) + \g(\inter_{j\in[m]}(1-t)\nu_j\star A_j)\\
        &\overset{\ref{lem:intersect_minkowski}}{\subseq}\inter_{i\in[m]}\left(t\mu_i\star
	A_i + (1-t)\nu_i\star A_i\right).\label{eq:m_sum_cvx5}
    \end{align}
    From \eqref{eq:m_sum_cvx2a}, $\forall t\in(0,1)$,
    \begin{gather}
        \inter_{i\in[m]}(t\mu_i\star A_i + (1-t)\nu_i\star A_i)=\inter_{i\in[m]}(t\mu_i+(1-t)\nu_i)\star A_i.\label{eq:m_sum_cvx6}
    \end{gather}
    Again the convexity of $M$ guarantees that  $t\mu+(1-t)\nu \in M$,
    and mirroring \eqref{eq:m_sum_cvxfinal}, $\forall{t\in(0,1)}, $
    \begin{align}
        tx + (1-t) y
        &\overset{\eqref{eq:m_sum_cvx5}}{\in} 
            \inter_{i\in[m]}\left(t\mu_i\star A_i + (1-t)\nu_i\star
	    A_i\right)\\
        &\overset{\eqref{eq:m_sum_cvx6}}{=} 
            \inter_{i\in[m]}(t\mu_i+(1-t)\nu_i)\star A_i\\
        &\overset{\hphantom{\eqref{eq:m_sum_cvx6}}}{\subseq}
            \union_{\mu\in M}\inter_{i\in[m]}\mu_i\star A_i,
    \end{align}
    which concludes the proof that $\dualmsum(A_1,\dots,A_m)$ is convex.
\end{proof}
\begin{proposition}\label{prop:msum_subset_prop}
    Let $M\in\cvx(\Rp^m\setminus{0})$, and $A_i\in\cvx(\pcone\setminus{0})$ for $i\in[m]$. Then
    \begin{gather}
	    \msum(A_1,\dots,A_m)\subseq \pcone\setminus{0}\mbox{\ \  and\ \ } 
	    \dualmsum(A_1,\dots,A_m)\subseq \pcone\setminus{0}.
    \end{gather}
\end{proposition}
\begin{proof}
    Since $\pcone\setminus{0}$ is a cone, it is closed under addition and positive multiplication. The set $M$ does not contain $0\in\R^m$ and since the $A_i$ are all subsets of $\pcone\setminus{0}$, for all $\mu\in M$ the following inclusions are immediate:
        \begin{gather}
          \mu_1\star A_1 + \dots + \mu_m\star A_m\subseq \pcone\setminus{0}
	  \mbox{\ \ and\ \ }
          \mu_1\star A_1 \cap \dots \cap \mu_m\star A_m\subseq \pcone\setminus{0}.
        \end{gather}
\end{proof}
\vspace*{-6mm}

\subsection{Closure}
\label{sub:M-closure}
Superprediction sets are closed by construction, so we also need to ensure
that our combination rules preserve closure. First we need the following
lemma. 
\begin{lemma}\label{lem:convergence_in_sets2}
    Let $A_i \in \cvx(\pcone)$ for $i\in[m]$. Let  $(\mu^{k})_{k\in\N} \to \mu$ with $\mu^k\in\Rp^m$. Then
    \begin{enumerate}
        \item $\lim_{k\to\infty}(\mu_1^{k}A_1 + \dots +\mu_m^{k}A_m) = \mu_1\star A_1 + \dots +\mu_m\star A_m,$\label{lem:convergence_in_sets_sum}
        \item $\lim_{k\to\infty}(\mu_1^{k} A_1 \cap \dots \cap\mu_m^{k}A_m) = \mu_1\star A_1 \cap \dots \cap\mu_m\star A_m.$ \label{lem:convergence_in_sets_inter}
    \end{enumerate}
\end{lemma}
\begin{proof}
  Define the set $C \coloneqq  \mu_1A_1 + \dots +\mu_mA_m$ and the sequence
  of sets $C_k \coloneqq  \mu_1^{k}A_1 + \dots +\mu_m^{k}A_m$. Likewise the
  set $D \coloneqq  \mu_1A_1 \cap \dots \cap\mu_mA_m$ and the sequence
  of sets $D_k \coloneqq  \mu_1^{k}A_1 \cap \dots \cap\mu_m^{k}A_m$. 
    
  Take an arbitrary convergent sequence $(x_k) \to x$ such that $x_k \in
  C_k$. Then  $x_k = \sum_{i\in[m]} \mu_i^{k}a_i^{k}$ with $a_i^{k}\in A_i$
  for $i\in[m]$ and each $k\in\N$. All the sequences $\mu_i^ka_i^k$ have
  convergent subsequences (Lemma \ref{lem:sequences_in_cone}) so it is
  without loss of generality to assume that they are convergent (by passing
  to  subsequence if necessary), and it follows that
  $\lim_{k_i\to\infty}\mu_i^{k_i}a_i^{k_i}$ exists for each $i\in[m]$ and 
  \begin{gather}
      x = \lim_{k_1\to\infty}\mu_1^{k_1}a_1^{k_1} + \dots 
      + \lim_{k_m\to\infty}\mu_m^{k_m}a_m^{k_m}
      \overset{\mathrm{P}\ref{prop:epim_continuity}}{\implies}
      x \in \mu_1\star A_1 + \dots + \mu_m\star A_m,
  \end{gather}
  since $\mu^k\in\Rp^n$ for all $k\in\N$, and we have 
  proven \eqref{lem:convergence_in_sets_sum}.
  
  Again take an arbitrary sequence $(x_k) \to x$ such that $x_k \in D_k$.
  Then $x_k = \mu_i^{k}a_i^{k}$ for some sequences $\mu_i^{k}a_i^{k}\in
  \mu_j^{k}A_j$ for all $i,j\in[m]$ and all $n\in\N$. Applying
  Proposition \ref{prop:epim_continuity} completes the proof of
  claim \ref{lem:convergence_in_sets_inter}.
\end{proof}

Our main result for this subsection is:
\begin{theorem}\label{thm:msum_closed}
   Let $M\in\cvx(\Rp^m\setminus{0})$, $A_i\in\cvx(\pcone\setminus{0})$ for
   $i\in[m]$. Then  $\msum(A_1,\dots,A_m)$ and $\dualmsum(A_1,\dots,A_m)$
   are both closed. 
\end{theorem}
\begin{proof}
    Take an arbitrary sequence $(x_k) \to x$ such that $x_k \in
    \msum(A_1,\dots,A_m)$. Then there exists a sequence $(\mu^k)_{k\in\N}$
    with $\mu_k\in M$ so that $x_k \in \sum_{i\in[m]}\mu_i^k\star A_i$ for
    all $k\in\N$. Assume the sequence $(\mu^k)_{k\in\N}$ is bounded. Then
    without loss of generality we may assume it is convergent (by passing 
    to a subsequence if necessary) with limit $\mu$. 
    Since $M$ is closed, $\mu\in M$. It follows that
    \begin{align}
        x &\ \ \, =\lim_{k\to\infty} x_k \in 
	\lim_{k\to\infty}(\mu_1^k\star A_1 + \dots + \mu_m^k\star A_m) \\
	& \overset{\mathrm{L}\ref{lem:convergence_in_sets2}
	(\ref{lem:convergence_in_sets_sum})}{=} \mu_1\star A_1 + \dots + 
	\mu_m\star A_m \subseq \msum(A_1,\dots,A_m).
    \end{align}
    
    A proof by contradiction shows that the sequence $(\mu^k)_{k\in\N}$ is
    bounded. Assume $(\mu^k)_{k\in\N}$ is unbounded. Then we can write
    $\mu^k = \nu^k·\norm{\mu^k}$ where $\nu^k =
    \varfrac{\mu^k}{\norm{\mu^k}}$ for each $k\in\N$. Thus
    $(\nu^k)_{k\in\N}$ is bounded and we may assume it is convergent with
    limit $\nu$. Therefore
    \begin{gather}
        x_k \in \mu_1^k\star A_1 + \dots + \mu_m^k\star A_m
        \iff 
        \frac{x_k}{\norm{\mu_k}} \in \nu_1^k\star A_1 + \dots + \nu_m^k\star A_m\\
        \implies 
        \lim_{k\to\infty} \frac{x_k}{\norm{\mu_k}} \in \lim_{k\to\infty}\nu_1^k\star A_1 + \dots + \nu_m^k\star A_m
	\overset{\mathrm{L}\ref{lem:convergence_in_sets2}(\ref{lem:convergence_in_sets_sum})}{\iff} 
        0 \in \nu_1\star A_1 + \dots + \nu_m\star A_m,
    \end{gather}
    which contradicts Proposition \ref{prop:msum_subset_prop} (taking
    $M=\set{\nu}$). Thus $\msum(A_1,\dots,A_m)$ is
    closed.
    
    Again take an arbitrary sequence $(x_k) \to x$ such that $x_k \in
    \dualmsum(A_1,\dots,A_m)$. Then $x_k = \mu_i^{k}a_i^{k}$ for some
    sequences $\mu_i^{k}a_i^{k}\in \mu_j^{k}A_j$ for all $i,j\in[m]$ and
    all $n\in\N$. Clearly the sequences $(\mu_i^k)_{k\in\N}$ must be
    bounded for all $i\in[m]$. And so, as before, without loss of
    generality we assume it has limit $\mu\in M$.  Applying
    Proposition \ref{prop:epim_continuity} completes the proof that
    $\dualmsum(A_1,\dots,A_m)$ is closed.
\end{proof}

\begin{corollary}\label{por:msum_closed}
   Let $M\in\cvx(\R^m)$, $A_i\in\cvx(\R^m)$ for $i\in[m]$. Assume $M$ and
   each $A_i$ for $i\in[m]$ are compact. Then both $\msum(A_1,\dots,A_m)$
   and $\dualmsum(A_1,\dots,A_m)$ are closed. 
\end{corollary}
\begin{proof}
    The above result follows by an almost identical proof to
    Theorem \ref{thm:msum_closed}, however since $M$ and each $A_i$ for
    $i\in[m]$ are closed and bounded, this rules out the above pathologies
    when it comes to building bounded sequences $(a_i^k)_{k\in\N}$ with
    $a_i^k\in A_i$ for all $i\in[m]$.
\end{proof}

\subsection{Orientation}
\label{sec:M-orientation}

\begin{figure}[t]
    \centering
    \tikzsetnextfilename{p_addition_of_losses}
    \begin{tikzpicture}
        \begin{axis}[domain=0.01:0.99, xmin=0, xmax=3, ymin=0, ymax=3, width=0.8\textwidth, height=0.8\textwidth,
            xlabel={${p_1, \ell(p;\rv y_1)}$}, ylabel={$p_2, \ell(p;{\rv y}_2)$}]
            \coordinate (o) at (axis cs: 0,0);
            \path[name path=upper_axis] (axis cs:0,5) -- (axis cs:5,5) -- (axis cs:5,0);

            \addplot[set, color=Cyan, domain=0.1:0.9, name path=msum] ({
            0.515464*(0.742448-2.39536*x+4.17582*x^2-4.13537*x^3+1.61246*x^4+(0.231263-0.925053*x+1.38758*x^2-0.925053*x^3+0.231263*x^4)*ln(1-x)^2+x*(-0.828736+0.242016*x+2.00218*x^2-1.41546*x^3)*ln(x)+0.231263*(1-x)^2*x^2*ln(x)^2+ln(1-x)*(-0.828736+2.99435*x-4.01062*x^2+2.35315*x^3-0.508137*x^4-0.462526*x*(-1+x)^3*ln(x)))/(-0.215414-0.86792*x+x^2+(0.120225-0.120225*x)*ln(1-x)+0.120225*x*ln(x))^2},{0.515464*(x^2*(4.18906-7.80369*x+3.72574*x^2+0.231263*(1-x)^2*ln(1-x)^2+(-0.828736+0.508137*x)*x*ln(x)+0.231263*x^2*ln(x)^2+ln(1-x)*(-2.75233+5.18406*x-2.43173*x^2+(0.462526-0.462526*x)*x*ln(x))))/(-0.215414-0.86792*x+x^2+(0.120225-0.120225*x)*ln(1-x)+0.120225*x*ln(x))^2}) 
                node[pos=0.28, anchor=south, sloped] {$\msum_{\super(\bar\ell_{\half})}\big(\super\big(\expandafter\bar\logloss'), \super\big(\bar\ell_{\half}))$};
            \addplot[fill=LightCyan] fill between[of=msum and upper_axis];
            
            \addplot[set, dashed, color=Green, domain=0.01:0.99, name path=cvnorm] ({4/((1/(1-x)+1/x)^2*x^2)},{4/((1/(1-x)+1/x)^2*(1-x)^2)}) node[pos=0.21, anchor=south, sloped] {$\bd(\super\bar\ell_{\half})$};
            \addplot[set, dashed, color=Blue, domain=0.001:0.99, name path=shifted_log] ({(ln(2) - ln(x))/(ln(2) + ln(4))}, {(-ln(4/3) + ln(2) + ln(4) - ln(1 - x))/(ln(2) + ln(4))}) node[pos=0.21, anchor=south, sloped] {$\bd\big(\super\expandafter\bar\logloss')$};
            
            \draw[faint] (o) -- (axis cs: 5,5);
            \coordinate (un) at (axis cs: 1,1);
            \draw[dashed, faint] ($(o)!(un)!(axis cs:0,5)$) -- (un) -- ($(o)!(un)!(axis cs:5,0)$);
            \draw (un) node[dot] {};
        \end{axis}
        \draw (o) node[anchor = north east] {$0$};
    \end{tikzpicture}
    \caption{Illustration of Corollary \ref{cor:msum_finale}. We take the
    normalised log loss $\expandafter\bar\logloss$
    (\S\ref{sub:canonical_normalisation}) and shift the maximum of its
    conditional Bayes risk (see \S\ref{sec:shifting_the_max}) to
    $(\varfrac{3}{4},\varfrac{1}{4})$ to generate the loss function
    $\expandafter\bar\logloss'$. We then show the resulting renormalised
    $M$-sum of this loss with the normalised concave norm loss
    $\bar\ell_{\half}$, and
    $M\coloneqq \super(\bar\ell_{\half})$.}\label{fig:msum_finale}
\end{figure}
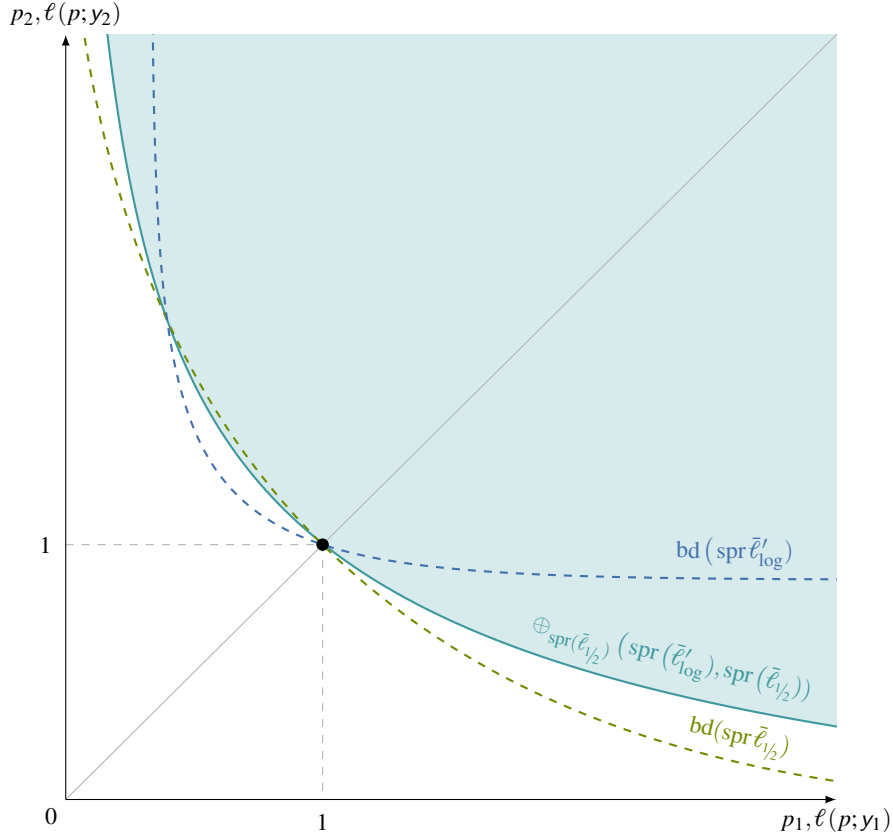

Superprediction sets recess to the positive orthant. Thus we also need to
ensure this property is preserved under our combination rules.  This is
captured by the main result of this subsection:
\begin{theorem}\label{thm:m_sum_rec_cone}
    Let $M\subseq\Rp^m$, $A_i\in\prop(\pcone)$ for $i\in[m]$. Then
    \begin{enumerate}
        \item $\rec(\msum(A_1,\dots,A_m)) = \pcone $, and \label{thm:m_sum_rec_cone1}
        \item $\rec(\dualmsum(A_1,\dots,A_m)) = \pcone.$ \label{thm:m_sum_rec_cone2}
    \end{enumerate}
\end{theorem}
\begin{proof}  %
    Let $\mu\in M$, $B_\mu \coloneqq  \sum_{i\in[m]}\mu_i\star A_i$, $C_\mu
    \coloneqq \inter_{i\in[m]} \mu_i\star A_i$. Then
    \begin{gather}
        \forall{\mu \in M},\ 
        \rec(B_\mu)
	\overset{\mathrm{P}\ref{prop:rccalc}(\ref{prop:rccalc_sum})}{\supseq} \sum_{i\in[m]}\rec(\mu_i\star A_i)  
	\overset{\mathrm{L}\ref{lem:epim_rec_cone}(\ref{lem:epim_rec_cone_eq})}{=}
        \sum_{i\in[m]} \rec(A_i)
	\overset{\mathrm{P}\ref{prop:rccalc}(\ref{prop:rccalc_cvx})}{=} \pcone. \label{eq:bums1}
    \end{gather}
    For the equivalent result for $C_\mu$ by Lemma \ref{lem:prop_intersect},
    $\inter_{i\in[m]} A_i \neq \varnothing$,  Proposition
    \ref{prop:rccalc}(\ref{prop:rccalc_inter}) implies
    \begin{gather}
        \forall{\mu \in M},\ 
        \rec(C_\mu) 
	\overset{\mathrm{P}\ref{prop:rccalc}(\ref{prop:rccalc_inter})}{=} \inter_{i\in[m]}\rec(\mu_i\star A_i)  
	\overset{\mathrm{L}\ref{lem:epim_rec_cone}(\ref{lem:epim_rec_cone_eq})}{=} \inter_{i\in[m]} \rec(A_i)= \pcone.\label{eq:bums2}
    \end{gather}
    It follows that
    \begin{gather}
        \rec(\msum(A_1,\dots,A_m))=  \rec(\union_{\mu\in M}
	B_\mu)\overset{\mathrm{P}\ref{prop:rccalc}(\ref{prop:rccalc_union})}{\supseq} \union_{\mu\in M} 
	\rec(B_\mu) 
        \overset{\eqref{eq:bums1}}\supseq  \pcone,\label{eq:m_sum_rec_cone_incl1}
        \shortintertext{and}
        \rec(\dualmsum(A_1,\dots,A_m)) = \rec(\union_{\mu\in M} C_\mu)
	\overset{\mathrm{P}\ref{prop:rccalc}(\ref{prop:rccalc_union})}{\supseq} \union_{\mu\in M} \rec(C_\mu) 
            \overset{\eqref{eq:bums2}}= \pcone.\label{eq:m_sum_rec_cone_incl2}
    \end{gather}
    
    The reverse inclusion is shown by contradiction. Suppose the inclusions
    in \eqref{eq:m_sum_rec_cone_incl1} and \eqref{eq:m_sum_rec_cone_incl2}
    are all strict. Then there exists
    \begin{gather}
	    d\in \rec(\msum(A_1,\dots,A_m)) \mbox{\ where\ }  d\notin \pcone.\label{eq:rec_cone_contradiction}
    \end{gather}
    From Proposition \ref{prop:rccalc}(\ref{prop:rccalc_seq}) this means there are
    sequences $(t_k)_{k\in\N}\searrow 0$, $t_k\in(0,1]$, and
    $(x_k)_{k\in\N}$, $x_k\in\rec(\msum(A_1,\dots,A_m))$ such that
    $\lim_{k\to\infty}t_k x_k = d$. From the definition of
    $\msum(A_1,\dots,A_m)$ there must be a sequence
    $(\mu^{k})_{k\in\N}\subseq M$ such that
    \begin{gather}
       x_k \in \mu^{k}_1 \star A_1 + \dots + \mu^{k}_m \star A_m.
    \end{gather} 
    By assumption we have $ A_i \subseq  \rec(A_i)=\pcone$ for $i\in[m]$, which implies $
        \mu_i^k \star A_i \subseq \pcone$ for $i\in[m]$. Since $\pcone$ is a cone we have 
    \begin{gather}
        t_kx_k \in \sum_{i\in[m]}t_k\mu_i^{k} \star A_i \subseq  \pcone.
    \end{gather}
    By hypothesis $\pcone$ is closed, and therefore contains
    $\lim_{k\to\infty}t_kx_k$, giving us a contradiction in
    \eqref{eq:rec_cone_contradiction}. Thus equality holds throughout in
    \eqref{eq:m_sum_rec_cone_incl1}. By an identical argument,
    \textit{mutatis mutandis},  applied to $\dualmsum$ we have equality 
    throughout in \eqref{eq:m_sum_rec_cone_incl2}, proving
    claim \ref{thm:m_sum_rec_cone2}.
\end{proof}
\begin{table}[t]
	    \arrayrulecolor{lightgray}
    \centering
    \small
    \begin{tabular}{l l l L{2.7cm} L{2.7cm}}
	    & $(A_i)_{i\in[m]}$ & $M$ &
	    \vspace*{-2mm}
	    ${\oplus_M(A_1,\dots,A_m)}$\vspace{2mm} &\vspace*{-2mm}
	    $\varoplus_M^*(A_1,\dots,A_m)$\vspace{2mm} \\ 
	 \toprule
        Theorem \ref{thm:msum_cvx_to_cvx} & $A_i\in\cvx(\bnch)$ & $M\in\cvx(\R^m)$ & convex & convex  \\ 
        Theorem \ref{thm:msum_closed} &
	$A_i\in\cvx(\pcone\!\setminus\!\{0\})$ &
	$M\in\cvx(\Rp^m\!\setminus\!\{0\})$ & closed & closed  \\ 
        Theorem \ref{thm:m_sum_rec_cone} & $A_i\in\prop(\pcone)$ & $M\in\Rp^m$ & $\pcone$-oriented & $\pcone$-oriented \\
        Proposition \ref{prop:msum_subset_prop} & $A_i\in\prop(\pcone)$ & $M\in\Rp^m\setminus{0}$ & subset of $\pcone\setminus{0}$ & subset of $\pcone\setminus{0}$ \\
        Corollary \ref{cor:msum_finale} & $A_i\in\prop(\pcone)$ & $M\in\prop(\Rp^m)$ & in $\prop(\pcone)$ & in $\prop(\pcone)$\\
        \bottomrule
    \end{tabular}
    \caption{Summary of $M$-sum structure results.}
    \label{tab:msum_summary}
\end{table}
The collection of results amassed in this section is summarised in
Table \ref{tab:msum_summary}, collectively they imply:
\begin{corollary}\label{cor:msum_finale}
    Let $M\subseq\prop(\Rp^m)$. Then $\msum$ maps from \(\prop(\pcone)^m\)
    to $\prop(\pcone)$, and $\dualmsum$ maps from \(\prop(\pcone)^m\) to
    $\prop(\pcone)$.
\end{corollary}
With Corollary \ref{cor:msum_finale} we see that the family of
superprediction sets of proper losses $\prop(\Rp^n)$ is closed under the
the $\msum$ and $\dualmsum$ operations.
Corollary \ref{cor:msum_finale} is illustrated in Figure \ref{fig:msum_finale}.

\subsection{The Functional \texorpdfstring{$M$}{M}-Sum}
\label{sec:the_functional_m-sum}

While the superprediction sets are our starting point, to be able to
derive proper loss functions we work with the support function of these
sets. The combination rules for the sets have an analog for their
corresponding support functions. We first introduce the functional $M$-sum
and in the following subsection justify the overloading of the naming and
notation. 

Let $f_1,\dots,f_m\colon\bnch \to \R$ be convex functions. For convex
$g\colon\R^m\to\Rx$ define the \Def{convex functional $M$-sum}
\begin{align}
    \bnch \ni x \mapsto & \Def{\msum_g(f_1,\dots,f_m)(x)}{g(f_1(x),\dots,
    f_m(x))}\label{eq:functional_msum_defn}\\
    \intertext{and the \Def{dual convex functional $M$-sum}}
    \bnch \ni x \mapsto 
 & \Def{\dualmsum_g(f_1,\dots,f_m)(x)}{\inf\set{g(f_1(a_1),\dots, f_m(a_{m}));
       a_{1} + \dots + a_{m} = x}}.\label{eq:cx_dual_functional_msum_defn}
\end{align}

If $f_1,\dots,f_m$ and $g$ are concave functions the above two notations
are overloaded with the \Def{concave functional $M$-sum} and \Def{dual
concave functional $M$-sum}:
\begin{align}
    \bnch \ni x \mapsto & \Def{\msum_g(f_1,\dots,f_m)(x)}{g(f_1(x),\dots, 
       f_m(x))}\label{eq:concave_msum_defn}\\
    \intertext{and}
    \bnch \ni x \mapsto & \Def{\dualmsum_g(f_1,\dots,
         f_m)(x)}{\sup\set{g(f_1(a_{1}),\dots, f_m(a_{m})); a_{1} + 
	 \dots + a_{m} = x}}.\label{eq:dual_concave_msum_defn}
\end{align}
The overload of notation can be defended since we will be either dealing
with convex or concave functions but not combinations of the two.

\subsection{Support Functions} 
\label{sub:support_functions}

We now justify the overload of the name $M$-sum for both the set operation
and the functional operation, via the following theorem.
\begin{theorem}\label{thm:functional_msum}
    Let  $A_i\subseq \bnch$ for $i\in[m]$ and either \begin{enumerate}
	    \item  $M\subseq\Rp^m$ with $\relint(M)≠\varnothing$, or
		    \label{thm:functional_msum_unbounded}
	\item  $M\subseq\R^m$, and $A_i$ for $i\in[m]$ are each bounded.
		\label{thm:functional_msum_bounded}
    \end{enumerate} Then
    $\msum_{\cxsprt_M}(\cxsprt_{A_1},\dots,\cxsprt_{A_m}) =
    \cxsprt_{\msum(A_1,\dots,A_m)}.$
\end{theorem}
\begin{proof}
    From the definition of the convex support function:
    \begin{gather}
	 \forall{x^*\in \bnch^*},\ \cxsprt_{\msum(A_1,\dots,A_m)}(x^*) =
	 \sup_{s\in \msum(A_1,\dots,A_m)}\inner{x^*;s}= \sup_{\mu\in
	 M}\,\,\sup_{s \in \!\!\sum\limits_{i\in[m]} \mu_i\star A_i}
	 \inner{x^*;s}.\label{eq:convex_sprt_epim}
    \end{gather}
    To simplify analysis it is useful to be able to replace the
    epimultiplication in \eqref{eq:convex_sprt_epim} with ordinary scalar
    multiplication. 
    
    \!\!{}Assume claim \ref{thm:functional_msum_unbounded}.
    Proposition \ref{prop:epim_continuity} shows epimultiplication is continuous in
    the Painleve--Kuratowski sense with respect to sequences of positive
    scalars, this means that \eqref{eq:convex_sprt_epim} implies the
    existence of a sequence $(\mu^n)_{n\in\N}$ with $\mu^n\in\relint(M)$
    such that 
    \begin{gather}
	\lim_{n\to\infty}\sup_{s \in\!\! \sum\limits_{i\in[m]}\! \mu_i^n\star A_i}
	\inner{x^*;s} = \lim_{n\to\infty}\sup_{s \in\!\!
		\sum\limits_{i\in[m]}\!
\mu_i^n·A_i} \inner{x^*;s} = \sup_{\mu\in M}\sup_{s\in \!\!
	\sum\limits_{i\in[m]}\!
\mu_i\star A_i} \inner{x^*;s},
    \end{gather}
    where the first equality follows since $\relint(M)\subseq\relint(\Rp^m)
    = \Rpp^m$. Therefore replacing $M$ by $\relint(M)$ in the supremum has
    no effect. Thus the expression in \eqref{eq:convex_sprt_epim}
    simplifies, giving $\forall{x^*\in\bnch^*},$
    \begin{align}
	    &\phantom{=}\  \ 
		\sup_{\mu\in M}\ \sup_{s \in\!\! \sum\limits_{i\in[m]}\!\mu_i\star A_i}
		\inner{x^*;s}\\
        &= \sup_{\mu\in \relint(M)}\ \sup_{s \in\!\!
		\sum\limits_{i\in[m]}\! \mu_i· A_i} \inner{x^*;s}\\
        &= \sup_{\mu\in \relint(M)}\sup_{a_{1} \in A_1}\left(\sup_{a_{2} 
               \in A_2}\left (\cdots \sup_{a_{m} \in A_m}
	          \inner{x^*; \mu_1 a_{1}+\cdots +\mu_m a_m}\right)\right)\\
        &= \sup_{\mu\in \relint(M)}\sup_{a_{1} \in A_1}\cdots \sup_{a_{m} \in A_m}\Biggl(\inner{x^*;\mu_1 a_{1}} + \dots + \inner{x^*;\mu_m a_{m}}\Biggr)\\
        &= \sup_{\mu\in \relint(M)}\left(\sup_{a_{1} \in
		A_1}\mu_1\inner{x^*;a_{1}} +\dots +\sup_{a_{m} \in
		A_m}\mu_m\inner{x^*;a_{m}}\right)\\
        &= \sup_{\mu\in \relint(M)}\left(\mu_1\cxsprt_{A_1}(x^*) +\dots
		+\mu_m\cxsprt_{A_m}(x^*)\right)\\
        &=  \sup_{\mu\in \relint(M)}\inner{(\cxsprt_{A_1}(x^*) ,\dots,
	      \cxsprt_{A_m}(x^*));\mu}\\
        &= \cxsprt_{\relint(M)}(\cxsprt_{A_1}(x^*) ,\dots ,\cxsprt_{A_m}(x^*))\\
        &\overset{\mathrm{L}\ref{lem:closed_convex_hull_cvx}}{=}  
        \cxsprt_{M}(\cxsprt_{A_1}(x^*) ,\dots ,\cxsprt_{A_m}(x^*)),\label{eq:msum_support}
    \end{align}
     as desired. 
     
     Now assume claim \ref{thm:functional_msum_bounded}. Since each of the sets
     $A_i$ are bounded, Proposition \ref{prop:rccalc}(\ref{prop:rccalc_bounded}) 
     implies that
     epimultiplication reduces to scalar multiplication. By a similar
     argument to \eqref{eq:msum_support} \emph{mutatis mutandis} (we no
     longer need to replace $M$ by $\relint(M)$) we are able to derive the
     same identity. 
\end{proof}

The relationship between convex and concave support functions \eqref{eq:cvx_to_ccv_sprt} yields the following corollary:

\begin{corollary}\label{cor:functional_msum}
    Let  $A_i\subseq \bnch$ for $i\in[m]$ and either \begin{enumerate}
            \item  $M\subseq\Rp^m$ with $\relint(M)≠\varnothing$, or
            \item  $M\subseq\R^m$, and $A_i$ for $i\in[m]$ are each bounded.
        \end{enumerate} Then $\msum_{\cvsprt_M}(\cvsprt_{A_1},\dots,\cvsprt_{A_m}) =\cvsprt_{\msum(A_1,\dots,A_m)}$.
\end{corollary}

\begin{corollary}\label{cor:proper_loss_to_proper_loss}
Let $\ell_1,\dots,\ell_m$ be a sequence of proper loss functions with
conditional Bayes risks $\minL_1,\dots,\minL_m$. Let
$\minM\colon\Rp^m\to\Rx$ be a conditional Bayes risk function. Let
\begin{gather}
	\jay\in\subdiff\msum_{\minM}(\minL_1,\dots,\minL_m ) \mbox{\ and\ }
	\kay\in\subdiff\dualmsum_{\minM}(\minL_1,\dots,\minL_m).
\end{gather}
Then $\jay$ and $\kay$ are proper loss functions.
\end{corollary}
This corollary shows how we can create new proper losses from old via the
$M$-sums and dual $M$-sums.  The following is a restatement in the
terminology of $M$-sums of an old result which we require subsequently.

\begin{proposition}[\protect{\citealt[Theorem~D.4.3.1]{hiriarturruty2001fca}}]\label{thm:composition_subdifferential}\ \\
Let $f_1,\dots,f_m\colon\R^n\to\R$ be convex. Let $g\colon\R^m\to\R$ be
convex and increasing component-wise in the sense that $x\succeq_{\Rp^m}y$
implies $g(x)≥ g(y)$. Define $F \coloneqq  x\mapsto (f_1(x),\dots,f_m(x))$. Then
    \begin{align}
        \forall{x\in\R^n},\ \subdiff (g°F)(x) = \msum_{\subdiff g(F(x))}(\subdiff f_1(x),\dots,\subdiff f_m(x)).\label{eq:subdiff_is_msum}
    \end{align}
\end{proposition}

\begin{corollary}\label{cor:m-sum_subdifferential}
Let $(\minL_i)_{i\in[m]}$, and $\minM$ be as defined above, and let
$\ell_i=\subdiff\minL_i$ for $i\in[m]$ and 
$\emm=\subdiff \minM$. 
Then
    \begin{align}
        \forall{p\in \probm},\ 
        &\underset{n\times m}{\begin{pmatrix}
                    \ell_1(p, \rv y_1) & \dots  & \ell_m(p, \rv y_1)\\
                            \vdots     & \ddots & \vdots            \\
                    \ell_1(p, \rv y_n) & \dots  & \ell_m(p, \rv y_n)
                \end{pmatrix}}
        \underset{m\times 1}{\emm(\minL_1(p),\dots,\minL_m(p))\vphantom{\begin{pmatrix}
                    \ell_1(p, \rv y_1) & \dots  & \ell_m(p, \rv y_1)\\
                            \vdots     & \ddots & \vdots            \\
                    \ell_1(p, \rv y_n) & \dots  & \ell_m(p, \rv y_n)
                \end{pmatrix}}}
        \in \subdiff \msum_{\minM}(\minL_1,\dots,\minL_m)(p)\subseq \R^n.
    \end{align}
\end{corollary}

\begin{lemma}\label{lem:dual_msum_convex}
    Let $f_1,\dots,f_m\colon\bnch\to\R$ be convex (resp.\,concave)
    functions then $\dualmsum_{\cxsprt_M}\!(f_1,\dots,f_m)$ is convex (resp.\
    $\dualmsum_{\cvsprt_M}(f_1,\dots,f_m)$ is concave).
\end{lemma}
\begin{proof}
    Let $f_1,\dots,f_m\colon\bnch\to\R$ be convex. Fix arbitrary $(a_i)_{i\in[m]}$ and $(b_i)_{i\in[m]}$ with $a_i,b_i\in \dom(f_i)$ for $i\in[m]$, and pick an arbitrary $t\in(0,1)$. Then
    \begin{alignat}{4}
        &\forall{i\in[m]},\ &&&f_i(ta_i + (1-t)b_i)&\ ≤\  t f_i(a_i) + (1-t)f_i(b_i)\\
        &\mathcslap{\implies}{\forall{i\in[m]},\ }&&&
        \sup_{\mu\in M}\sum_{i\in[m]}\mu_if_i(ta_i + (1-t)b_i)
            &\ ≤\ t\sup_{\mu\in M}\sum_{i\in[m]}\mu_i f_i(a_i) + (1-t)\sup_{\nu\in M}\sum_{j=1}^m\nu_jf_j(b_j).
    \end{alignat}
    Since $(a_i)_{i\in[m]}$ and $(b_i)_{i\in[m]}$ are arbitrary, taking the infimum over both sides yields
    \begin{align}
        \MoveEqLeft[6]\inf_{\substack{a_1+\dots+a_m=x\\b_1+\dots+b_m=y}}
	   \left(\sup_{\mu\in M}\sum_{i\in[m]}\mu_if_i(ta_i + (1-t)b_i)\right)\\
            &≤
	    \inf_{\substack{a_1+\dots+a_m=x\\b_1+\dots+b_m=y}}\left(t\sup_{\mu\in
	    M}\sum_{i\in[m]}\mu_i f_i(a_i) + (1-t)\sup_{\nu\in
	    M}\sum_{j=1}^m\nu_jf_j(b_j)\right)\\
            &=  t\g(\inf_{a_1+\dots+a_m=x}\sup_{\mu\in M}\sum_{i\in[m]}\mu_i f_i(a_i)) + (1-t)\g(\inf_{b_1+\dots+b_m=y}\sup_{\nu\in M}\sum_{j=1}^m\nu_jf_j(b_j))\\
	    &=  t\dualmsum_{\cxsprt_M}(f_1,\dots,f_m)(x) + 
	    (1-t)\dualmsum_{\cxsprt_M}(f_1,\dots,f_m)(y).
	    \label{eq:dual-m-sum-cvx-1}
    \end{align}
    To complete the proof
    \begin{align}
        \MoveEqLeft[3]\inf_{\substack{a_1+\dots+a_m=x\\b_1+\dots+b_m=y}}
	   \left(\sup_{\mu\in M}\sum_{i\in[m]}\mu_if_i(ta_i + (1-t)b_i)\right) \\
        &= \inf\set{\sup_{\mu\in M}\sum_{i\in[m]}\mu_if_i(ta_i +
	(1-t)b_i); \begin{aligned} &\forall{i\in[m]},\  a_i,b_i \in\dom(f_i)
			  \mbox{\ and\ }\\
		     & \sum_{i\in[m]}a_i = x\mbox{\ and\ }\sum_{i\in[m]}b_i = y
	     \end{aligned}}\\
        &= \inf\set{\sup_{\mu\in
	M}\sum_{i\in[m]}\mu_if_i(c_i);\forall{i\in[m]},\  c_i
	\in\dom(f_i)\mbox{\ and\ }\sum_{i\in[m]}c_i = tx + (1-t)y}\\
        &= \dualmsum_{\cxsprt_M}(f_1,\dots,f_m)(tx + (1-t)y),
	    \label{eq:dual-m-sum-cvx-2}
    \end{align}
    where the second equality follows since $\dom(f_i)$ is convex for each
    $i\in[m]$. Since $\eqref{eq:dual-m-sum-cvx-2} \le
    \eqref{eq:dual-m-sum-cvx-1}$, we have that
    $\dualmsum_{\cxsprt_M}(f_1,\dots,f_m)$ is convex.
    By an identical argument, \textit{mutatis mutandis}, the
    concave result for the concave functional $M$-sum follows.
\end{proof}
\begin{lemma}\label{lem:conjugate_epim_sprt}
    Let $S\in\cvx(\bnch)$, and $\mu\in\Rp$. Then 
    $x\mapsto\sup_{x^*\in\dom(\cxsprt_S)}
    \big(\inner{x^*;x} - \mu\cxsprt_S(x^*)) = (\cxsprt_{\mu\star A})^*$.
\end{lemma}
\begin{proof} Let $\mu>0$. Then $\mu\cxsprt_{S} = \cxsprt_{\mu A} =
	\cxsprt_{\mu\star A}$ \citep[Theorem~C.3.3.2]{hiriarturruty2001fca}
	and the result follows by conjugation. Now assume $\mu = 0$. Then
	from Lemma \ref{lem:sprt_domain}, 
	$\overline{\dom(\cxsprt_S)} = -(\rec S)^*$ and $\forall{x\in\bnch}$,
    \begin{align}
	\sup_{x^*\in\dom(\cxsprt_S)}\left(\inner{x^*;x} -
	\mu\cxsprt_S(x^*)\right)
        &=\sup_{x^*\in - (\rec S)^*}\inner{x^*;x}\\
        &=\sup_{x^*\in(\rec S)^*}-\inner{x^*;x}\\
        &=\begin{cases}
             0 & x \in \rec(S)\\
             \infty      & \text{otherwise},
        \end{cases}
        \label{eq:indicator_thing}
    \end{align}
    where the final equality follows from the definition of the dual cone \eqref{eq:dual_cone_def}. Noticing \eqref{eq:indicator_thing} is the indicator of the set $\rec(S)$, it therefore is also the convex conjugate of the support function $\cxsprt_{S}$ \citep[Theorem~13.2, p.~114]{Rockafellar:1970}.
\end{proof}
\begin{theorem}\label{thm:dual_functional_msum}
    Let $M\in\rdnt(\R^m)$ be compact, $N\in\shdy(\R^m)$, and $A_i \in
    \cvx(\bnch)$ for $i\in[m]$ with $\cap_{i\in[m]}A_i≠\varnothing$.  Then
    \begin{enumerate}
        \item $\cxsprt_{\dualmsum(A_1,\ldots,A_m)} = \overline{\dualmsum_{\cxsprt_M}(\cxsprt_{A_1},\ldots, \cxsprt_{A_m})}$, and \label{thm:dual_functional_msum1}
        \item $\cvsprt_{\dualmsum_N(A_1,\ldots,A_m)} = \overline{\dualmsum_{\cvsprt_N}(\cvsprt_{A_1},\ldots, \cvsprt_{A_m})}$. \label{thm:dual_functional_msum2}
    \end{enumerate}
\end{theorem}
\begin{proof}
    \ref{thm:dual_functional_msum1}. Let $D\coloneqq
    \dom(\cxsprt_{A_1}) \times \cdots \times \dom(\cxsprt_{A_m})\subseq
    (\bnch^*)^m$ and  $a\coloneqq (a_1,\dots,a_m)\in D$. The
    Legendre-Fenchel conjugate \eqref{eq:convex_conjugate} of the
    right-hand side    is $\forall{x\in\bnch},$
    \begin{align}
        \MoveEqLeft[7]\ \g(\dualmsum_{\cxsprt_M}(\cxsprt_{A_1}, \dots, \cxsprt_{A_m}))^* (x) \\
        &=
        \sup_{x^*\in \bnch^*}\left(\inner{x^*;x} -
            \inf_{ a_1 + \dots + a_m = x^*}
            \cxsprt_M((\cxsprt_{A_1}(a_1), \dots,
            \cxsprt_{A_m}(a_m)))\right)\\
        &= \sup_{x^*\in \bnch^*}\left(\sup_{ a_1 + \dots + a_m =
	x^*}{\inner{x^*;x}-\cxsprt_M(( 
                                    \cxsprt_{A_1}(a_1),\ldots,
                                    \cxsprt_{A_m}(a_m)))}\right)\\
        &= \sup_{a_1,\dots,a_m\in \bnch^*}\left(
            \inner{a_1;x} +\dots +\inner{a_m;x} 
            - 
            \cxsprt_M((\cxsprt_{A_1}(a_1),
            \ldots,\cxsprt_{A_m}(a_m)))\right)
            \\
        &= \sup_{a\in D}\left(
            \inner{a_1;x} +\dots +\inner{a_m;x} 
            -\cxsprt_M((\cxsprt_{A_1}(a_1),\dots,\cxsprt_{A_m}(a_m)))\right)\\
        &= \sup_{a\in D}\left(
            \inner{a_1;x} +\dots +\inner{a_m;x} 
            -\sup_{\mu\in
	    M}\inner{(\cxsprt_{A_1}(a_1),\dots,\cxsprt_{A_m}(a_m));\mu}\right)\\
        &= \sup_{a\in D}\inf_{\mu\in M}\g\left(
            \inner{a_1;x} +\dots +\inner{a_m;x} 
            -\inner{(\cxsprt_{A_1}(a_1),\dots,\cxsprt_{A_m}(a_m));\mu}\right)\\
        &= \sup_{a\in D}\inf_{\mu\in M} L_x(a,\mu), \label{eq:dfm1}
    \end{align}
    where for $x\in\bnch$ we have $L_x\colon D\times M\to\R$ with
    \begin{gather}
	    \forall{a\in D},\ \forall \mu\in M,\ L_{x}(a,\mu)\coloneqq 
        \inner{a_{1};x} +\dots + \inner{a_{m};x} - 
        \inner{(\cxsprt_{A_1}(a_{1}),\dots,\cxsprt_{A_m}(a_{m}));\mu}.
    \end{gather}
    
    Immediately $L_x(a,\mu)$ is concave and upper semi-continuous in $a$, and convex and lower semi-continuous in $\mu$. Both $D$ and $M$ are convex and $M$ is compact, and so we can apply \citeauthor{Sion1958}'s minimax theorem \citeyearpar{Sion1958} and write
    \begin{align}
        \sup_{a\in D}\inf_{\mu\in M} L_x(a,\mu) 
        &= \inf_{\mu\in M}\sup_{a\in D} L_x(a,\mu)\\
        &= \inf_{\mu\in M}\:\sum_{i\in[m]}\sup_{x^*\in
		\dom(\cxsprt_{A_i})}\left(\inner{x^*;x}-
		\mu_i\cxsprt_{A_i}(x^*)\right)\\
        &= \inf_{\mu\in M}\sum_{i\in[m]} r_{A_i,\mu_i}(x),\label{eq:dfm2}
    \end{align}
    where $r_{A_i,\mu_i}(x) \coloneqq  \sup_{x^*\in \dom
	    \cxsprt_{A_i}}(\inner{x^*;x}- \mu_i\cxsprt_{A_i}(x^*))$.
	    Examining the functions $r_{A_i,\mu_i}$ with
	    Lemma \ref{lem:conjugate_epim_sprt}, we see that $r_{A_i,\mu_i} =
	    \cxsprt_{\mu_i\star A_i}^*$ and in summary we have
		$\forall{x\in\bnch}$,
    \begin{align}
	\g(\dualmsum_{\cxsprt_M}(\cxsprt_{A_1}, \dots, \cxsprt_{A_m}))^* (x)
        &\overset{\eqref{eq:dfm1}}{=} \sup_{a\in D}\inf_{\mu\in M} L_x(a,\mu)\\
        &\overset{\eqref{eq:dfm2}}{=} \inf_{\mu\in M}\sum_{i\in[m]} r_{A_i,\mu_i}(x)\\
	&\overset{\mathrm{L}\ref{lem:conjugate_epim_sprt}}{=} \inf_{\mu\in M}(\cxsprt_{\mu_1\star A_1}^*(x) +\dots + \cxsprt_{\mu_m\star A_m}^*(x))\\
        &\overset{\hphantom{\eqref{eq:dfm2}}}{=} \inf_{\mu\in M}(\cxsprt_{\mu_1\star A_1} \infconv\cdots \infconv \cxsprt_{\mu_m\star A_m})^*(x),\label{eq:conjugate}
    \end{align}
    where $\infconv$ denotes infimal convolution \eqref{eq:infimal-convolution-def}.
    We now apply the biconjugate theorem to both sides to obtain
        $\forall{x^*\in\bnch^*}$,
    \begin{align}
	\g(\dualmsum_{\cxsprt_M}(\cxsprt_{A_1}, \dots, \cxsprt_{A_m}))^{**}(x^*)
        &= \sup_{x\in\bnch}\left(\inner{x^*;x} - \inf_{\mu\in
	M}(\cxsprt_{\mu_1\star A_1} \infconv\cdots \infconv
	\cxsprt_{\mu_m\star A_m})^*(x)\right)\\
        &= \sup_{\mu\in M}\sup_{x\in\bnch}\left(\inner{x^*;x}
	-(\cxsprt_{\mu_1\star A_1} \infconv\cdots \infconv
	\cxsprt_{\mu_m\star A_m})^*(x)\right)\\
        &= \sup_{\mu\in M}\overline{(\cxsprt_{\mu_1\star A_1} \infconv\cdots \infconv \cxsprt_{\mu_m\star A_m})}(x^*)\\
        &= \sup_{\mu\in M}\cxsprt_{\mu_1\star A_1 \cap \dots \cap \mu_m\star A_m}(x^*)
        \\
        &= \cxsprt_{\!\!\union\limits_{\mu\in M} \!
                            \mu_1 \star A_1 \cap \cdots 
                            \cap \mu_m\star A_m}(x^*)\\
        &=
        \cxsprt_{\dualmsum(A_1,\dots,A_m)}(x^*),
    \end{align}
    where the final two equalities are due to
    \citet[Theorem~C.3.3.2]{hiriarturruty2001fca}. From Lemma
    \ref{lem:dual_msum_convex} we have
    $\g(\dualmsum_{\cxsprt_M}(\cxsprt_{A_1}, \dots, \cxsprt_{A_m}))^{**}=
    \overline{\dualmsum_{\cxsprt_M}(\cxsprt_{A_1}, \dots, \cxsprt_{A_m})}$,
    which completes the proof of claim \ref{thm:dual_functional_msum1}.

    We now turn our attention to claim \ref{thm:dual_functional_msum2}. Let
    $E\coloneqq \dom(\cvsprt_{A_1}) \times \cdots \times
    \dom(\cvsprt_{A_m})\subseq (\bnch^*)^m$ and  $a\coloneqq (a_1,\dots,a_m)\in E$. The concave conjugate \eqref{eq:concave_conjugate} of the right-hand side
    is $\forall{x\in\bnch}$,
    \begin{align}
        \MoveEqLeft[6]\ \g(\dualmsum_{\cvsprt_N}(\cvsprt_{A_1}, \dots, \cvsprt_{A_m}))_* (x) \\
        &=
        \inf_{x^*\in \bnch^*}\left(\inner{x^*;x} -
            \sup_{ a_1 + \dots + a_m = x^*}
            \cvsprt_N((\cvsprt_{A_1}(a_1), \dots,
            \cvsprt_{A_m}(a_m)))\right)\\
        &= \inf_{a\in E}\left(
            \inner{a_1;x} +\dots +\inner{a_m;x} 
            -\inf_{\mu\in
	    N}\inner{(\cvsprt_{A_1}(a_1),\dots,\cvsprt_{A_m}(a_m));\mu}\right)\\
        &= \inf_{a\in E}\sup_{\mu\in M}\left(
            \inner{a_1;x} +\dots +\inner{a_m;x} 
            -\inner{(\cvsprt_{A_1}(a_1),\dots,\cvsprt_{A_m}(a_m));\mu}\right)\\
        &= \inf_{a\in E}\sup_{\mu\in N} E_x(a,\mu), \label{eq:dfm4}
    \end{align}
    where for $x\in\bnch$ we have $E_x\colon E\times N\to\R$ with
    \begin{gather}
	    \forall{a\in E},\ \forall\mu\in M,\ E_x(a,\mu)\coloneqq 
        \inner{a_{1};x} +\dots + \inner{a_{m};x} - 
        \inner{(\cvsprt_{A_1}(a_{1}),\dots,\cvsprt_{A_m}(a_{m}));\mu}.
    \end{gather}
    
    We immediately have that $E_x(a,\mu)$ is convex and lower
    semi-continuous in $a$, and concave and upper semi-continuous in $\mu$.
    In order to apply Lemma \ref{thm:minimax} we need to find certain sets $E'\subseq E$ and $N'\subseq N$ such that we satisfy \eqref{eq:minimax_cond}. From the definition of $E$ we have $0\in E$. From the 1-homogeneity of the functions $\cvsprt_{A_m}$ we know $E_x(0,\mu)=0$ for all $\mu\in N$. Therefore
       $\forall{\mu\in N}, $
    \begin{gather}
       \inf_{a\in E} E_x(a,\mu) ≤ E_x(0,\mu) \iff  \inf_{a\in E} E_x(a,\mu) ≤ \inf_{b\in\single{0}}E_x(b,\mu)=0. \label{eq:minimax_cond_first}
    \end{gather}
    Let $N'\coloneqq  N\cap\ball_{\agauge_N(0)+1}\subseq N$, where
    $\ball_{\agauge_N(0)+1}$ is the norm ball of radius $\agauge_N(0)+1$.
    Then $N'\subseq N$ is compact, and $E'\coloneqq \single{0}\subseq E$ is convex and nonempty. From \eqref{eq:minimax_cond_first} we have
    \begin{gather}
        \inf_{a\in E}\sup_{\mu\in N} E_x(a,\mu) ≤ 0 = \inf_{a\in E'}\sup_{\mu\in N'}E_x(a,\mu),
    \end{gather}
    and therefore we satisfy \eqref{eq:minimax_cond} and we can apply Lemma \ref{thm:minimax}.
    
    From here we proceed with an argument that parallels the proof for
    claim \ref{thm:dual_functional_msum1}, \emph{mutatis mutandis} (infima and
    suprema exchanged, convex conjugates replaced with concave conjugates,
    inf convolution replaced with sup convolution) and we find
    $\cvsprt_{\dualmsum_{N}(A_1,\ldots,A_m)} =
    \overline{\dualmsum_{\cvsprt_N}(\cvsprt_{A_1},\ldots, \cvsprt_{A_m})}$,
    completing the proof of claim \ref{thm:dual_functional_msum2}.
\end{proof}

\subsection{The \texorpdfstring{$M$}{M}-Sum Polar}
\label{sub:the_m-sum_polar}

The polar (antipolar) operation plays an important role in our theory, and
thus it is natural to ask how polarity interacts with the $M$-sum
operations.  We first need the following proposition.
\begin{proposition}\label{prop:polar_prop}
    Let $A_i\subseq {\bnch}$ for $i\in I$ be an arbitrary collection of sets with index set $I$, let $R\in\rdnt(\bnch)$ and $S\in\shdy(\bnch)$. Then
    \begin{enumerate}
        \item  $\g(\union_{i\in I}A_i)^\polar= \inter_{i\in I}\g(A_i^\polar)$, and $\g(\union_{i\in I}A_i)^\apolar= \inter_{i\in I}\g(A_i^\apolar)$; \label{prop:polar_prop_interunion}
        \item $R = \lev_{≤1}(\gauge_{R})$, and $S = \lev_{≥1}(\agauge_{S})$. \label{prop:polar_prop_gauge}
    \end{enumerate}
\end{proposition}
\begin{proof}
    The polar identity in claim \ref{prop:polar_prop_interunion} is a standard
    result in functional analysis \citep[Corollary~5.104,
    p.~218]{aliprantis2006infinite}. The proof for the antipolar is
    identical modulo the reversal of some inequalities.
    Claim \ref{prop:polar_prop_gauge} is immediate from the definition of
    star-shaped and co-star-shaped sets in \S\ref{sec:gauge_functions_and_polar_duality}.
\end{proof}
We now show that the polarity operation preserves the radiant and shady
nature of the sets.
\begin{theorem}\label{thm:radshad_to_radshad}
    Let $A_i\in\rdnt(\bnch)$ and $B_i\in\shdy(\pcone)$ for $i\in[m]$, $M\in\rdnt(\R^m)$ compact, $N\in\shdy(\Rp^m)$. Then
        \begin{enumerate}
            \item $\msum(A_1,\dots,A_m) \in \rdnt(\bnch)$, and $\dualmsum(A_1,\dots,A_m) \in \rdnt(\bnch)$;
            \item $\msum_N(B_1,\dots,B_m) \in \shdy(\bnch)$, and $\dualmsum_N(B_1,\dots,B_m) \in \shdy(\bnch)$.
        \end{enumerate}
\end{theorem}
\begin{proof}
    We take the first case:
    \begin{align}
        (0,1]·\msum(A_1,\dots,A_m)
        & = (0,1]·\union_{\mu\in M} \sum_{i\in[m]}\mu_i\star A_i\\
        &= \union_{\mu\in M} \sum_{i\in[m]} (0,1]·\mu_i\star A_i\\
        &= \union_{\mu\in M} \sum_{i\in[m]}\mu_i\star (0,1]·A_i\\
        &= \union_{\mu\in M} \sum_{i\in[m]}\mu_i\star A_i,
    \end{align}
    and so it is clear that $\msum(A_1,\dots,A_m)$ is star-shaped. By the same argument, \textit{mutatis mutandis} $\dualmsum(A_1,\dots,A_m)$ is star-shaped, and $\msum_N(B_1,\dots,B_m)$ and $\dualmsum_N(B_1,\dots,B_m)$ are co-star-shaped. 
    
    Theorem \ref{thm:msum_cvx_to_cvx} guarantees convexity, and Theorem
    \ref{thm:msum_closed} guarantees closure. Proposition \ref{prop:msum_subset_prop} guarantees the exclusion of the origin in the shady case, and so all that is left is to show $0\in\interior(\msum(A_1,\dots,A_m))$, and $0\in\interior(\dualmsum(A_1,\dots,A_m))$.
    
    By hypothesis $A_i\in\rdnt(\bnch)$ for each $i\in[m]$ and so $0\in\interior(A_i)$. By definition there exists $(r_i)_{i\in[m+1]}$ with $r_i\in\Rpp$ such that
    \begin{gather}
	    \forall{i\in[m]},\ \ball_{r_i}\subseq A_i\subseq\bnch\mbox{\ and\ } \ball_{r_{m+1}}\subseq M\subseq\R^m.
    \end{gather}
    Let $r\coloneqq \min_{i\in[m]}r_i$,
    $\mu'\in\Rpp^m\cap\ball_{r_{m+1}}\subseq M$, and $\mu'_{\min}
    \coloneqq  \min_{i\in[m]}\mu_i'>0$. Then
    \begin{gather}
        \msum(A_1,\dots,A_m) = \union_{\mu\in M}\sum_{i\in[m]} \mu_i\star A_i 
        \supseq \union_{\mu\in M}\sum_{i\in[m]} \mu_i\star \ball_{r}
        \supseq \sum_{i\in[m]} \mu_i'\ball_{r}
        \supseq \ball_{r·\mu'_{\min}},
        \shortintertext{and}
        \dualmsum(A_1,\dots,A_m) = \union_{\mu\in M}\inter_{i\in[m]} \mu_i\star A_i 
        \supseq \union_{\mu\in M}\inter_{i\in[m]} \mu_i\star \ball_{r}
        \supseq \inter_{i\in[m]}\mu'_{i}\cdot\ball_{r}
        = \ball_{r·\mu'_{\min}},
    \end{gather}
    completing the proof.
\end{proof}
\begin{lemma}\label{lem:gauge_epim}
    Let $R\in\rdnt(\bnch)$, $S\in\shdy(\bnch)$ with $S\subseq\rec(S)$, $\alpha≥0$. Then \begin{enumerate}
        \item $\lev_{≤\alpha}\gauge_R = \alpha\star R$, and \label{lem:gauge_epim_rad}
        \item $\lev_{≥\alpha}\agauge_S = \alpha\star S$. \label{lem:gauge_epim_shad}
    \end{enumerate}
\end{lemma}
\begin{proof}
    Suppose $\alpha>0$ and let $x^*\in\lev_{≤\alpha}(\gauge_{R})$, then exploiting the 1-homogeneity of $\gauge_{R}$:
    \begin{gather}
        \gauge_{R}(x^*) ≤ \alpha \iff \oneon{\alpha} \gauge_{R}(x^*) ≤ 1 \iff 
        \gauge_{\alpha R}(x^*) ≤ 1.
    \end{gather}
    Thus applying Proposition
    \ref{prop:polar_prop}(\ref{prop:polar_prop_gauge}) we find
    $\lev_{≤\alpha}\gauge_R = \lev_{≤1}( \gauge_{\alpha R})=\alpha R$.
    \citet[Proposition 2.3]{Penot2000} prove $\lev_{=0}(\gauge_{R}) =
    \rec(R)$. Since gauge functions are nonnegative \eqref{eq:gauge_defn},
    this shows claim \ref{lem:gauge_epim_rad}.
    
    Suppose $\alpha>0$ and let $x^*\in\lev_{≥\alpha}(\agauge_{S})$, then
    appealing to the 1-homogeneity of $\agauge_{S}$:
    \begin{gather}
        \agauge_{S}(x^*) ≥ \alpha \iff \oneon{\alpha} \agauge_{S}(x^*) ≥ 1 \iff 
        \agauge_{\alpha S}(x^*) ≥ 1.
    \end{gather}
    Thus applying Proposition
    \ref{prop:polar_prop}(\ref{prop:polar_prop_gauge}) we find
    $\lev_{≥\alpha}(\agauge_S) = \lev_{≥1}( \agauge_{\alpha S})=\alpha S$.
    Now suppose $\alpha=0$. \citet[Proposition 2.4]{Penot2000} prove
    $\lev_{=0}(\agauge_{S}) = \rec(S)\setminus\cone(S)$, and
    $\lev_{>0}(\agauge_{S}) = \cone(S)$, giving $\lev_{≥0}(\agauge_S)
    =\rec(S)$. This shows claim \ref{lem:gauge_epim_shad} and completes the
    proof.
\end{proof}
The following general duality result extends a range of classical results,
as well as results in \citep{Seeger1990}; it demonstrates the appealling
fact that the polar of an $M$-sum of $(A_i)_i$ is the dual (polar-$M$)-sum of
the polars of $(A_i)_i$; and similarly for antipolars.
\begin{theorem}
    \label{thm:dual_msum}
    Let $A_i\in\rdnt(\bnch)$ and $B_i\in\shdy(\bnch)$ with $B_i\subseq\rec(B_i)$ for $i\in[m]$, $M\in\rdnt(\R^m)$, and $N\in\shdy(\Rp^m)$. Assume $M$ and $A_i$ for $i\in[m]$ are compact. Then
    \begin{enumerate}
        \item $\msum(A_1,\dots,A_m)^{\polar} = {\dualmsum_{M^\polar}(A_1^\polar, \dots,A_m^\polar)}$, and \label{thm:dual_msum_cvx}
        \item $\msum_N(B_1,\dots,B_m)^{\apolar} = {\dualmsum_{N^\apolar}(B_1^\apolar, \dots,B_m^\apolar)}$. \label{thm:dual_msum_ccv}
    \end{enumerate}
\end{theorem}
\begin{proof}
    Calculating the polar of the left hand side of claim \ref{thm:dual_msum_cvx} we have
    \begin{align}
        \g\left(\msum(A_1,\dots,A_m)\right)^{\polar}
        &\overset{\:\:\,\eqref{eq:polar_antipolar_dfn}\:\:\,}{=}
        \lev_{≤1}\left(\cxsprt_{\msum(A_1,\dots,A_m)}\right)\\
	&\overset{\:\mathrm{T}\ref{thm:functional_msum}\:}{=}
        \lev_{≤1}\left(\msum(\cxsprt_{A_1},\dots,\cxsprt_{A_m})\right)\\
	&\!\!\!\overset{\hphantom{\mathrm{L}\ref{lem:gauge_epim}(\ref{lem:gauge_epim_rad})}}{=}
        \lev_{≤1}\left(x^*\mapsto
	\cxsprt_M((\cxsprt_{A_1}(x^*),\dots,\cxsprt_{A_m}(x^*)))\right)\\
        &\overset{\:\,\eqref{eq:gauge_support_polar}\:\,}{=}
        \lev_{≤1}\left(x^*\mapsto
	\gauge_{M^\polar}((\gauge_{A^\polar_1}(x^*),\dots,\gauge_{A^\polar_m}(x^*)))\right)\\
	&\!\!\overset{\mathrm{P}\ref{prop:polar_prop}(\ref{prop:polar_prop_gauge})}{=}
        \left\{x^* \in \bnch^*\st
		(\gauge_{A^{\polar}_1}(x^*),\dots,\gauge_{A^{\polar}_m}(x^*))\in
	M^\polar\right\}\\
	&\!\!\overset{\hphantom{\mathrm{L}\ref{lem:gauge_epim}(\ref{lem:gauge_epim_rad})}}{=}
        \union_{\mu^*\in M^\polar}\set{x^* \in \bnch^*; (\gauge_{A^{\polar}_1}(x^*),\dots,\gauge_{A^{\polar}_m}(x^*)) =\mu^*}\\
	&\!\!\overset{\hphantom{\mathrm{L}\ref{lem:gauge_epim}(\ref{lem:gauge_epim_rad})}}{=}
        \union_{\mu^*\in M^\polar}\set{x^* \in \bnch^*; \forall{i\in[m]},\
	\gauge_{A^{\polar}_i}(x^*) =\mu_i^*}.\label{eq:dual_msum_1}
    \end{align}
    Observe that $M^\polar$ is star-shaped and closed, thus $0\in M^\polar$ and $[0,1]·M^\polar = M^\polar$. Hence
    \begin{align}
        \g\big(\msum(A_1,\dots,A_m))^{\polar}
        &\overset{\eqref{eq:dual_msum_1}}{=}
        \union_{\mu^*\in M^\polar}\set{x^* \in \bnch^*; \forall{i\in[m]},\ \gauge_{A^{\polar}_i}(x^*) =\mu_i^*}\\
	& \, =
        \union_{\mu^*\in [0,1]·M^\polar}\set{x^* \in \bnch^*;
		\forall{i\in[m]},\ \gauge_{A^{\polar}_i}(x^*) =\mu_i^*}\\
	&=
        \union_{\mu^*\in M^\polar}\union_{\lambda\in[0,1]}\set{x^* \in
		\bnch^*; \forall{i\in[m]},\ \gauge_{A^{\polar}_i}(x^*) =\lambda\mu_i^*}\\
	& =
        \union_{\mu^*\in M^\polar}\set{x^* \in \bnch^*; \forall{i\in[m]},\ \gauge_{A^{\polar}_i}(x^*) ≤\mu_i^*}\\
	& =
        \union_{\mu^*\in M^\polar}\inter_{i\in[m]}\lev_{≤\mu_i^*}(\gauge_{A^{\polar}_i})\\
	&\!\!\!\!\overset{\smash{\mathrm{L}\ref{lem:gauge_epim}(\ref{lem:gauge_epim_rad}})}{=}
        \union_{\mu^*\in M^\polar}\inter_{i\in[m]}\mu_i^* \star A^{\polar}_i\\
	&\!\!\!\overset{\hphantom{\mathrm{L}\ref{lem:gauge_epim}(\ref{lem:gauge_epim_rad})}}{=}
        \dualmsum_{M^\polar}(A_1^\polar, \dots,A_m^\polar),\label{eq:pp1}
    \end{align}
    which completes the proof of claim \ref{thm:dual_msum_cvx}.  The proof of
    claim \ref{thm:dual_msum_ccv} proceeds much like the proof of
    claim \ref{thm:dual_msum_cvx}.   Calculating the left hand side of
    claim \ref{thm:dual_msum_ccv}, by a similar argument to \eqref{eq:pp1},
    \emph{mutatis mutandis}
    \begin{align}
        \g\left(\msum_N(B_1,\dots,B_m)\right)^{\apolar}
        &\overset{\:\:\,\eqref{eq:polar_antipolar_dfn}\:\:\,}{=}
        \lev_{≥1}\left(\cvsprt_{\msum_N(B_1,\dots,B_m)}\right)\\
	&\overset{\:\mathrm{C}\ref{cor:functional_msum}\:}{=}
        \lev_{≥1}\left(\msum_N(\cxsprt_{B_1},\dots,\cxsprt_{B_m})\right)\\
        & \overset{\hphantom{L\ref{lem:gauge_epim}\ref{lem:gauge_epim_rad}}}{=}
        \lev_{≥1}\left(x^*\mapsto
	\cvsprt_N((\cvsprt_{B_1}(x^*),\dots,\cvsprt_{B_m}(x^*)))\right)\\
        &\overset{\:\,\eqref{eq:gauge_support_polar}\:\,}{=}
        \lev_{≥1}\left(x^*\mapsto
	\agauge_{N^\apolar}((\agauge_{B^\apolar_1}(x^*),\dots,\agauge_{B^\apolar_m}(x^*)))\right)\\
	&\!\overset{\mathrm{P}\ref{prop:polar_prop}(\ref{prop:polar_prop_gauge})}{=}
        \set{x^*\in\bnch^*;(\agauge_{B^\apolar_1}(x^*),\dots,\agauge_{B^\apolar_m}(x^*))\in N^\apolar}\\
        &\overset{\hphantom{L\ref{lem:gauge_epim}\ref{lem:gauge_epim_rad}}}{=}    
        \union_{\mu^*\in N^\apolar}\set{x^*\in\bnch^*;(\agauge_{B^\apolar_1}(x^*),\dots,\agauge_{B^\apolar_m}(x^*))=\mu^*}\\
        &\!\!\overset{\hphantom{L\ref{lem:gauge_epim}(\ref{lem:gauge_epim_rad})}}{=}    
        \union_{\mu^*\in N^\apolar}\set{x^*\in\bnch^*;\forall{i\in[m]},\ \agauge_{B^\apolar_i}(x^*)=\mu^*_i}.\label{eq:dual_msum_2}
        %
    \end{align}
    Observe that $N^\apolar$ is co-star-shaped, thus $[1,\infty)·N^\apolar = N^\apolar$. Hence
    \begin{align}
        \g\big(\msum_N(B_1,\dots,B_m))^{\apolar}
        &\overset{\:\,\eqref{eq:dual_msum_2}\:\,}{=}
        \union_{\mu^*\in N^\apolar}\set{x^*\in\bnch^*;\forall{i\in[m]},\ \agauge_{B^\apolar_i}(x^*)=\mu^*_i}\\
        &\overset{\hphantom{L\ref{lem:gauge_epim}\ref{lem:gauge_epim_rad}}}{=}    
        \union_{\mu^*\in
	[1,\infty)·N^\apolar}\set{x^*\in\bnch^*;\forall{i\in[m]},\ \agauge_{B^\apolar_i}(x^*)=\mu^*_i}\\
        &\overset{\hphantom{L\ref{lem:gauge_epim}\ref{lem:gauge_epim_rad}}}{=}    
        \union_{\mu^*\in
	N^\apolar}\union_{\lambda\in[1,\infty)}\set{x^*\in\bnch^*;\forall{i\in[m]},\ \agauge_{B^\apolar_i}(x^*)=\lambda\mu^*_i}\\
        &\overset{\hphantom{L\ref{lem:gauge_epim}\ref{lem:gauge_epim_rad}}}{=}    
        \union_{\mu^*\in N^\apolar}\set{x^*\in\bnch^*;\forall{i\in[m]},\ \agauge_{B^\apolar_i}(x^*)≥\mu^*_i}\\
        &\overset{\hphantom{L\ref{lem:gauge_epim}\ref{lem:gauge_epim_rad}}}{=}    
        \union_{\mu^*\in N^\apolar}\inter_{i\in[m]}\lev_{≥\mu_i^*}(\agauge_{B^{\apolar}_i})\\
	&\!\overset{\mathrm{L}\ref{lem:gauge_epim}(\ref{lem:gauge_epim_shad})}{=}
        \union_{\mu^*\in N^\apolar}\inter_{i\in[m]}\mu_i^*\star B^{\apolar}_i\\
        &\overset{\hphantom{L\ref{lem:gauge_epim}\ref{lem:gauge_epim_rad}}}{=}    
        \dualmsum_{N^\apolar}(B_1^\apolar,\dots,B_m^\apolar),
    \end{align}
    which completes the proof of claim \ref{thm:dual_msum_ccv}.
\end{proof}

We can take the polars of both sides of the above theorem to obtain the
result that the 
polar of the dual polar $M$-sum of the polars of $(A_i)_i$ is the $M$-sum
of $(A_i)_i$:
\begin{corollary}
    \label{cor:dual_msum}
    Let $A_i\in\rdnt(\bnch)$ and $B_i\in\shdy(\bnch)$ with $B_i\subseq\rec(B_i)$ for $i\in[m]$, $M\in\rdnt(\R^m)$, and $N\in\shdy(\Rp^m)$. Assume $M$ and $A_i$ for $i\in[m]$ are compact. Then
    \begin{enumerate}
        \item $\msum(A_1,\dots,A_m) = \g(\dualmsum_{M^\polar}(A_1^\polar, \dots,A_m^\polar))^\polar$, and
        \item $\msum_N(B_1,\dots,B_m) = \g(\dualmsum_{N^\apolar}(B_1^\apolar, \dots,B_m^\apolar))^\apolar$.
    \end{enumerate}
\end{corollary}
\begin{proof}
    From Theorem \ref{thm:radshad_to_radshad} we know
    $\msum(A_1,\dots,A_m)$ is radiant and $\msum_N(B_1,\dots,B_m)$ is
    shady. The bipolar theorem \eqref{eq:bipolar_theorem} applied to
    Theorem \ref{thm:dual_msum} gives
    \begin{gather}
        \overline{\msum(A_1,\dots,A_m)} = \g(\dualmsum_{M^\polar}(A_1^\polar, \dots,A_m^\polar))^\polar
	\mbox{\ \ and\ \ }
        \overline{\msum_N(B_1,\dots,B_m)} = \g(\dualmsum_{N^\apolar}(B_1^\apolar, \dots,B_m^\apolar))^\apolar. \label{eq:closed_apolar}
    \end{gather}
    Theorem \ref{thm:msum_closed} and Corollary \ref{por:msum_closed} makes the explicit closure operations redundant.
\end{proof}

The above duality result can also be expressed in terms of gauges:
\begin{corollary}\label{cor:gauge_msum}
   Let $A_i\in\rdnt(\bnch)$ and $B_i\in\shdy(\bnch)$ with $B_i\subseq\rec(B_i)$ for $i\in[m]$, $M\in\rdnt(\R^m)$, and $N\in\shdy(\Rp^m)$. Assume $M$ and $A_i$ for $i\in[m]$ are compact. Then
    \begin{enumerate}
        \item $\gauge_{\msum(A_1,\dots,A_m)} = \overline{\dualmsum_{\gauge_{M}}(\gauge_{A_1},\dots,\gauge_{A_m})}$, and\label{cor:gauge_msum1}
        \item $\agauge_{\msum_N(B_1,\dots,B_m)} = 
                    \overline{\dualmsum_{\agauge_{N}}(\agauge_{B_1},\dots,\agauge_{B_m})}$.\label{cor:gauge_msum2}
    \end{enumerate}
\end{corollary}
\begin{proof}
    The two identities are derived as follows:
    \begin{gather}
            \begin{aligned}
                \gauge_{\msum(A_1,\dots,A_m)} 
                &\!\!\!\overset{\:\,\eqref{eq:gauge_support_polar}\:\,}{=}
                \cxsprt_{\g(\msum(A_1,\dots,A_m))^\polar}\\
		&\overset{{\mathrm{T}\ref{thm:dual_msum}(\ref{thm:dual_msum_cvx})}}{=}
                \cxsprt_{{\dualmsum_{M}(A_1^\polar,\dots,A_m^\polar)}}\\
		&\overset{\,\mathrm{T}\ref{thm:dual_functional_msum}\,}{=}
                \overline{\dualmsum_{\cvsprt_{M^\polar}}(\cxsprt_{A_1^\polar},\dots,\cxsprt_{A_m^\polar})}\\
                &\overset{\:\,\eqref{eq:gauge_support_polar}\:\,}{=}
                \overline{\dualmsum_{\gauge_{M}}(\gauge_{A_1},\dots,\gauge_{A_m})},
            \end{aligned}
	    \hspace*{1cm}
            \begin{aligned}
                \agauge_{\msum(B_1,\dots,B_m)} 
                &\overset{\:\,\eqref{eq:gauge_support_polar}\:\,}{=}
                \cvsprt_{\g(\msum_N(B_1,\dots,B_m))^\apolar}\\
		&\!\!\overset{{\mathrm{T}\ref{thm:dual_msum}(\ref{thm:dual_msum_ccv})}}{=}
                \cvsprt_{{\dualmsum_{N^\apolar}(B_1^\apolar,\dots,B_m^\apolar)}}\\
		&\overset{\,\mathrm{T}\ref{thm:dual_functional_msum}\,}{=}
                \overline{\dualmsum_{\cvsprt_{N^\apolar}}(\cvsprt_{B_1^\apolar},\dots,\cvsprt_{B_m^\apolar})}\\
                &\overset{\:\,\eqref{eq:gauge_support_polar}\:\,}{=}
                \overline{\dualmsum_{\agauge_{N}}(\agauge_{B_1},\dots,\agauge_{B_m})}.
            \end{aligned}
        \end{gather}    
\end{proof}
Finally, the duality result implies the following result expressed in terms
of proper loss functions.
\begin{corollary}\label{cor:proper_loss_msum_polar}
    Let $\ell_1,\dots,\ell_m$ be a sequence of proper loss functions with
    conditional Bayes risks $\minL_1,\dots,\minL_m$. Let
    $\minM:\Rp^m\to\Rx$ be another conditional Bayes risk function. Then
    \begin{gather}
            \ell\in\subdiff\msum_{\minM}(\minL_1,\dots,\minL_m ) 
	    \iff \ell^\apolar\in\subdiff\overline{\dualmsum_{\minM^\apolar}(\minL_1^\apolar,\dots,\minL_m^\apolar)}.
    \end{gather}
\end{corollary}

\subsection{Conclusion on $M$-sums and New Losses from Old}
\label{sub:common_texorpdfstring}

The above development shows a general and powerful way of combining
existing proper losses into new ones. An intriguing and satisfying feature
of the results is that, in essence, the way you combine several proper
losses into a new one, is to combine them using yet another proper loss!
The $M$-sum operations are thus a very natural means to combine multiple
existing proper loss functions to create a new proper loss function.

To provide some intuition, several classical 
examples of \emph{convex} $M$-sums are presented in Table
\ref{table:common-M-sums}. Confer \citep{Seeger1990} who uses different
notation.  See also \citep{Mesikepp:2016tf} and \citep{Gardner:2013aa} and
\cite[Chapter 1]{Kusraev:1995aa}.

    \begin{table}
	    \arrayrulecolor{lightgray}
        \centering
	\small
        \begin{tabular}{c c l c }\toprule
            $M$ 
                & $\msum_{M}(A_1,\dots,A_m)$ 
                        & Operation
                                & $\msum_{\cxsprt_M}(\cxsprt_{A_1},\dots,\cxsprt_{A_m})$
		\\\midrule
            $\single{1_m}$
                & $A_1+\dots+A_m$
			& Minkowski sum
			& $\displaystyle\sum_{i=1}^m \cxsprt_{A_i}$
			\\[1mm]
	    $\probm([m])$
	    &  $\co\left(A_1\cup\cdots \cup A_m\right)$ 
	    &  Convex union
			&
			$\displaystyle\bigvee_{i=1}^m\sigma_{A_i}$
			\\[2mm]
	\toprule
	    $M$
                & $\dualmsum_{M}(A_1,\dots,A_m)$ 
                        & 
                                & $\dualmsum_{\cxsprt_M}(\cxsprt_{A_1},\dots,\cxsprt_{A_m})$ %
					\\\midrule
            $\single{1_m}$
                & $A_1\cap\dots\cap A_m$
                        & Intersection
                                & $\cxsprt_{A_1}\infconv\dots\infconv\cxsprt_{A_m}$
				\\[1mm]
	    $\probm([m])$
                &  $A_1\mathbin{\Diamond}\cdots\mathbin{\Diamond} A_m$
		&  Inverse sum 
			&  $\displaystyle\inf_{a_1+\cdots a_m=x}
			\bigvee_{i=1}^m \sigma_{A_i}$
			\\
            \bottomrule
        \end{tabular}
	\caption{\label{table:common-M-sums} Examples of \emph{convex}
	$M$-sums and dual $M$-sums. The inverse sum is discussed in
	\citep[page 21]{Rockafellar:1970}.}
    \end{table}
\section{Conclusion} 
\label{sec:conclusion}
We have presented a geometric theory of proper losses, whereby we take an
unbounded convex set with particular properties as the starting point, and
derive the proper loss as the (sub)-gradient of the support function of the
set. The new perspective shows the natural duality between a loss and
the ``antipolar loss'' newly introduced here, which is of value in understanding
Vovk's aggregating algorithm \citep{Kamalaruban:2015aa}. It also shows how
one can generally combine multiple proper losses to create new proper
losses. In developing that combinatorial theory, we extended a number of
results on $M$-sums, and in particular proved an elegant and general
duality theorem. The theory shows a deep but simple relationship between
loss functions and concave gauges, the concave analog of a norm. Such
concave gauges arise naturally in economics, but until now they have not
been explicitly utilised within machine learning.  We have presented the
theory for finite outcomes ($n$-class probability estimation). Many of the
results extend to a more general setting \citep{Cranko:2021aa}.

The geometry of losses complements existing geometric approaches to machine
learning, which have focussed on the geometry of the data distribution
(through its likelihood function) \citep{Amari:2016aa}, the geometry of
the prior \citep{Mahony:2001aa}, and the geometry of the model class
\citep{Lee:1998wr,Mendelson:2002wk}.

The geometric theory developed here enables a general and insightful
perspective relating Bayes risks and measures of information extending the
results in \citep{Reid2011}  and \citep{Garcia-Garcia2012}. 
By developing measures of information in
terms of convex sets (directly related to the superprediction sets used in the
present paper) one can extend and refine 
the famous data processing theorem of information theory
\citep{Williamson:2022aa}.

Given the foundational role loss functions play in a wide range of machine
learning problems, it seems reasonable to suppose that the theory presented
here will lead to further insights. One concrete direction for future work
is to relate the geometry of the loss function developed here to the
geometry of hypotheses classes and thus illuminate the interaction between
losses and hypothesis classes that controls the speed of convergence in
learning problems \citep{van-Erven:2015aa}. Ideally one would have a
geometric theory that simultaneously incorporated the loss function $\ell$, 
the hypothesis class $\mathcal{F}$, the distribution of the data $P$,
and prior knowledge or belief encoded in some structure such as a
Riemannian metric \citep{Mahony:2001aa}. 



\section{Acknowledgements} 
\label{sec:acknowledgements}
This work was inspired by conversations between RW and  Mark Reid during
2009--2012. An early version of some of the results appeared in COLT2014.
RW's contribution was funded in part by NICTA and
the Deutsche Forschungsgemeinschaft
under Germany’s Excellence Strategy –-
EXC number 2064/1 –- Project number 390727645.   Thanks to an
extraordinarily assiduous referee for catching many typos and glitches in an earlier
version.

\setlength{\bibsep}{0.0pt}
\bibliography{bibliography.bib}
\end{document}